%% file: main.tex
\documentclass{article}

\usepackage[utf8]{inputenc} 
\usepackage[T1]{fontenc}    
\usepackage{hyperref}       
\usepackage{url}            
\usepackage{booktabs}       
\usepackage{amsfonts}       
\usepackage{nicefrac}       
\usepackage{microtype}      
\usepackage{xcolor}         

\usepackage{multirow}
\usepackage{stfloats}
\usepackage{array}
\usepackage[export]{adjustbox}
\usepackage{placeins}
\usepackage{float}
\usepackage{amsmath}
\usepackage{subcaption}
\usepackage{authblk}

\usepackage{algorithm}
\usepackage{algpseudocode}
\usepackage{amssymb}

\usepackage{enumerate}

\usepackage{makecell}

\usepackage[margin=1in]{geometry}

\usepackage{natbib}

\newtheorem{theorem}{Theorem}[section]
\newtheorem{proposition}[theorem]{Proposition}

\newtheorem{assumption}{Assumption}
\usepackage[dvipsnames]{xcolor}
\usepackage[]{color-edits}
\addauthor{ab}{blue}
\addauthor{abb}{red}
\addauthor{gt}{violet}

\title{Gradual Fine-Tuning for Flow Matching Models}

\author[1]{Gudrun Thorkelsdottir}
\author[1]{Arindam Banerjee}

\affil[1]{University of Illinois Urbana-Champaign}
\affil[ ]{\texttt{gudrunt2@illinois.edu, arindamb@illinois.edu}}

\date{}

\begin{document}

\maketitle

\begin{abstract}
  Fine-tuning flow matching models is a central challenge in settings with limited data, evolving distributions, or computational constraints. While recent work has produced significant advances, particularly in the area of reward-based fine-tuning, current methods fail to demonstrate both theoretical correctness as well as strong empirical results in terms of stability, efficiency, and diversity preservation. In this work, we propose Gradual Fine-Tuning (GFT), a simple yet principled annealing-based framework for fine-tuning flow generative models when only samples from the target distribution are available. For stochastic flows, GFT defines a temperature-controlled sequence of intermediate objectives that smoothly interpolate between the pretrained and target drifts, provably approaching the true target as the temperature approaches zero. We analytically demonstrate that sample generation after GFT can be made substantially more efficient with the use of arbitrary (e.g., optimal transport) couplings, as well as by utilizing few-step inference methods. Empirically, GFT significantly improves convergence stability, while maintaining or improving generation quality, training speed, and generation diversity compared to other fine-tuning methods. Our results position GFT as a simple yet theoretically grounded and practically effective alternative for scalable adaptation of flow matching models under distribution shift.
\end{abstract}

\section{Introduction}
\label{sec:intro}

\input{introduction}

\section{Related Work}
\label{sec:related}

\input{related}

\section{Background and Preliminaries}
\label{sec:prelim}
\input{preliminaries}

\section{Gradual Fine-Tuning}
\label{sec:gft}
\input{gft}

\section{Experiments}
\label{sec:expt}
\input{experiments}

\section{Discussion}
\label{sec:discuss}
\input{discussion}

\newpage
\bibliographystyle{plainnat}
\bibliography{refs}


\appendix

\input{app_tech1}

\input{app_expt}

\input{app_tech2}


\end{document}

%% file: introduction.tex
Recent advances in flow-based generative modeling \cite{lipman2023flow, liu2023flow} have achieved remarkably high generation quality, and have seen widespread use in image generation \cite{pmlr-v235-esser24a}, as well as scientific applications such as molecule generation \cite{jing2024alphafold, EqFM, EqFM_Molecule}, biological sequence modeling \cite{stark2024dirichlet, Fisher}, and atmospheric forecasting \cite{fotiadis2025adaptive, bao2024ensemble}. In settings characterized by limited computational resources, data-scarcity, or evolving data distributions, fine-tuning a pretrained flow matching model can be the most effective approach for achieving accurate generation. Unlike training from random initialization, fine-tuning must reconcile the competing objectives of adapting to a new target distribution while preserving the beneficial properties learned during pretraining, such as short probability paths or high diversity. 

In order to achieve this balance, an ideal fine-tuning method for flow generative models must satisfy several key criteria. Importantly, it should be \textit{theoretically correct}, with provable convergence to the desired target distribution under reasonable assumptions. To ensure practical utility, the method must maintain \textit{stability}, with limited performance oscillations, and should ideally maintain the efficient \textit{simulation-free} training of flow matching models \cite{lipman2023flow}. Fine-tuned models should also have \textit{high inference efficiency} by facilitating few-step inference \cite{geng2026mean, boffi2025buildconsistencymodellearning} and through efficient coupling between source and target distributions, e.g., based on Optimal Transport (OT), that accelerate generation \cite{tong2024improving, multisampleFM}. Importantly, these properties must produce high empirical performance, matching or outperforming alternative fine-tuning methods in generation accuracy and diversity.

Recent work in fine-tuning for flow matching models has made significant progress, particularly in reward-based methods. However, existing methods do not fully satisfy the key criteria listed above. Adjoint Matching \cite{domingo-enrich2025adjoint} casts reward fine-tuning as Stochastic Optimal Control (SOC), and provides convergence guarantees but requires multiple trajectory simulations for each fine-tuning step. Additionally, instead of directly optimizing the parameters of the pretrained model, Adjoint Matching and other SOC-based methods optimize separate control vector fields which are added linearly to the pretrained drift, making them incompatible with nonlinear fine-tuning techniques such as full fine-tuning and LoRA \cite{hu2022lora}. Beyond SOC-based fine-tuning there exists a large class of reward-based methods, handling both differentiable \cite{clark2024directly, xu2023imagereward, prabhudesai2023aligning} and non-differentiable rewards \cite{black2024training, fan2023reinforcement, fan2025online}. However, these methods largely suffer from high gradient instability, and inherently require trajectory simulation to calculate rewards, sacrificing the efficiency of simulation-free training.

An important and practically relevant setting remains underexplored: fine-tuning when samples from the target distribution are directly available. This scenario arises naturally in domain adaptation and dataset drift applications, and removes the need to define or learn an external reward function. An existing method in this category is standard fine-tuning, defined as continued training with a new dataset. Standard fine-tuning provides no principled mechanism for controlled adaptation, treating the pretrained model merely as an initialization and ignoring its learned dynamics during fine-tuning, often leading to high instability. 

We propose \textbf{Gradual Fine-Tuning} (GFT), a principled framework for adapting pretrained flow generative models with access to target distribution samples. GFT uses  annealed regularization to create a temperature-controlled sequence of intermediate objectives that smoothly interpolate between pretrained and target dynamics. The result is a gradual, highly stable, and reliably controllable transition to the fine-tuning target. We show that GFT addresses the limitations of existing approaches while satisfying all key desiderata for flow model fine-tuning outlined above. The main contributions of our work are as follows:

\begin{enumerate}[(C-1)]  
    \item We prove that GFT converges to the target distribution under standard assumptions. 
    \item We establish equivalent results for GFT when conditioning on arbitrary source-target couplings, allowing for the use of training techniques that enhance inference efficiency. 
    \item We demonstrate the natural extension of GFT to fine-tuning for few-step inference methods without sacrificing theoretical correctness.
    \item We empirically demonstrate the strong performance of GFT across several experimental settings, showing that it achieves comparable or superior generation quality to other methods while providing improved training stability, generation diversity, and consistently shorter probability paths.
\end{enumerate}

This work presents the first systematic investigation of annealed gradual fine-tuning for flow matching models using a temperature parameter, and establishes a foundation for scalable and efficient adaptation of flow-based generative models.

%% file: related.tex
Previous work in fine-tuning for flow generative models can be organized along two key dimensions: the type of supervision available, and the methodological approach.

\paragraph{Available Supervision}   A substantial body of work focuses on fine-tuning generative models to adapt to a tilted distribution implicitly defined by a reward function. When the reward function is differentiable, these methods can directly backpropagate the reward gradients \cite{clark2024directly, domingo-enrich2025adjoint, xu2023imagereward, prabhudesai2023aligning}. In contrast, policy gradient and reinforcement learning methods are commonly used when the reward function is not differentiable \cite{black2024training, fan2023reinforcement, fan2025finetuningflowmatchinggenerative}. Although reward-based methods can be sample efficient and highly generalizable, they often face high training instability and reward hacking when the true reward is approximated by a trained model. Another class of methods directly optimize over pairwise preference data when a reward function is unavailable or difficult to define \cite{wallace2023diffusion, liang2025aestheticposttrainingdiffusionmodels}. When samples from the fine-tuning target distribution are available, the most straightforward approach is standard fine-tuning. GFT also belongs to this category, but incorporates annealed regularization to yield strong theoretical guarantees and improve upon the instability associated with standard fine-tuning.

\paragraph{Methodology}   Several recent works formulate fine-tuning as stochastic optimal control (SOC) \cite{domingo-enrich2025adjoint, uehara2024finetuningcontinuoustimediffusionmodels, potaptchik2025tiltmatchingscalablesampling}. Although SOC methods provide strong theoretical guarantees, they are incompatible with few-step inference methods, and the required trajectory simulation during training also makes them time and memory intensive. Beyond SOC formulations, many methods perform direct optimization using reward gradients \cite{clark2024directly, prabhudesai2023aligning} or samples, such as standard fine-tuning and GFT. A separate class of methods apply inference-time guidance or conditioning to alter the generation of samples instead of fine-tuning the pretrained model \cite{ho2022classifierfreediffusionguidance, ben-hamu2024dflow}. While computationally appealing, these methods typically increase inference cost, and may not be capable of handling large distribution shifts as the parameters of the pretrained model remain unchanged.

%% file: preliminaries.tex
\subsection{Flow Matching} 

Flow Matching \citep{lipman2023flow, liu2023flow} is a family of generative modeling approaches which iteratively simulate a continuous process between a source distribution \(p_0\) and a data-defined target distribution \(p_1\). Flow Matching defines a generative model by learning a time-varying vector field \(v_\theta(X_t, t):\mathbb{R}^d \times \mathbb{R} \rightarrow \mathbb{R}^d\) which transports a family of probability densities \(\{p_t\}_{t \in [0,1]}\) along a deterministic path. The evolution of a single sample through this vector field is described by the following ordinary differential equation (ODE)
\begin{equation}
    dX_t = v_\theta(X_t, t) dt~, \qquad X_0 \sim p_0 ~.\label{eq:ODE}
\end{equation}

Integrating \eqref{eq:ODE} over time therefore enables simulated generation from the target distribution. The vector field \(v_\theta\) also describes the deterministic shift of probability mass from \(p_0\) to \(p_1\). To achieve a well-defined continuous transport process, each intermediate density \(p_t\) must satisfy the continuity equation \citep{villani2008optimal} \eqref{eq:continuity} with the vector field \(v_t(\cdot) := v_\theta(\cdot, t)\),
\begin{equation}
    \partial_tp_t + \nabla_x \cdot (p_t v_t) = 0~, \label{eq:continuity}
\end{equation}

where \(\nabla_x \cdot\) denotes the divergence operator with respect to \(x\). The continuity equation enforces the conservation of probability mass across time, such that the change in density at any spatial point is exactly determined by the local flow of probability mass caused by the vector field. As a result, given the vector field and the initial density, the continuity equation uniquely determines the entire continuous sequence of intermediary densities \(\{p_t\}_{t \in [0,1]}\). This is a crucial feature of deterministic flow-based generative models, as it enables exact likelihood calculation through the instantaneous change of variables formula, which can be directly derived using the continuity equation.

Despite achieving impressive generation quality in many applications, the iterative nature of flow-based generative models significantly reduces their inference efficiency compared to earlier single-step generative models \citep{Kingma2014, NIPS2014_f033ed80}. However, several variations of the training algorithm, such as minimizing vector field curvature \cite{liu2023flow} or sampling according to OT plans \cite{tong2024improving}, have proven to lead to highly efficient inference-time generation.

\subsection{Stochastic View of Flow Matching}

Despite their deterministic formulation, flow matching models can be embedded into a class of stochastic differential equations (SDEs) with time marginals that match \(p_t\). Consider the following SDE
\begin{equation}
    dX_t = \tilde{v}_t(X_t)dt + \sigma_t(X_t)dB_t, \qquad X_0 \sim p_0 ~, \label{eq:SDE}
\end{equation}
with drift term \(\tilde{v}:\mathbb{R}^d \times \mathbb{R} \rightarrow \mathbb{R}^d\) and diffusion term \(\sigma:\mathbb{R}^d \times \mathbb{R} \rightarrow \mathbb{R}^{d \times d}\). Assuming isotropic diffusion, the Fokker-Planck equation describing the evolution of the density \(p_t\) under the dynamics of this SDE is
\begin{equation}
    \partial_tp_t = -\nabla_x \cdot (\tilde{v}_t p_t) + \frac{1}{2}\nabla_x \cdot \nabla_x \cdot (\sigma_t^2 I  p_t) \label{eq:FP}~.
\end{equation}

The Fokker-Planck equation can be described as a natural generalization of the continuity equation to the stochastic setting. The change in density at a given spatial point is still described by the deterministic flow of probability mass caused by the drift term, now in addition to the spread of probability mass due to the stochastic noise \(dB_t\). When \(\sigma_t = 0\), the noise term vanishes and the Fokker-Planck equation reduces exactly to the continuity equation. 

In order to exactly match the ODE density evolution, we substitute the continuity equation \eqref{eq:continuity} into the Fokker-Planck equation \eqref{eq:FP}, and solve for the drift term \(\tilde{v}_t\) (details shown in Appendix \ref{ap:recover_score}).
\begin{align}
    \nabla_x \cdot (p_t v_t) & = \nabla_x \cdot (\tilde{v}_t p_t) - \frac{1}{2} \nabla_x \cdot \nabla_x \cdot (\sigma_t^2 I p_t)~, \\
    \Rightarrow \quad \tilde{v}_t & = v_t + \frac{1}{2} \sigma_t^2 \nabla_x \log p_t + Z_t~,
\end{align}
where \(Z_t\) is a divergence free velocity which can simply be set to 0. This formulation explicitly introduces the score function of the density \(p_t\), and has therefore motivated the exploration of the relationship between deterministic Neural ODEs and diffusion models \cite{song2021scorebased}, as well as the related stochastic interpolant model \cite{albergo2025stochasticinterpolantsunifyingframework}. We recover the exact ODE drift term by setting \(\sigma_t = 0\).

%% file: gft.tex
\subsection{Problem Setup}
Assume that we have access to a pretrained flow matching model with parameters \(\theta_0\) which generates samples from a distribution \(p_1\). The model has been trained to simulate the base continuous process 
\begin{equation}
    \mathbb{P}_{\theta_0} : dX_t = v_{\theta_0}(X_t, t)dt + \sigma_tdB_t, \qquad X_0 \sim p_0~,
\end{equation}
where \(v_{\theta_0}\) is the vector field transforming samples from the source distribution \(p_0\) to \(p_1\), and \(B_t\) is a standard \(d\)-dimensional Brownian motion. We now want to fine-tune this model to generate samples from a new target distribution \(q\), from which we have access to samples. The target fine-tuned continuous process is defined as
\begin{equation}
    \mathbb{P}_q : dX_t = v_q(X_t, t)dt + \sigma_tdB_t, \qquad X_0 \sim p_0~.
\end{equation}
where \(v_q\), the vector field generating samples from \(q\), is the target vector field which is unknown during fine-tuning. The goal of fine-tuning is to learn a new process \(\mathbb{P}_\theta\) which closely approximates \(\mathbb{P}_q\), 

\begin{equation}
    \mathbb{P}_\theta : dX_t = v_\theta(X_t, t)dt + \sigma_tdB_t, \qquad X_0 \sim p_0~.
\end{equation}
The fine-tuned vector field \(v_\theta\) is initialized with the parameters of the base vector field \(v_{\theta_0}\), and then optimized to match the target vector field \(v_q\) through fine-tuning. 

Here, \(\mathbb{P}_{\theta_0}, \mathbb{P}_q, \text{and } \mathbb{P}_\theta\) are  probability path measures on the path space \(C([0,1], \mathbb{R}^d)\) induced by their corresponding SDEs \cite{alma9919303981206531}. As such, they assign likelihoods to trajectories \(\{X_t\}_{t \in [0,1]}\), rather than individual points \(X_t\), and therefore allow us to reason about the full generative process. For each process, the distribution of the terminal state \(X_1\) generated by the SDE is defined by its corresponding terminal marginal. In particular, under \(\mathbb{P}_{\theta_0}\) we have \(X_1 \sim p_1\), and under \(\mathbb{P}_q\) we have \(X_1 \sim q\). Therefore, when fine-tuning successfully adapts \(\mathbb{P}_\theta\) to closely match \(\mathbb{P}_q\), its terminal marginal will match the target distribution \(q\), under suitable assumptions.

\subsection{Gradual Fine-Tuning Objective}\label{sc:GFT_objective}

Consider the following optimization over SDE-induced path measures:
\begin{equation}
\begin{split}
    \min_{\theta} ~& ~~ \text{KL}(\mathbb{P}_{\theta} \| \mathbb{P}_q) + \beta \text{KL}(\mathbb{P}_{\theta} \| \mathbb{P}_{\theta_0})   \\
    \text{s.t.} &~~ dX_t = v_{\theta}(X_t, t) dt + \sigma_t dB_t~.
\end{split}
\label{eq:GFT_objective_PM}
\end{equation}
The first term of the cost expression promotes alignment with the target process $\mathbb{P}_q$, while the second term acts as a divergence regularizer that penalizes large deviations from the pretrained dynamics. The coefficient $\beta \in \mathbb{R}_+$ is a temperature parameter that controls the strength of this regularization. The GFT objective can therefore be viewed as an interpolation between two transport processes, one which guides samples to the terminal distribution \(q\) by matching the target process \(\mathbb{P}_q\), and another which maintains the pretrained dynamics by penalizing deviation from \(\mathbb{P}_{\theta_0}\). Balancing these costs suggests that the optimal fine-tuned model should be a weighted average of the base and target dynamics, controlled by the temperature \(\beta\). We now formalize this intuition in the following theorem. 
\begin{theorem}\label{thm:opt_GFT}
    Let \(\mathbb{P}_{\theta_0}, \mathbb{P}_q, \text{and } \mathbb{P}_\theta\) be path measures induced by SDEs with drift terms \(v_{\theta_0}, v_q, \text{and } v_\theta\), respectively. Assume that these processes share the diffusion coefficient \(\sigma_t\). Then, for a given temperature \(\beta\), the vector field minimizing the GFT objective (\ref{eq:GFT_objective_PM}) is given by the convex combination
    \begin{equation}
        v_\theta^*(X_t, t) = \left( \frac{1}{1 + \beta} \right) v_q(X_t, t) + \left( \frac{\beta}{1 + \beta} \right) v_{\theta_0}(X_t, t) ~ .
    \label{eq:GFT_optimum}
    \end{equation}
    Proof given in Appendix \ref{ap:min_GFT}.
\end{theorem}
Theorem \ref{thm:opt_GFT} states that minimizing the GFT objective (\ref{eq:GFT_objective_PM}) results in a closed-form and continuous optimal drift which is a weighted arithmetic mean of the pretrained and target vector fields. The weighting is directly controlled by the temperature \(\beta\), resulting in limiting behavior which recovers the pretrained and target dynamics:
\begin{equation}
    \lim_{\beta \rightarrow \infty} v^*_\theta = v_{\theta_0}~,  \qquad  \lim_{\beta \rightarrow 0^+} v^*_\theta = v_q \label{eq:beta_lim}~.
\end{equation}
\paragraph{Annealed Regularization.} The limiting behavior of the optimal vector field \(v_\theta^*\) provides a  clear motivation for the gradual nature of our method. To control the tradeoff between preserving pretrained knowledge and adapting to the new target distribution, GFT uses a time-dependent cooling schedule. At the start of fine-tuning, a high temperature strongly penalizes deviation from the base model, keeping updates close to the pretrained drift and preventing premature collapse towards the target. As training progresses, \(\beta\) is gradually annealed toward \(0^+\), smoothly relaxing this constraint and allowing the fine-tuned drift \(v_\theta\) to asymptotically recover the target vector field \(v_q\) (\ref{eq:beta_lim}). This progression creates a gradual transition between the pretrained and target dynamics, improving training stability and final generation diversity. Similar gradual adaptation methods have previously been shown to stabilize and improve classical domain adaptation \cite{gradual_DA}, and related trust region methods have recently been considered for diffusion model fine-tuning \cite{blessing2025trust}.

\paragraph{Tractable Reformulation.}
Although the GFT objective \eqref{eq:GFT_objective_PM}  provides a simple and elegant formulation for gradual fine-tuning of flow-based models, it cannot be directly optimized for two reasons. First, it implicitly depends on the vector field \(v_q\) through the path measure \(\mathbb{P}_q\), which is unknown during fine-tuning. Second, the objective is defined over the space of path measures, which are not differentiable with respect to the model parameters \(\theta\). Fundamentally, the neural network being optimized during fine-tuning represents the drift term of the SDE which induces the path measure \(\mathbb{P}_\theta\), not the measure itself. In order to perform optimization, the objective must be rewritten in terms of the corresponding drift terms of each path measure. In the following proposition, we demonstrate that the application of Girsanov's Theorem achieves this result. 
\begin{proposition}\label{pr:KL_Girsanov}
    Let \(\mathbb{P}_f\) and \(\mathbb{P}_g\) be path measures induced by SDEs with drift terms \(f_t\) and \(g_t\), respectively. Assume that these processes share the diffusion coefficient  \(\sigma_t\), are mutually absolutely continuous, and draw initial samples from the same source distribution. Following Girsanov's theorem, the KL divergence between these path measures is given by:
    \begin{equation}
        KL(\mathbb{P}_f \| \mathbb{P}_g) = \mathbb{E}_{ X_{[0,1]} \sim \mathbb{P}_f} \left[ \frac{1}{2} \int_0^1 \| \sigma_t^{-1} (f_t(X_t) - g_t(X_t)) \|^2 dt \right]
    \label{eq:KL_girsanovs}
    \end{equation}
    Proof given in Appendix \ref{ap:PL_KL}.
\end{proposition}
We apply Proposition \ref{pr:KL_Girsanov} to (\ref{eq:GFT_objective_PM}) to write the tractable form of the GFT objective (details given in Appendix \ref{ap:Prac_Impl}). Note that we set \(\sigma_t = \textit{I}\), as we find that this leads to empirically strong results while maintaining simplicity. 
\begin{equation}
    \mathcal{L}(\theta) = \mathbb{E}_{ \mathbb{P}_{\theta}} \left[ \frac{1}{2} \int_0^1 (\| v_\theta - v_{q} \|^2 + \beta \| v_\theta - v_{\theta_0} \|^2) dt \right] 
\label{eq: GFT_objective_emp}
\end{equation}
While this reformulation successfully shifts optimization from the path space to the space of vector fields, enabling direct optimization of the neural network \(v_\theta\), the dependence on \(v_q\) still exists. In the next section, we eliminate this dependency by following the CFM framework to reformulate the marginal GFT objective as a conditional objective.

\subsection{Conditional GFT with Source-Target Couplings}\label{sc:Cond_GFT}

A key property of GFT is that conditioning on arbitrary couplings between the source and target distributions does not alter the unconditional optimization results. This enables the use of OT couplings during fine-tuning, creating an opportunity to substantially improve generation efficiency. 

Let \(\pi \in \Pi(p_0, q)\) be any coupling between the source and target distributions with marginals \(\int \pi(X_0, X_1)dX_1=p_0(X_0)\) and \(\int \pi(X_0, X_1)dX_0=q(X_1)\). Suppose that training samples are drawn jointly, and define \(Z := (X_0, X_1) \sim \pi\). We can then define the conditional GFT objective as 
\begin{equation}
    \mathcal{L}_\pi(\theta) = \mathbb{E}_{Z \sim \pi} \big[ KL(\mathbb{P}_\theta \| \mathbb{P}_q(\cdot | Z)) + \beta KL(\mathbb{P}_\theta \| \mathbb{P}_{\theta_0}) \big]
\label{eq:cond_obj_intractable}
\end{equation}
where \(\mathbb{P}_q(\cdot | Z)\) is the conditional target path measure induced by the SDE   \(dX_t = v_q(X_t, t| Z)dt + \sigma_t dB_t\). Recall that the pretrained base process \(\mathbb{P}_{\theta_0}\) is fixed, and therefore remains unconditional. Crucially, the fine-tuned process \(\mathbb{P}_\theta\) must also be unconditional in order to perform generation at inference time without access to target distribution samples. Instead, its underlying drift term \(v_\theta\) is trained to approximate the optimal conditional vector field marginalized over \(Z \sim \pi\). 

Following CFM, the unknown conditional target drift is replaced by a tractable conditional vector field which marginalizes to the true target, such that \(\mathbb{E}_{Z \sim \pi}[v_q(X_t, t|Z)] = v_q(X_t, t) \) \cite{lipman2023flow}. A common choice for the conditional vector field is the linear interpolation path between the given source and target samples, which satisfies this marginalization property \cite{liu2023flow}. Applying Proposition \ref{pr:KL_Girsanov} to \eqref{eq:cond_obj_intractable} and substituting the conditional target yields the final GFT loss:
\begin{equation}
    \mathcal{L}_\pi(\theta) = \mathbb{E}_{Z \sim \pi,~X_t \sim \mathbb{P}(\cdot | Z),~t \sim \mathcal{U}[0,1]} \bigg[  \frac{1}{2}  ( \|v_\theta(X_t, t) - v_q(X_t, t|Z)\|^2 + \beta \|v_\theta(X_t, t) - v_{\theta_0}(X_t, t) \|^2 ) \bigg]~.
\label{eq:cond_GFT_tractable}
\end{equation}
This loss is now tractable, and can be optimized directly over the parameters of the neural network \(v_\theta\). When the unconditional base and target vector fields are defined as the marginals of their conditional counterparts, we can show that the gradients of the conditional objective \(\mathcal{L}_\pi\) will be exactly equivalent to the gradients of the unconditional GFT objective \(\mathcal{L}\) (proof given in Appendix \ref{ap:Grad_equivalence}). Optimizing this conditional loss therefore preserves the theoretical guarantees of the unconditional objective, while enabling the use of arbitrary couplings, including OT, during fine-tuning without introducing bias.

\vspace{-0.8em}
\paragraph{Remark on relationship to CFM. } We note the close similarity between the tractable conditional GFT objective and the original CFM objective proposed in \cite{lipman2023flow}. The first term inside the expectation of \eqref{eq:cond_GFT_tractable} is exactly the regression loss used in the CFM objective. The second term is of the same form, but matches the pretrained vector field rather than the target vector field. The result of this relationship is that GFT is highly related to standard CFM, and therefore inherits its training efficiency and sampling accuracy. This also has the beneficial effect that any future advancements made to the flow matching algorithm will likely be applicable to GFT without significant changes. We make further remarks on the relationship between GFT and CFM in the following section.

\subsection{GFT Convergence}
\label{sc:Convergence}

The gradual adaptation resulting from the annealed regularization of GFT represents its key difference from standard CFM fine-tuning. After initialization, standard fine-tuning ignores the dynamics of the pretrained model, immediately adapting directly to the new target distribution. In section \ref{sec:expt} we empirically demonstrate that this sudden shift often leads to highly unstable fine-tuning. In contrast, the sequence of intermediary targets created by GFT results in a smooth and controlled transition to the new target distribution, creating stability while maintaining accurate convergence. The temperature parameter \(\beta\), which controls the distance between each subsequent GFT target, can be used to directly control the rate of convergence. This intuition is formally stated in Theorem \ref{thm:convergence}.

\begin{assumption}[Function Space Minimizer]
     For any given \(\beta > 0\) the vector field \(v_\theta\) exactly minimizes \(\mathcal{L_\pi(\theta)}\) in the function space, resulting in a minimizer of the form \eqref{eq:GFT_optimum}, which we will call \(v_\beta^*\).
\label{as:minimizer}
\end{assumption}

\begin{assumption}[Bounded Vector Field Mismatch]
    The pretrained vector field \(v_{\theta_0}\) and the marginal target vector field \(v_q\) satisfy:
        \begin{equation}
            C' := \int_0^1 \mathbb{E}_{\mathbb{P}_q} \|v_{\theta_0}(X^q_t,t) - v_q(X^q_t,t)\|^2\,dt < \infty~.
        \label{eq:Cprime}
        \end{equation}

        The constant \(C'\) then measures the \(L^2\) distance between the pretrained and target vector fields under the target path measure, and increases as the distribution shift between \(p_1\) and \(q\) increases.
\label{as:mismatch}
\end{assumption}

\begin{assumption}[Bounded Conditional Variance]
    For any jointly drawn sample \(Z\), the conditional variance of \(v_q(X_t, t|Z)\) around its mean \(v_q(X_t, t)\) is finite, such that 

        \begin{equation}
            V := \mathbb{E}_{Z \sim \pi}\int_0^1 \| v_q(X_t,t) - v_q(X_t,t \mid Z)\|^2\,dt < \infty~.
            \label{eq:V}
        \end{equation}

        Note that this quantity is equivalent to the standard CFM loss calculated at the sample \(Z\).
\label{as:variance}
\end{assumption}

\begin{assumption}[Lipschitz Vector Fields]
  Both $v_q$ and $v_{\theta_0}$ are Lipschitz continuous in $x$, uniformly in $t$, such that there exist constants $L_q, L_0 < \infty$ that for all $x, y \in \mathbb{R}^d$ and $t \in [0,1]$:
        \begin{equation}
            \|v_q(x,t) - v_q(y,t)\| \leq L_q\|x-y\|, \qquad \|v_{\theta_0}(x,t) - v_{\theta_0}(y,t)\| \leq L_0\|x-y\|
            \label{eq:lipschitz}
        \end{equation}
        We write $L = \max(L_q, L_0)$.
\label{as:lipschitz}
\end{assumption}

\begin{theorem}\label{thm:convergence}

    Under Assumptions \ref{as:minimizer}--\ref{as:lipschitz}, the following hold.

    \begin{enumerate}
    \item[(i)] $L^2$ \textbf{convergence of vector fields.}
    \(v_\beta^*\) converges to  \(v_q\) in \(L^2(\mathbb{P}_q)\) as \(\beta \to 0^+\),
    with the explicit rate:
    \begin{equation}
        \int_0^1 \mathbb{E}_{\mathbb{P}_q}\|v_\beta^*(X_t,t) - v_q(X_t,t) \|^2\,dt
        \leq \frac{\beta^2}{(1+\beta)^2}\,C'~.
        \label{eq:l2_convergence}
    \end{equation}

    \item[(ii)] \textbf{Terminal distribution convergence.}
    The terminal distribution $p_\beta^*$ of the GFT process
    converges to $q$ in Wasserstein-2 distance as $\beta \to 0^+$:
    \begin{equation}
        W_2^2(p_\beta^*, q)
        \leq \frac{\beta^2\,e^{2L+1}}{(1+\beta)^2}\,C' ~.
        \label{eq:w2_convergence}
    \end{equation}
    \end{enumerate}
    Therefore, as \(\beta \to 0^+\), the minimizer of the GFT objective converges to both the target vector field and target distribution. \\Proof given in Appendix \ref{ap:GFT_Convergence}.
\end{theorem}

Theorem \ref{thm:convergence} states that the minimizer of the GFT objective and its terminal distribution are bounded in distance from their corresponding targets by a function of the temperature \(\beta\). When the cooling schedule is chosen such that \(\beta \rightarrow 0^+\) at the end of fine-tuning, these bounds vanish and GFT converges to the same target vector field and terminal distribution as standard CFM. This provides a theoretical basis for our proposed annealed regularization, and is corroborated by our empirical results, which demonstrate the GFT reaches the same accuracy as standard CFM at the end of fine-tuning, while providing increased stability and sampling diversity.

The fine-tuned model \(v_\theta\) is a marginal model, and is trained to approximate the conditional target \(v_q(X_t, t| Z)\) marginalized over \(Z \sim \pi\) rather than the true conditional target itself. This introduces an inherent approximation error, even at the function-space minimizer \(v_\beta^*\). However, this does not affect the convergence guarantees of Theorem \ref{thm:convergence}, which are stated in terms of the marginal target \(v_q\) and the terminal distribution q (the correct targets for the unconditional model). The use of the marginal model is therefore a deliberate design choice of GFT, which enables the use of \(v_\theta\) as an unconditional generator at inference time. 

\vspace{-0.8em}
\paragraph{Remark on the degree of distribution shift. } The constant \(C'\) measures the \(L^2\) mismatch between the pretrained and target vector fields, and therefore captures the degree of distribution shift between \(p_1\) and \(q\). The upper bounds given in \((i)\) and \((ii)\) are both scaled proportionally to \(C'\) and \(\beta\). We can therefore conclude that the fine-tuning convergence rate is dependent on the relationship between the chosen cooling schedule and the degree of distribution shift between the pretraining and fine-tuning target distributions. This intuition is further support be our experiments across varying degrees of distribution shift. In settings with larger shifts, fine-tuning convergence required more epochs and a slower-decaying cooling schedule, while settings with smaller distribution shifts were able to reach high accuracy with a faster-decaying cooling schedule over fewer epochs.

\subsection{Extension to Few-Step Inference}

A significant consideration in the development of flow-based generative models is achieving efficient inference generation, which is severely reduced in iterative generative models. Several recent works have explored methods of modifying flow-based models to generate samples in one or few steps \cite{boffi2025buildconsistencymodellearning, geng2026mean} by defining a mapping mechanism to directly transport a given sample from time \(r\) to any time \(t\) along the learned probability density path. Such a mapping mechanism can then be used to achieve inference efficiency similar to previous one-step generative models while often retaining the accuracy of flow-based methods. For many SOC methods, the adjoint formulation inherently requires optimizing over full sampling trajectories \cite{domingo-enrich2025adjoint, havens2025adjoint}, and would therefore require a non-trivial reformulation of both the algorithm and theoretical framework to extend to such methods.

We now demonstrate the natural extension of GFT to Mean Flow (MF) \cite{geng2026mean}, a popular few-step generation method which performs mapping through a learned average velocity field \(u(X_t, r, t)\). Due to the linear relationship between the average and instantaneous velocity fields \eqref{eq:MF_identity}, it follows that the optimal average velocity field corresponding to the instantaneous optimum given in Theorem \ref{thm:opt_GFT} should also be a convex combination of two vector fields. This intuition is formalized in Proposition \ref{pr:mean_flow}.

\begin{proposition}\label{pr:mean_flow}

Let \(u_{\theta_0}\), \(u_q\), and \(u_\theta\) be average velocity fields related to their corresponding instantaneous fields \(v_{\theta_0}\), \(v_q\), and \(v_\theta\) via the MeanFlow Identity (equation (6) in \cite{geng2026mean}):

\begin{equation}
    u_\phi(X_t, r, t) = v_\phi(X_t, t) - (t-r) \frac{d}{dt}u_\phi(X_t, r, t)~, \qquad \phi = \{\theta_0, q, \theta\}~.
\label{eq:MF_identity}
\end{equation}

Assuming that the instantaneous vector field \(v_\theta\) takes the form given in \eqref{eq:GFT_optimum}, the corresponding optimal average vector field is:
\begin{equation}
    u_\theta(X_t, r, t) = \left(\frac{1}{1+\beta}\right) u_q(X_t, r, t) + \left(\frac{\beta}{1+\beta}\right) u_{\theta_0}(X_t, r, t)~.
\label{eq:GFT_avg_optimum}
\end{equation}
Proof given in Appendix \ref{ap:MeanFlow}.
\end{proposition}

Proposition \ref{pr:mean_flow} establishes the form of the average velocity field corresponding to the previously derived instantaneous optimum. Although this expression cannot be directly used during fine-tuning since the average target vector field \(u_q\) is unknown, this relationship can be used to naturally extend GFT to fine-tuning few-step inference models, as shown in Algorithm \ref{alg:GFT_MF}. Using Proposition \ref{pr:mean_flow} to lift the gradual sequence of instantaneous target vector fields created by GFT to the space of average vector fields, we retain the benefits of gradual adaptation when fine-tuning MF models without sacrificing generation quality. Extended results are shown in Appendix \ref{ap:MeanFlow}.

%% file: experiments.tex
\begin{table}[bp]
\centering
\caption{Convergence and stability analysis on the Camelyon17 dataset. GFT yields greater stability and a higher convergence rate compared to CFM. Extended results in Appendix \ref{ap:stability}.}
\label{tab:stability_main}

\makebox[\textwidth][c]{
\begin{tabular}{clccc}
\toprule
& \textbf{Objective} 
& \textbf{Inst. Variance} $\downarrow$
& \textbf{Convergence Rate} $\uparrow$
& \textbf{Spearman $\rho$} $\downarrow$ \\
\midrule

\multirow{3}{*}{Cross-Domain Adaptation} 
& Full (CFM)     & 99.242 & 3.133 $\times10^{-2}$ & -0.663 \\
& LoRA (CFM)   & 70.604 & 2.504 $\times10^{-2}$ & -0.597 \\
& Full (GFT)     & \textbf{63.132} & 3.647 $\times 10^{-2}$ & \textbf{-0.932} \\ 
& Full (CFM; RI)     & 859.632 & \textbf{7.801} $\times \textbf{10}^{\textbf{-2}}$ & -0.394 \\

\hline

\multirow{3}{*}{In-Domain Adaptation} 
& Full (CFM)    & 19.387 & 4.693 $\times10^{-3}$ & -0.253 \\
& LoRA (CFM)   & 13.134 & 1.182 $\times10^{-2}$ & -0.499 \\
& LoRA (GFT) & \textbf{8.429} & \textbf{2.213} $\times \textbf{10}^{\textbf{-2}}$ & \textbf{-0.966} \\ 

\bottomrule
\end{tabular}
}
\end{table}

We experimentally evaluate GFT by fine-tuning pretrained flow matching models and comparing the performance of GFT to standard and reward-based fine-tuning methods. To evaluate GFT under natural distribution shifts, we test it against standard fine-tuning with the CFM objective on three WILDS benchmark datasets \cite{wilds2021}. Each image in these datasets belongs to a single group of samples with common natural factors such as experimental batch, geographic region, or time of acquisition. The training and validation split of each dataset is comprised of disjoint sets of groups, creating a distribution shift between them. To ensure rigorous evaluation, we hold out a random 20\% of both the training and validation splits of each dataset for exclusive use in calculating Fréchet Inception Distance (FID). We define two distinct fine-tuning scenarios to evaluate GFT across varying degrees of distribution shift: 
\begin{enumerate}
    \item \textbf{Cross-domain adaptation.} We fine-tune a model pretrained on Cifar-10 to the training split of a WILDS dataset, and calculate FID on the held-out training samples. This represents a large shift from a general-purpose prior to a specific scientific domain.
    \item \textbf{In-domain adaptation.} Starting from models after cross-domain adaptation, we perform a second round of fine-tuning on the corresponding validation split, calculating FID on the held-out validation samples. This simulates a subtler distribution shift where adaptation is required within the same scientific context. 
\end{enumerate}

In addition to FID and average path length, we report three metrics to quantify the speed and stability of convergence. Instantaneous variance quantifies local volatility by taking the average of sliding window variance calculations relative to a radial basis function mean. Convergence rate measures adaptation speed by taking the average of the absolute slopes of linear regressions performed over sliding windows. Finally, we report Spearman correlation between training epochs and FID to assess global consistency. Additional implementation details are given in Appendix \ref{ap:imp_details}. 

To further compare the performance of GFT to reward-based fine-tuning methods, we shift to smaller scale experiments with simulated 2D data. This setting enables a fair comparison for unconditional generation without the need to incorporate additional elements to stabilize and guide the reward-based fine-tuning methods. In these experiments, a flow matching model is pretrained to generate samples from a unimodal Gaussian distribution, and then fine-tuned to a bimodal Gaussian target using GFT, as well as standard CFM fine-tuning \cite{lipman2023flow}, Adjoint Matching \cite{domingo-enrich2025adjoint}, and ORW-CFM-W2 \cite{fan2025online}. In these experiments, we report mode coverage as the ratio of generated samples that fall within a fixed radius of each target mean, and covariance error as the \(L^2\) norm of the difference between the empirical and true covariances for each mode.

\subsection{Fine-tuning for In-Domain Adaptation}\label{sc:small_ds}

\begin{figure*}
\centering
\begin{subfigure}{.33\textwidth}
  \centering
  \includegraphics[width=1.0\linewidth]{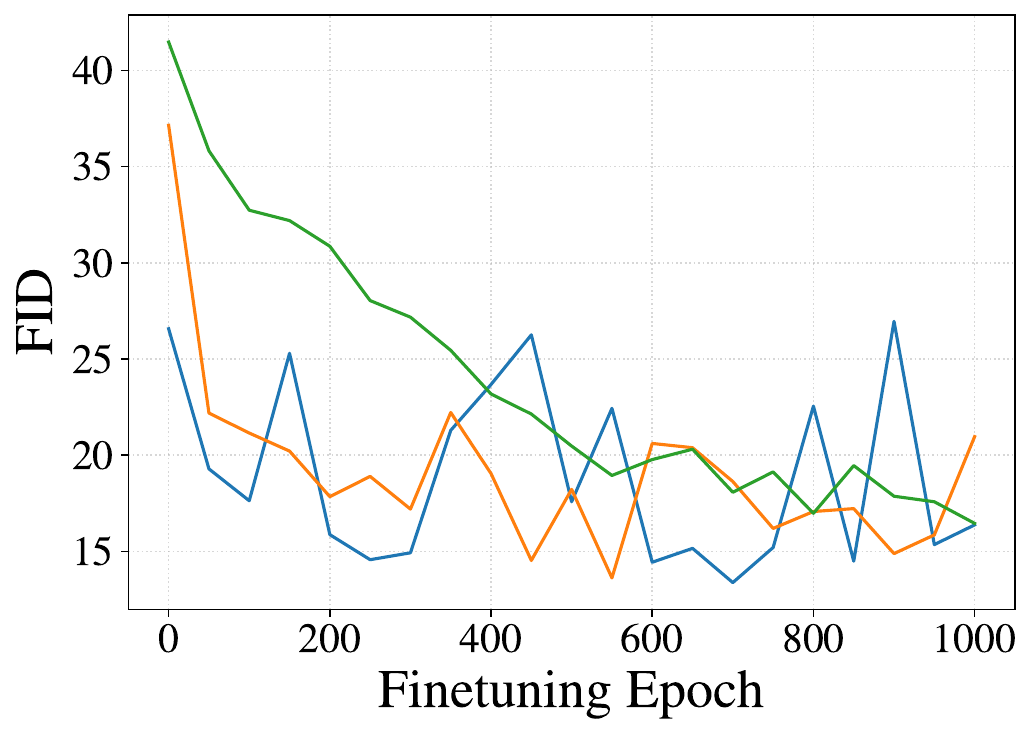}
  \label{fig:sub1}
\end{subfigure}
\hspace{0.05\textwidth}
\begin{subfigure}{.33\textwidth}
  \centering
  \includegraphics[width=1.45\linewidth]{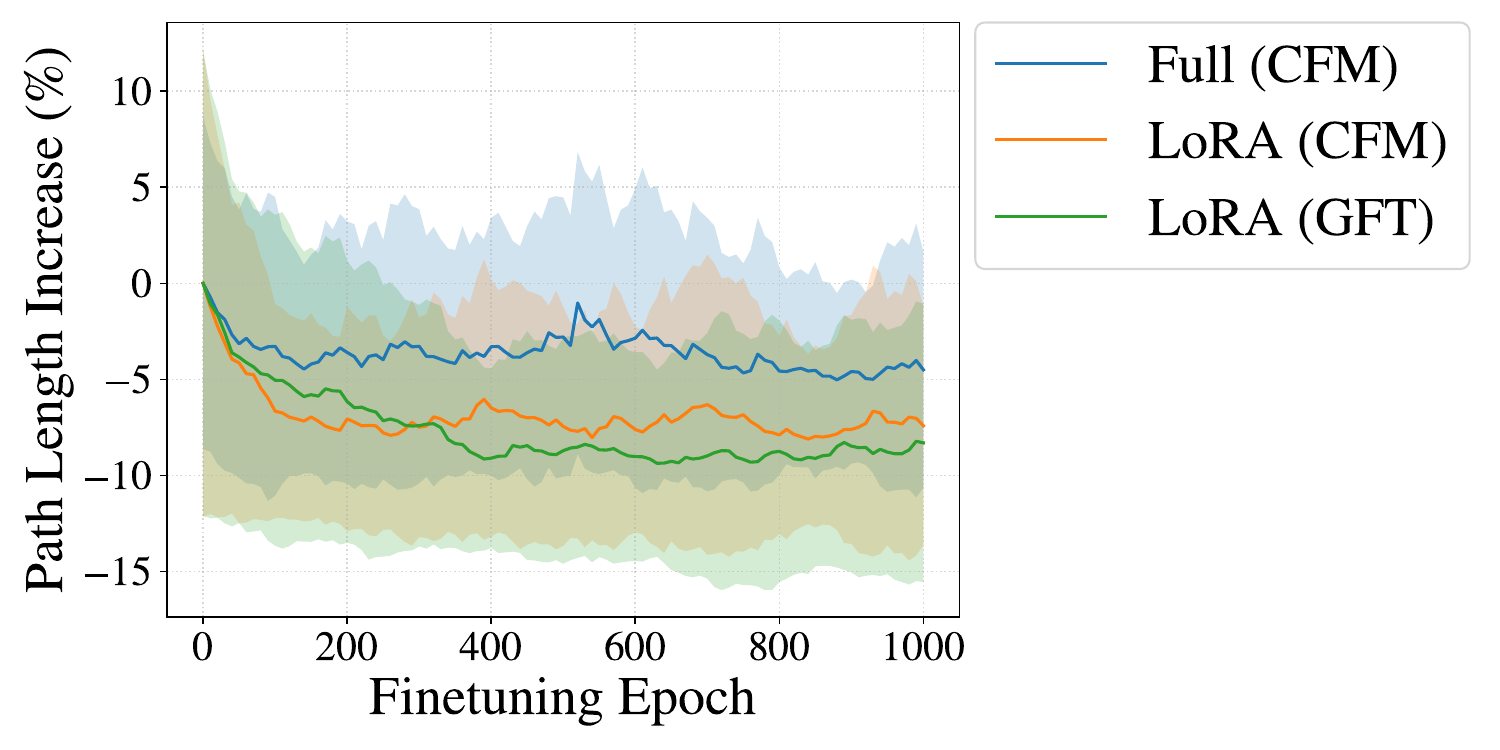}
  \label{fig:sub2}
\end{subfigure}
\caption{In-domain adaptation on Camelyon17. GFT achieves the same accuracy as CFM at the end of fine-tuning, while maintaining a lower average path length. Further results in Appendix \ref{ap:small_results}.}
\label{fig:small_shift}
\end{figure*}

\begin{figure*}
\centering
\begin{subfigure}{.33\textwidth}
  \centering
  \includegraphics[width=1.0\linewidth]{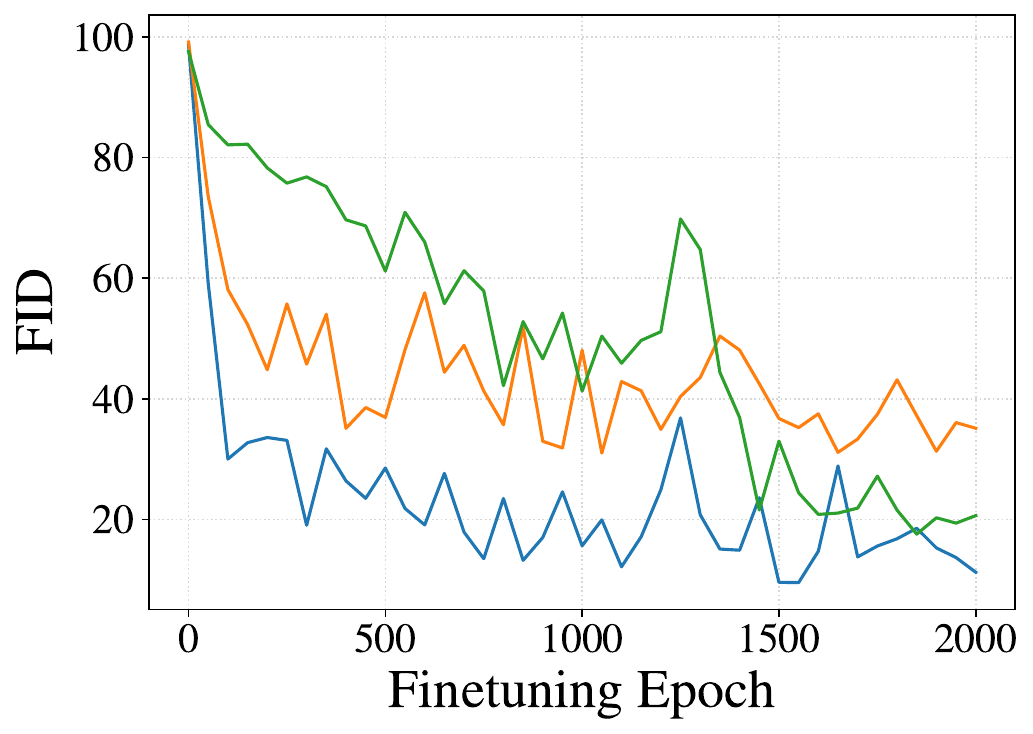}
  \label{fig:sub1}
\end{subfigure}%
\hspace{0.05\textwidth}
\begin{subfigure}{.33\textwidth}
  \centering
  \includegraphics[width=1.45\linewidth]{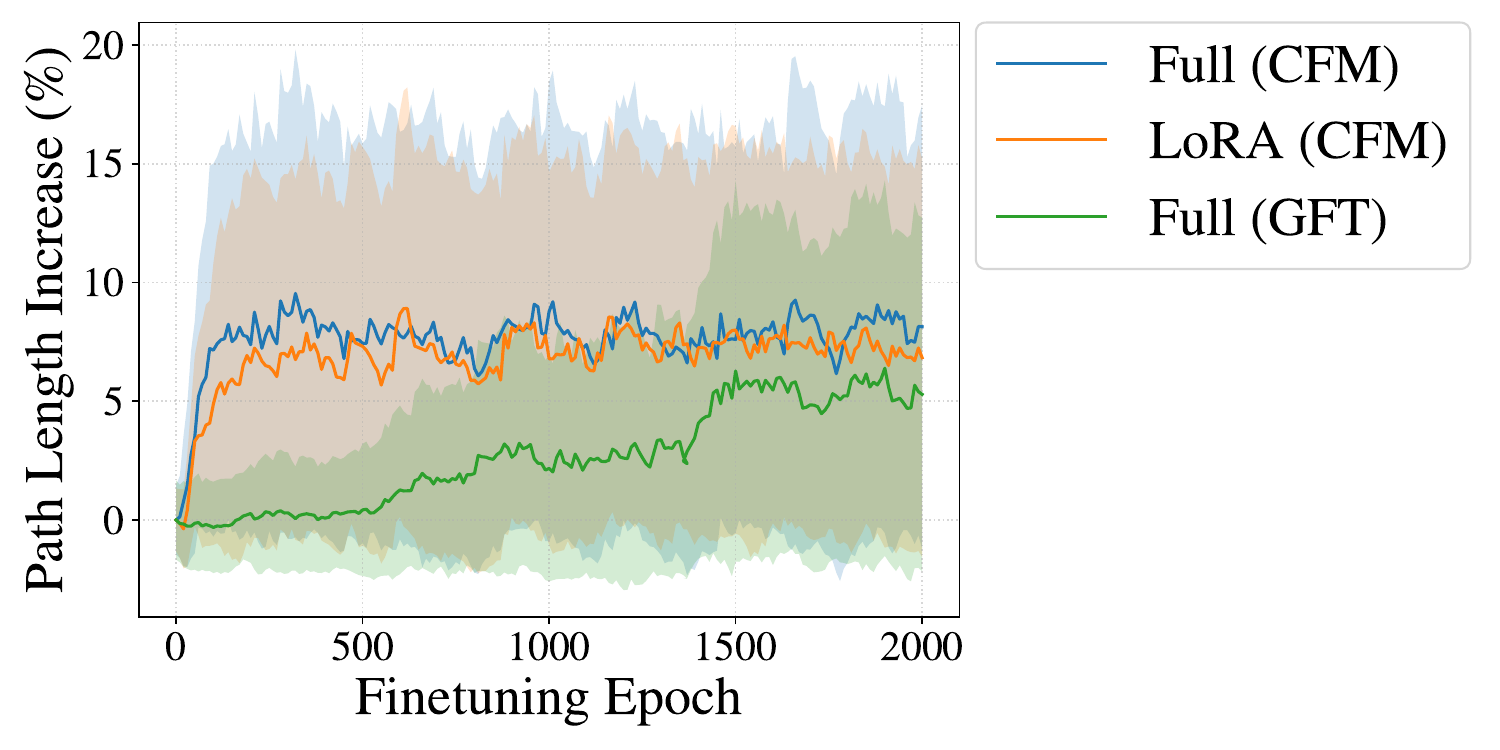}
  \label{fig:sub2}
\end{subfigure}
\caption{Cross-domain adaptation on Camelyon17. By the end of fine-tuning, GFT has achieved the same accuracy as CFM with a lower average path length. Further results in Appendix \ref{ap:large_results}.}
\label{fig:large_shift}
\end{figure*}

\begin{figure*}[bp]
\centering
\begin{subfigure}[t]{.18\textwidth}
  \centering
  \includegraphics[width=.85\linewidth]{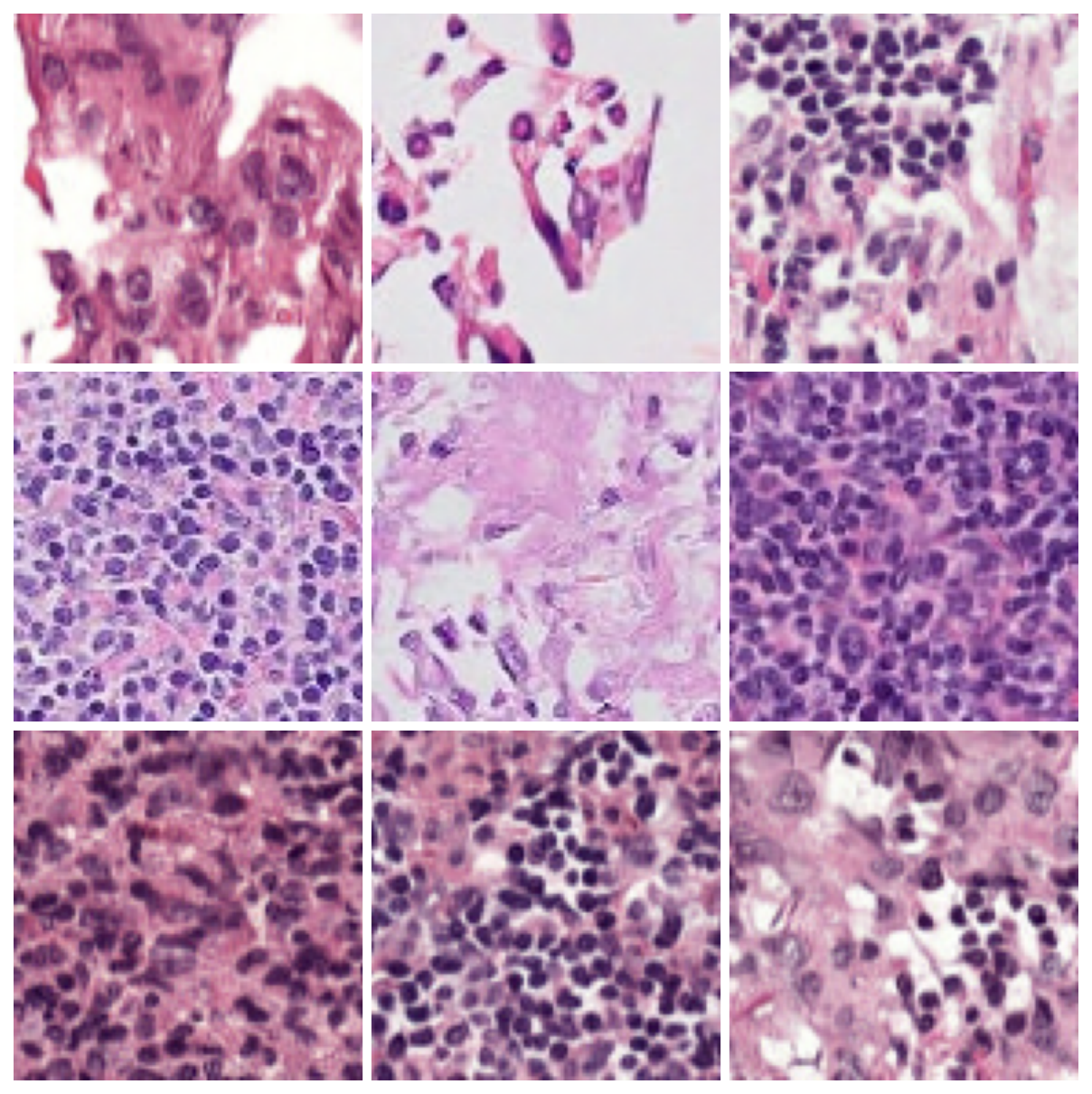}
  \caption{Full (CFM).}
  \label{fig:sub2}
\end{subfigure}\hfill
\begin{subfigure}[t]{.18\textwidth}
  \centering
  \includegraphics[width=.85\linewidth]{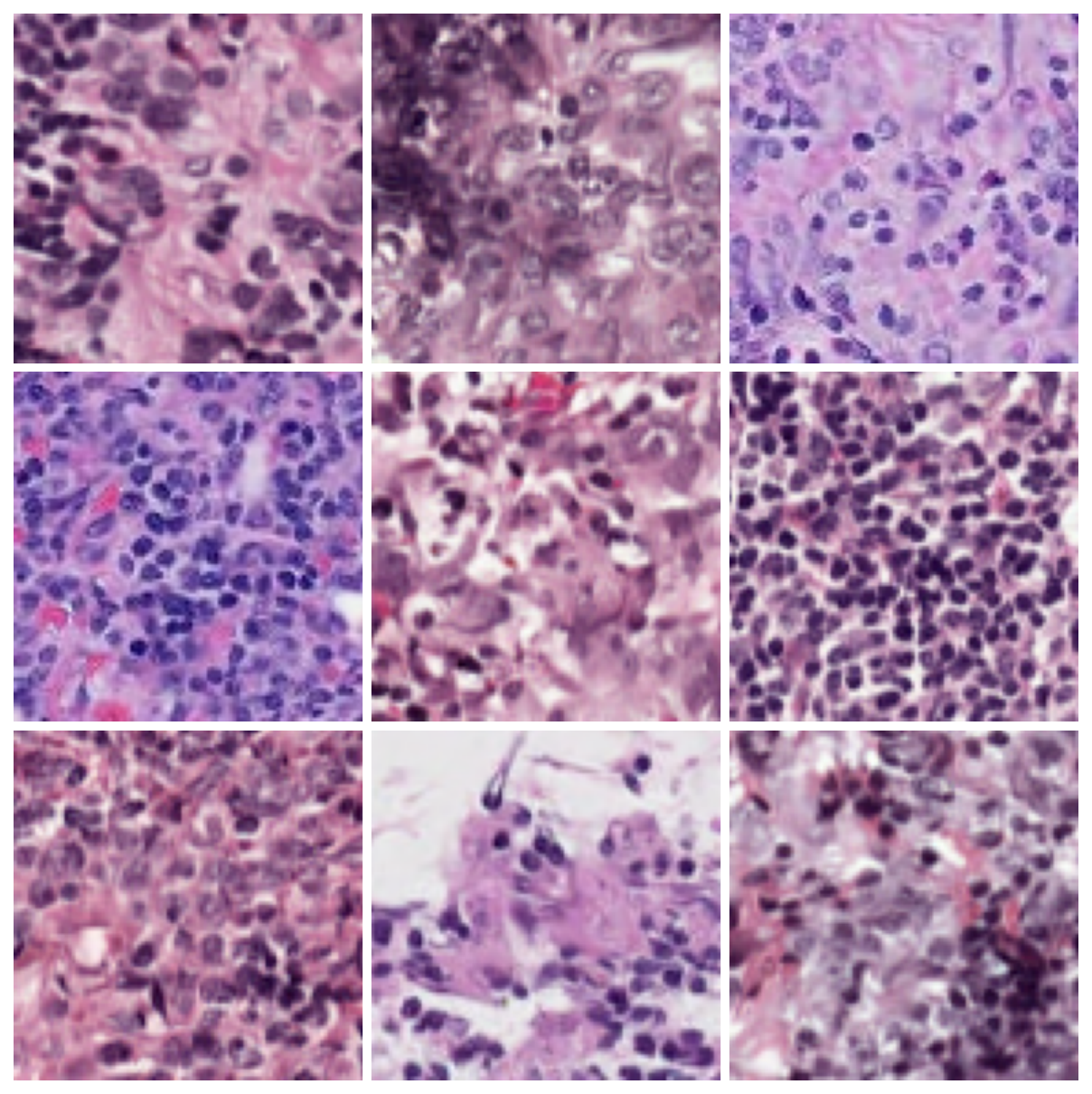}
  \caption{Full (GFT).}
  \label{fig:sub2}
\end{subfigure}\hfill
\begin{subfigure}[t]{.18\textwidth}
  \centering
  \includegraphics[width=.85\linewidth]{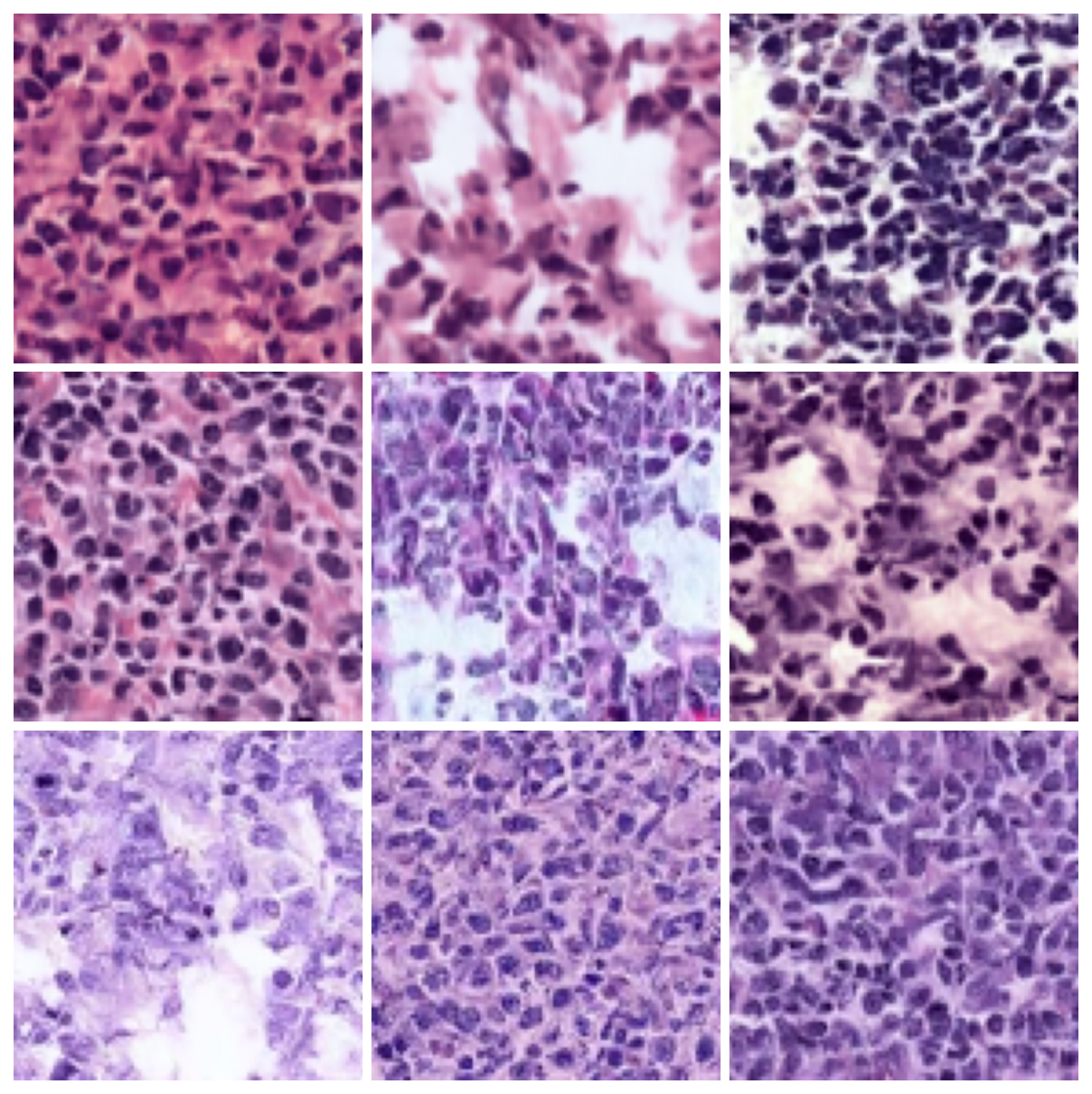}
  \caption{LoRA (CFM).}
  \label{fig:sub2}
\end{subfigure}\hfill
\begin{subfigure}[t]{.18\textwidth}
  \centering
  \includegraphics[width=.85\linewidth]{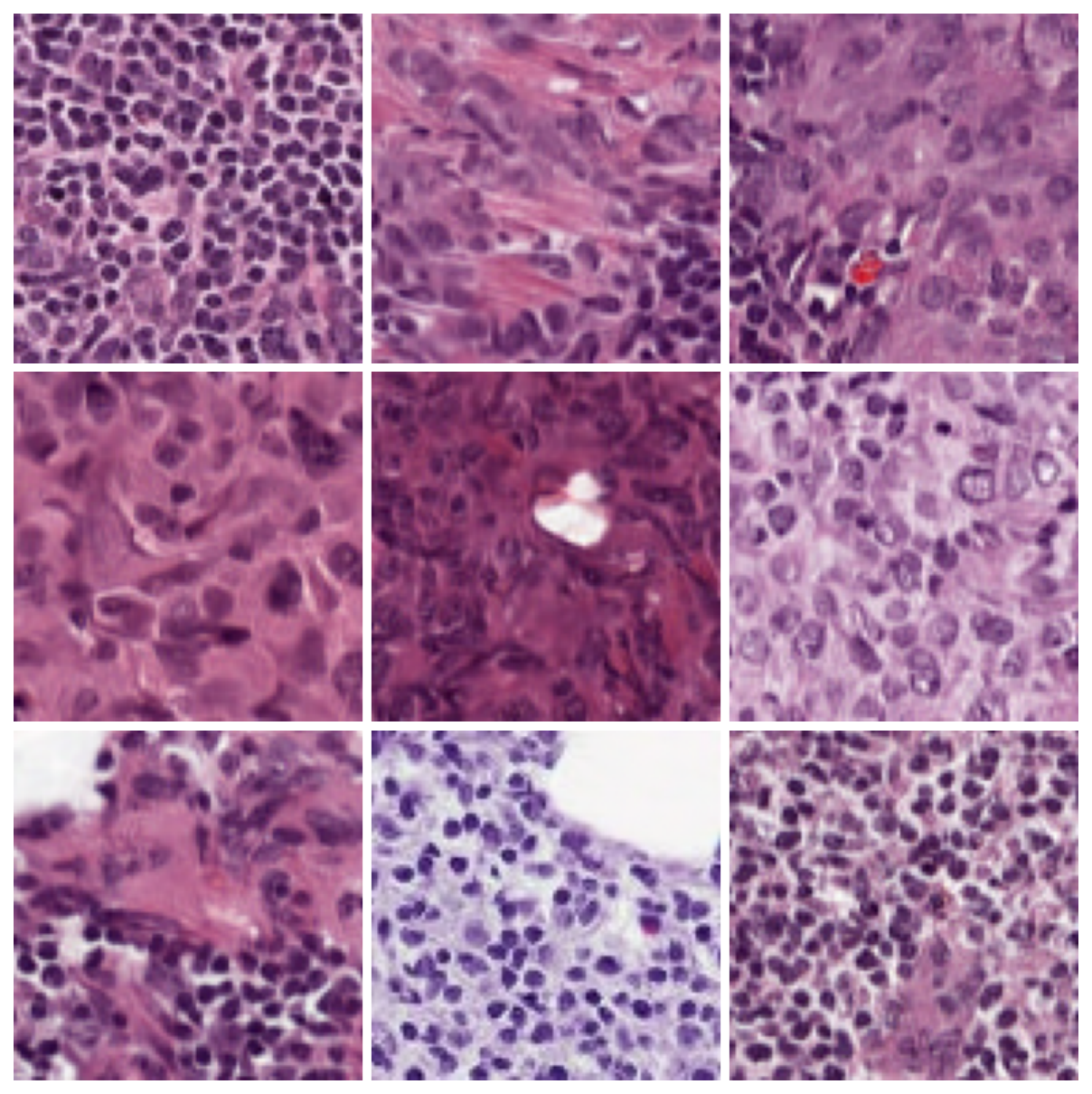}
  \caption{LoRA (GFT).}
  \label{fig:sub2}
\end{subfigure}\hfill
\begin{subfigure}[t]{.18\textwidth}
  \centering
  \includegraphics[width=.85\linewidth]{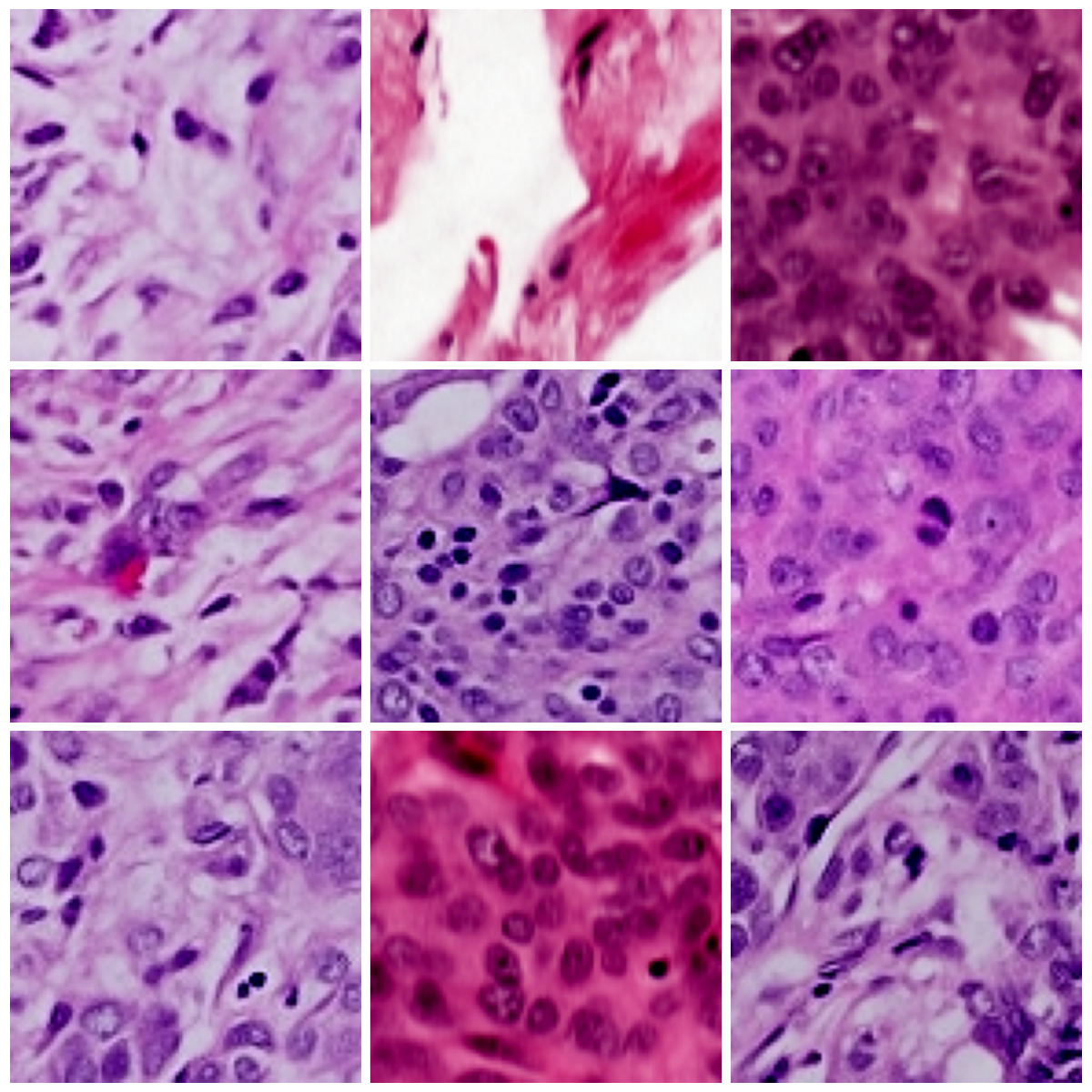}
  \caption{Full (GFT-MF).}
  \label{fig:sub2}
\end{subfigure}\hfill
\caption{Generated images of the Camelyon17 dataset with varying fine-tuning methods. Images shown in (e) are generated using two-step inference. Extended results in Appendix \ref{ap:images}.}
\label{fig:Camelyon17_images}
\end{figure*}

We begin our empirical evaluation with an investigation of fine-tuning for in-domain adaptation over 1000 epochs. For this setting, where the pretraining and fine-tuning target distribution are similar, low rank updates demonstrate superior performance. LoRA with the CFM objective matches the accuracy of full fine-tuning throughout training, and LoRA with the GFT objective reaches a similar accuracy after about 600 epochs (Figure \ref{fig:small_shift})). However, GFT converges with significantly higher stability than CFM (Table \ref{tab:stability_main}). GFT achieves the lowest instantaneous variance, indicating low oscillation throughout fine-tuning, and a Spearman correlation coefficient is close to -1, pointing to strong monotonic convergence towards optimal model parameters. In addition to high accuracy and stable convergence, GFT achieves a lower average path length than both CFM methods, supporting efficient inference-time generation.

\subsection{Fine-tuning for Cross-Domain Adaptation}\label{sc:large_ds}

We conduct a similar comparison of GFT to standard fine-tuning for cross-domain adaptation across large distribution shifts over 2000 epochs. The evolution of FID in these experiments demonstrates the key differences between LoRA and full fine-tuning in the setting of large distribution shifts, as well as the benefits of GFT (Figure \ref{fig:large_shift}). LoRA fine-tuning fails to match the accuracy of full fine-tuning, indicating that low-rank updates are too restrictive to capture the substantial differences between the pretrained and fine-tuned target distributions. Although full fine-tuning with GFT requires more epochs to reach high accuracy generation than CFM, it attains comparable FID at the end of fine-tuning, and does so with notably improved stability (Table \ref{tab:stability_main}) and higher diversity (Table \ref{tab:diversity}). This behavior is consistent with the use of the GFT gradual cooling schedule, which moderates model updates to achieve more stable optimization. In addition, GFT yields shorter average probability path lengths with lower variance (Figure \ref{fig:large_shift}), indicating better preservation of the pretrained vector field and correlating with more efficient generation at inference time. 

\paragraph{Mean Flow GFT} We perform further Cross-Domain Adaptation experiments using the MF-GFT algorithm (\ref{alg:GFT_MF}), starting from a MF model pretrained on Cifar10 and fine-tuning on the Camelyon17 dataset. Compared to standard MF fine-tuning, MF-GFT increases stability (Table \ref{tab:MF}) while reaching the same FID at the end of fine-tuning (Figure \ref{fig:mean_flow}). Notably, the MF-GFT model is almost able to match the FID of the iterative flow matching models, while achieving a 50-100x reduction in number of model evaluations. Further results are shown in Appendix \ref{ap:MeanFlow}.

\subsection{Diversity and Mode Collapse}

We compare GFT to reward-based fine-tuning methods on simulated 2D data, which enables a fair comparison of unconditional generation, without the need for the supplemental stability or guidance mechanisms often required for reward methods in higher dimensional settings. In these experiments, a flow matching model is pretrained to transport samples between two unimodal Gaussian distributions, and then fine-tuned to transport samples from the same unimodal Gaussian source to a bimodal target. 

\begin{figure*}
\centering
\includegraphics[width=0.9\linewidth]{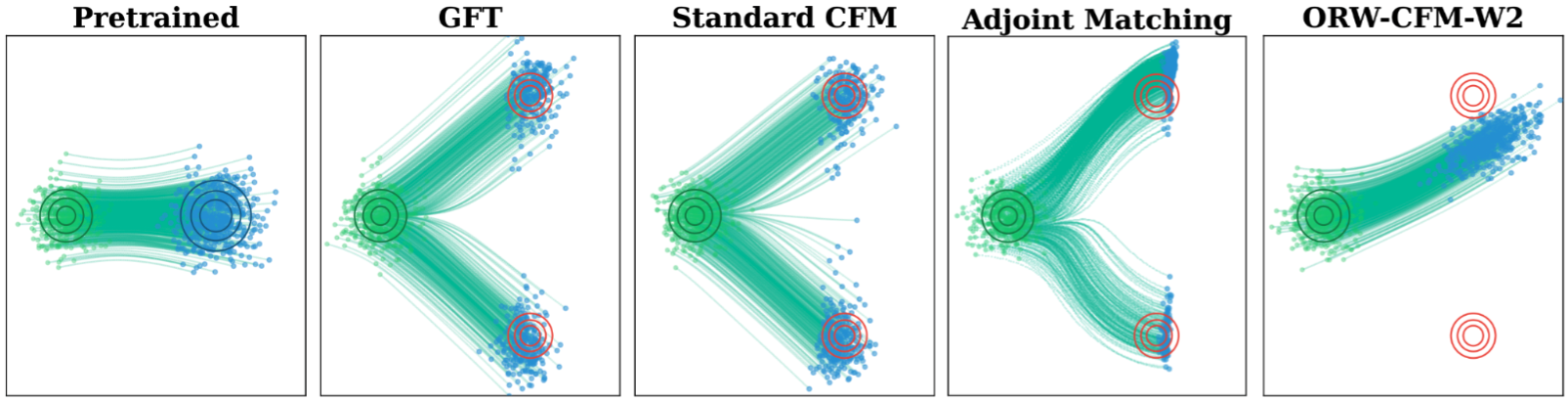}

\caption{Simulated trajectories for 2D experiments. GFT and CFM recover both target modes with high accuracy and diversity. The reward-based methods fall into varying degrees of mode collapse.}
\label{fig:2d_exp}
\end{figure*}

The resulting trajectory plots (Figure \ref{fig:2d_exp}) demonstrate the stronger performance of the sample-based fine-tuning methods over the reward methods. GFT and CFM are able to cover both fine-tuning target modes with high accuracy and diversity. Adjoint Matching suffers from significantly reduced diversity compared to the pretrained model, with all generated samples tightly clustered within the distribution. The reduced coverage and mode collapse demonstrated by Adjoint Matching in these experiments is further quantified in Table \ref{tab:2d}, where Adjoint Matching is shown to have significantly higher covariance error than both GFT and CFM. While ORW-CFM-W2 maintains a high generative diversity, it is only able to acquire one of the target distribution modes with low accuracy.

\begin{table}
\centering
\caption{Fine-tuning performance comparison for 2D experiments. GFT and CFM achieve better mode coverage and covariance accuracy than both reward-based methods. They also have significantly higher training efficiency, as reported in units of average seconds per epoch during pretraining.  }
\begin{tabular}{lccccc}
\toprule
& \multicolumn{2}{c}{\textbf{Upper Mode}} & \multicolumn{2}{c}{\textbf{Lower Mode}} \\
\cmidrule(r){2-3} \cmidrule(l){4-5}
\textbf{Method} & \textbf{Cov. Error} $\downarrow$ & \textbf{Coverage} $\uparrow$ & \textbf{Cov. Error} $\downarrow$ & \textbf{Coverage} $\uparrow$ & \textbf{Epoch Time}\\
\midrule
CFM              & 0.112 & \textbf{0.446} & \textbf{0.059} & 0.394 & 0.994\\
GFT              & \textbf{0.039} & 0.362 & 0.071 & \textbf{0.472} & 1.556\\
Adjoint Matching & 0.198 & 0.390  & 0.196 & 0.228 & 26.111\\
ORW-CFM-W2       & 0.211   & 0.042 & inf   & 0.000 & 3.722\\
\bottomrule
\end{tabular}
\label{tab:2d}
\end{table}

We further comment on the significantly increased training time required for reward-based compared to sample-based methods. Table \ref{tab:2d} reports average epoch time for each method as a unit of the average seconds per epoch during pretraining with the CFM objective. Standard CFM fine-tuning retains the same training speed, with GFT taking slightly longer due to the need for two model evaluations for each epoch. However, the epoch time of Adjoint Matching is over an order of magnitude higher than that of GFT, as it involves two expensive ODE simulations for each fine-tuning step.

\subsection{Effect of Regularization Strength}
\label{sc
:regularization}

\begin{figure*}[b]
\centering
\begin{subfigure}{.45\textwidth}
  \centering
  \includegraphics[width=0.9\linewidth]{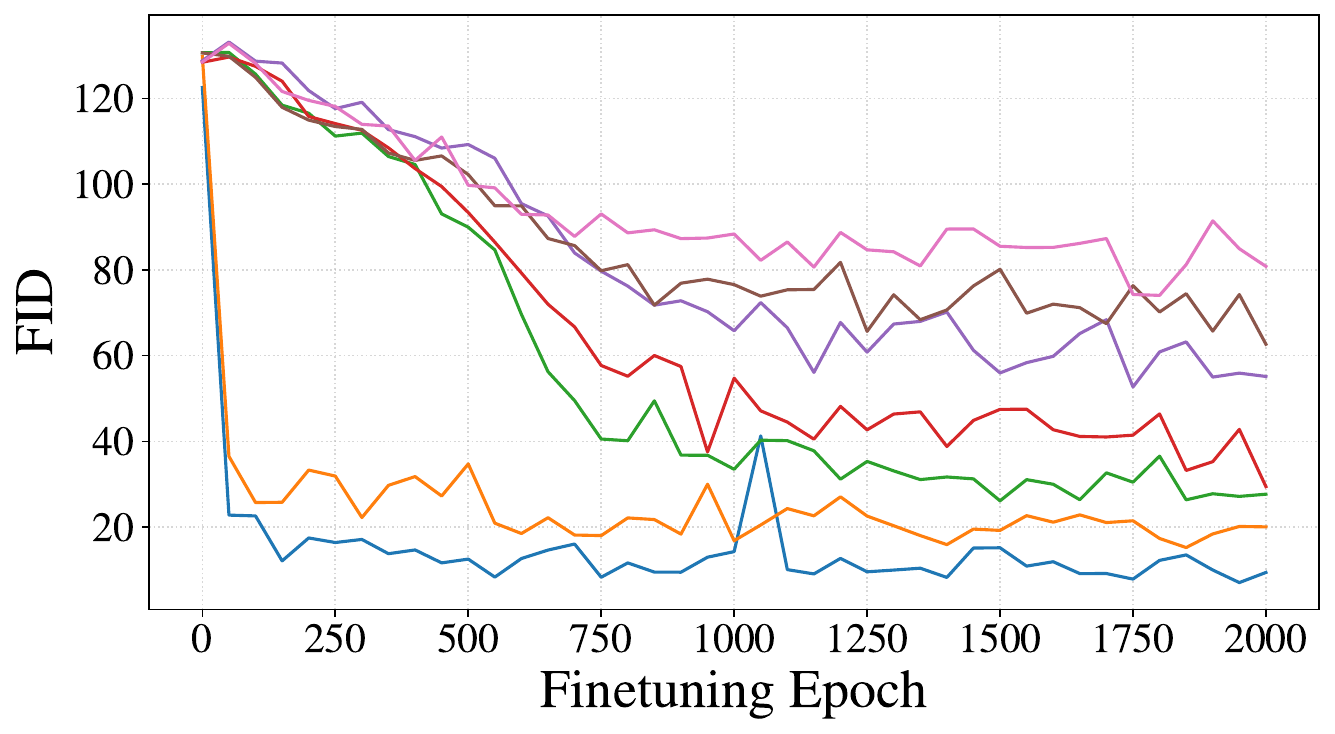}
  \caption{FID.}
  \label{fig:sub1}
\end{subfigure}%
\hspace{0.03\textwidth}
\begin{subfigure}{.45\textwidth}
  \centering
  \includegraphics[width=1.2\linewidth]{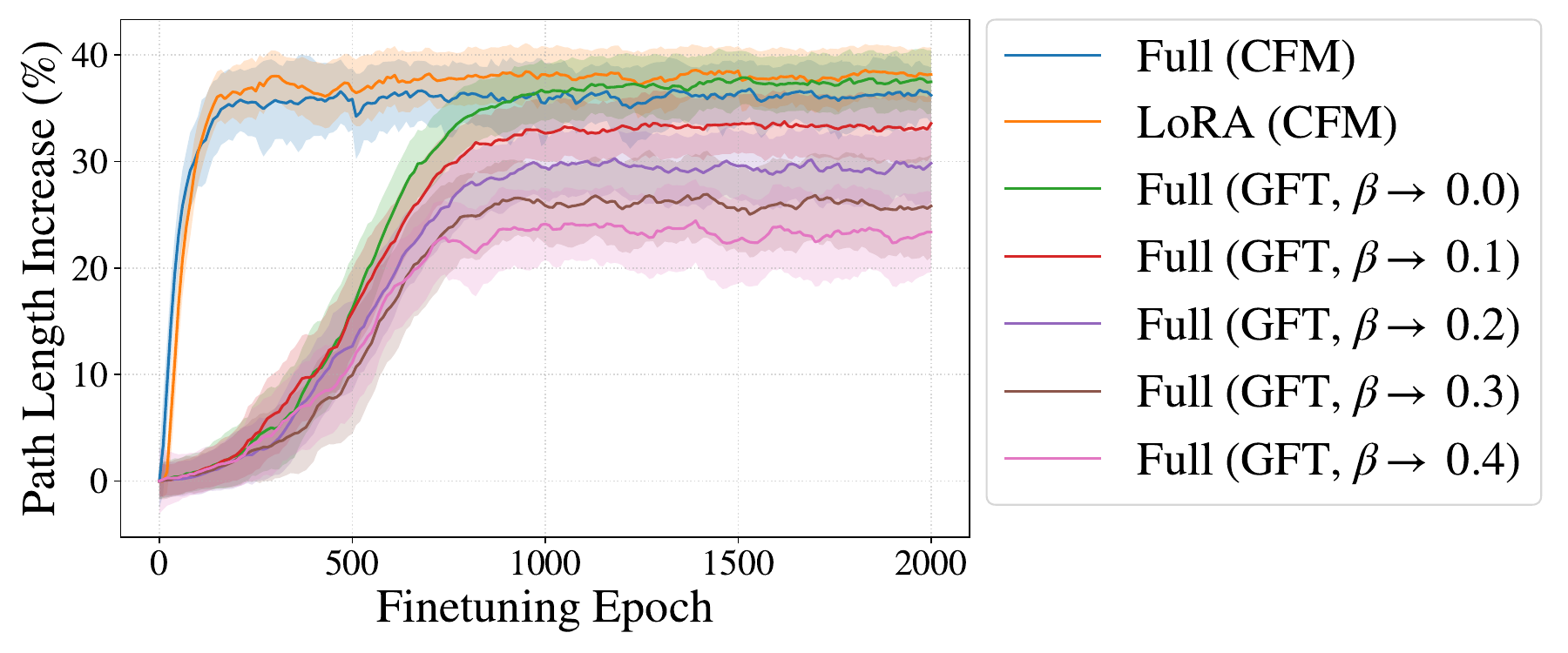}
  \caption{Path Length.}
  \label{fig:sub2}
\end{subfigure}
\caption{Cross-domain adaptation with varying terminal temperatures. A lower final path length corresponds to higher final FID, and this tradeoff can be predictably controlled by changing \(\beta\).}
\label{fig:beta_schedule_main}
\end{figure*}

To demonstrate the effect of regularization strength, we conduct a set of GFT experiments which vary only in the minimum temperature $\beta$ reached at the end of fine-tuning. This minimum value of the cooling schedule dictates the strength of regularization at the final stages of fine-tuning, and therefore determines the distribution from which the model generates images. The results of these experiments reveal an important inverse relationship between final FID and average path length, which can be reliably controlled by altering the cooling schedule (Figure \ref{fig:beta_schedule_main}). Setting the minimum temperature to zero for GFT results in a similar final path length and FID as the CFM objective. However, every uniform increase in the minimum $\beta$ value results in a highly consistent and predictable decrease in final average path length and corresponding increase in final FID. According to these results, the tradeoff between FID and path length is likely directly caused by the strength of regularization, and can therefore be explicitly controlled by varying the cooling schedule.

\subsection{Cross-Domain Adaptation vs Random Initialization}\label{sc:ft_vs_scratch}

\begin{figure*}
\centering
\begin{subfigure}{.33\textwidth}
  \centering
  \includegraphics[width=.99\linewidth]{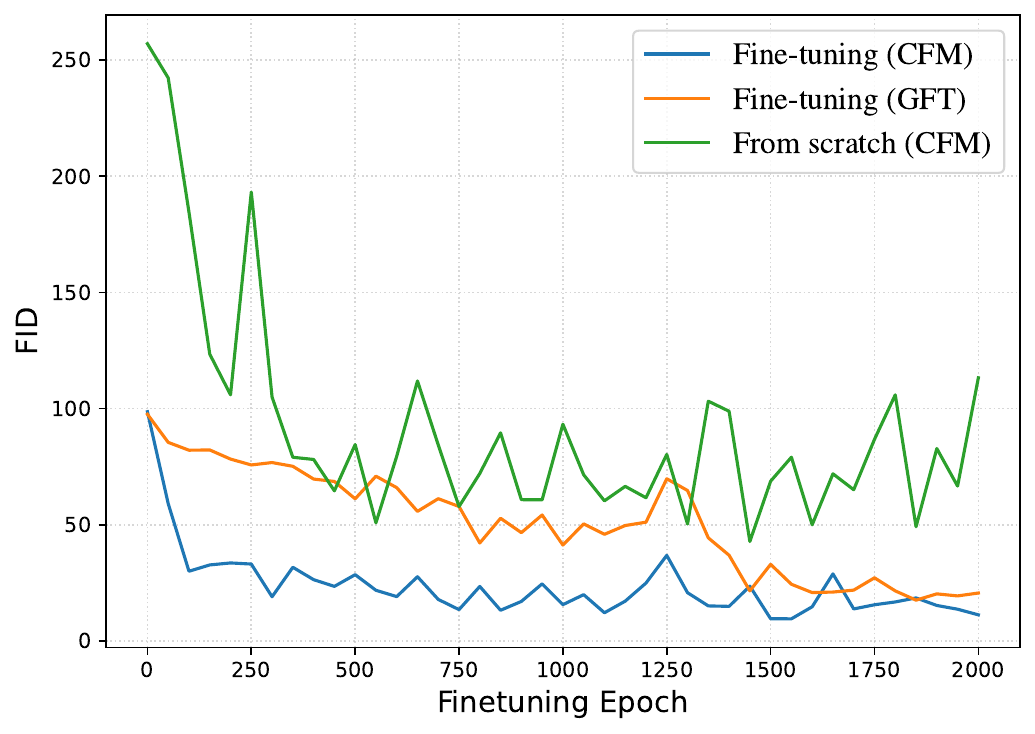}
  \caption{FID.}
  \label{fig:sub1}
\end{subfigure}%
\begin{subfigure}{.33\textwidth}
  \centering
  \includegraphics[width=.99\linewidth]{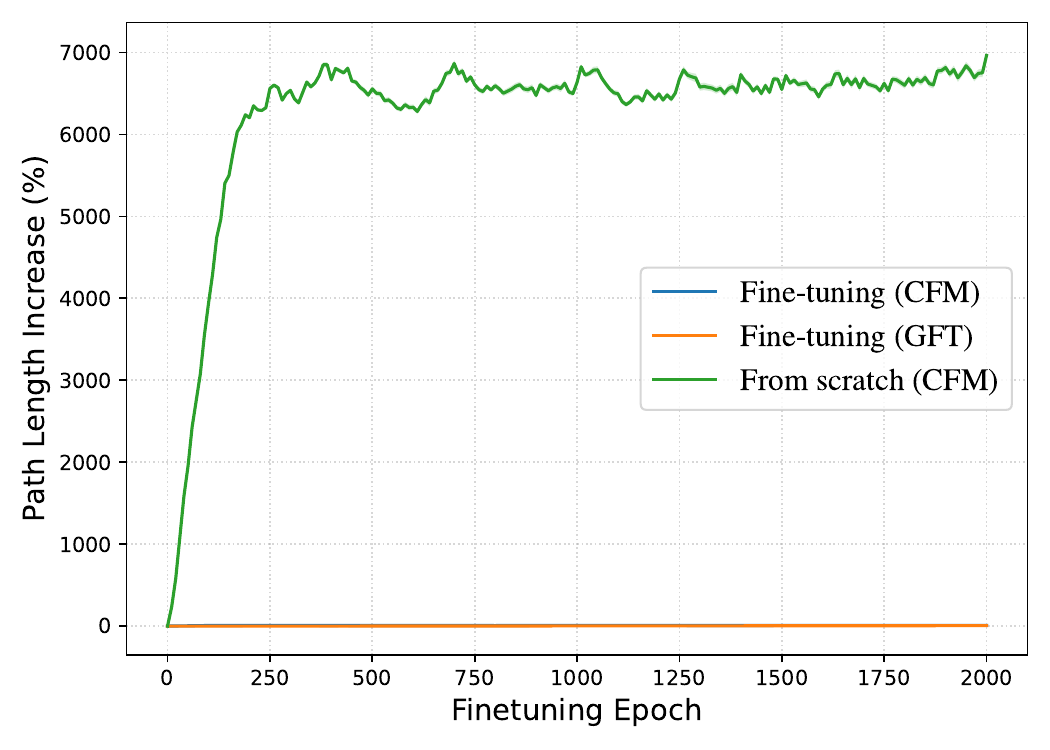}
  \caption{Path length.}
  \label{fig:sub2}
\end{subfigure}
\begin{subfigure}{.33\textwidth}
  \centering
  \includegraphics[width=.99\linewidth]{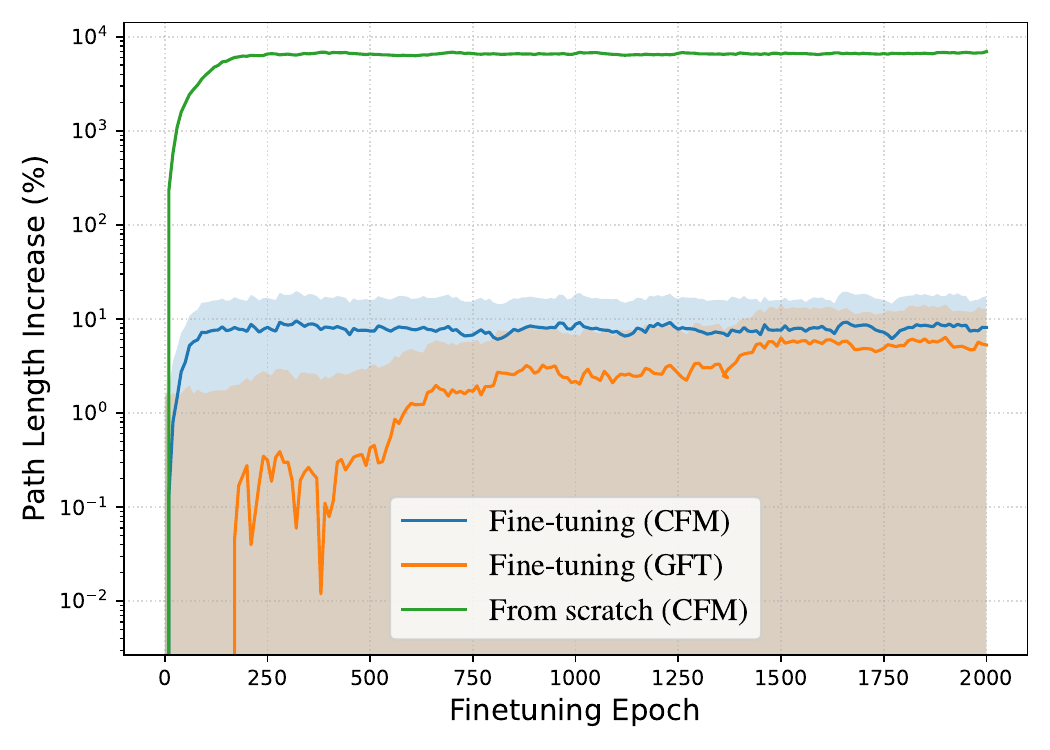}
  \caption{Path length (logarithmic scale).}
  \label{fig:sub2}
\end{subfigure}
\caption{Cross-domain adaptation results, as opposed to training from random initialization. Fine-tuning and pretraining are done on the Camelyon17 train dataset. The shaded regions of the path length graphs (b, c) represent one standard deviation from the mean.}
\label{fig:Camelyon17_from_scratch}
\end{figure*}

We perform an additional evaluation of fine-tuning performance for cross-domain adaptation by comparing the fine-tuning of a base Cifar-10 model on the Camelyon17 train dataset to training on the same dataset from random initialization (Figure \ref{fig:Camelyon17_from_scratch}).  FID analysis highlights the clear advantage of fine-tuning over training from random initialization (Figure \ref{fig:Camelyon17_from_scratch}a). Before fine-tuning has commenced, the base Cifar-10 model achieves a substantially lower FID than random initialization, despite the large distribution shift between it and Camelyon17. Fine-tuning with either of the tested objectives maintains this advantage over 2000 epochs. Despite its higher convergence rate, training from random initialization is highly unstable compared to fine-tuning, as measured by both instantaneous velocity and Spearman correlation coefficient (Table \ref{tab:stability_main}). The variation in path length also strongly points to the benefits of fine-tuning (Figures \ref{fig:Camelyon17_from_scratch}b, \ref{fig:Camelyon17_from_scratch}c). Training from random initialization results in a sharp rise in average path length, which does not decrease over the scope of the experiment. Although fine-tuning from the pretrained model also results in a slight increase in average path length, the relative scale is significantly smaller than training from scratch. This suggests that the pretrained Cifar-10 vector field provides a structural prior that remains partially valid even under large shifts.

%% file: discussion.tex
We have introduced \textbf{GFT}, a principled framework for fine-tuning flow matching generative models. GFT is derived from an optimization over the path space induced by stochastic flows, and yields a closed-form and continuous optimal drift that interpolates between pretrained and target dynamics through an annealing cooling schedule. The resulting method is intuitive and simple to implement, yet demonstrates both strong theoretical properties and empirical performance.

With the proposed GFT objective we address several critical gaps in the existing literature, and achieve all key criteria for an ideal finetuning method. GFT provably converges to the correct target distribution, with an explicit convergence rate that is directly controllable through the chosen cooling schedule. GFT exhibits highly monotonic and stable convergence compared to standard fine-tuning, as quantified by both Spearman correlation and instantaneous variance, and consistently maintains lower trajectory length. At training-time, GFT is simple to implement and highly efficient, retaining the simulation-free training of CFM. Inference efficiency is also strongly supported by GFT, which is compatible with OT source-target couplings to increase sampling efficiency, and can also be naturally extended to fine-tuning for few-step inference methods such as Mean Flow. Finally, across all experimental settings, GFT matches or outperforms alternative fine-tuning methods in generation accuracy and diversity.

By bridging the gap between high-quality pretraining and fine-tuning, GFT provides a robust solution for adapting generative models to complex, data-driven domains which require nuanced control of the outcome of fine-tuning. Our theoretical and empirical results position GFT as a competitive method for scalable adaptation of flow matching models under distribution shift.

%% file: app_tech1.tex
\section{Impact Statement}
\label{ap:impact}

This paper presents theoretical and methodological work with the primary goal of advancing the field of machine learning by improving the understanding and fine-tuning of flow-based generative models. The proposed method and accompanying analysis are general-purpose, and do not target a specific application domain. As such, the potential societal impacts of this work are indirect and depend on possible downstream use-cases, which may include both beneficial and harmful applications. We do not foresee any immediate ethical concerns directly related to this work, beyond those already broadly understood for generative models.

\section{Recovering the Score in the Stochastic View of Flow Matching\label{ap:recover_score}}

Substituting the continuity equation (\ref{eq:continuity}) into the Fokker-Planck equation (\ref{eq:FP}) yields 

\begin{equation}
    \nabla_x \cdot (p_t v_t) = \nabla_x \cdot (\tilde{v}_t p_t) - \frac{1}{2} \nabla_x \cdot \nabla_x \cdot (\sigma_t^2 I p_t).
\end{equation}

Since the diffusion coefficient \(\sigma_t^2\) depends only on the timestep \(t\), we can move it outside the divergence operators.

\begin{equation}
    \nabla_x \cdot (p_t v_t) = \nabla_x \cdot (\tilde{v}_t p_t) - \frac{1}{2} \sigma_t^2 (\nabla_x \cdot \nabla_x \cdot p_t)
\end{equation}

We rearrange terms to isolate the drift term divergence.

\begin{equation}
    \nabla_x \cdot (\tilde{v}_t p_t) = \nabla_x \cdot (p_t v_t) + \frac{1}{2} \sigma_t^2 (\nabla_x \cdot \nabla_x \cdot p_t)
\end{equation}

As the divergence operator commutes with scalar multiplication, the terms on the right hand side can be combined.

\begin{equation}
    \nabla_x \cdot (\tilde{v}_t p_t) = \nabla_x \cdot \left[ p_t v_t + \frac{1}{2} \sigma_t^2\nabla_x  p_t \right]
\end{equation}

Two vector fields have equivalent divergence when they differ only by a divergence-free component \(Z_t\), such that \(\nabla_x \cdot (p_t Z_t) = 0\). Therefore, this equivalence can be rewritten without the leading divergence operator.

\begin{equation}
    \tilde{v}_t p_t =  p_t v_t + \frac{1}{2} \sigma_t^2\nabla_x  p_t + p_t Z_t
\end{equation}

Finally, we divide by \(p_t\), using the identity that \(\frac{\nabla_x p_t}{p_t} = \nabla \log p_t\).

\begin{equation}
    \tilde{v}_t =  v_t + \frac{1}{2} \sigma_t^2\nabla_x \log p_t + Z_t
\end{equation}

\(Z_t\) is a divergence-free velocity that can simply be set to 0 to yields the final expression.

\section{Derivation of Path Length by KL Divergence}\label{ap:PL_KL}

We derive a distance measure between the probability distribution of the paths generated by two SDEs using Girsanov's Theorem. Let \(\mathbb{P}_f\) and \(\mathbb{P}_g\) be the path measures of two SDEs on the time interval \([0,1]\), with initial samples from the same source distribution. 

\begin{equation}
    \mathbb{P}_f: \quad dX_t = f_t(X_t, t) dt + \sigma_tdB_t, \qquad X_0 \sim p_0
\end{equation}
    
\begin{equation}
    \mathbb{P}_g: \quad dX_t = g_t(X_t, t) dt + \sigma_tdB_t, \qquad X_0 \sim p_0
\end{equation}

We assume that \(\mathbb{P}_f\) and \(\mathbb{P}_g\) are mutually absolutely continuous, and that the diffusion coefficient \(\sigma_t\) is non-zero. The KL-divergence between the probability measures of these processes is

\begin{equation}
    KL(\mathbb{P}_f \| \mathbb{P}_g) = \mathbb{E}_{\mathbb{P}_f} \left[ \log \frac{d \mathbb{P}_f}{d \mathbb{P}_g} \right] = - \mathbb{E}_{\mathbb{P}_f} \left[ \log \frac{d \mathbb{P}_g}{d \mathbb{P}_f} \right].
\end{equation}

Girsanov's Theorem provides the formulation for the Radon-Nikodym derivative \(\frac{d \mathbb{P}_g}{d \mathbb{P}_f}\), which specifies how the probability measure \(\mathbb{P}_g\) changes with respect to \(\mathbb{P}_f\). 

\begin{equation}
    \frac{d \mathbb{P}_g}{d \mathbb{P}_f} = \exp \left(  \int_0^1 \sigma_t^{-1} (g_t - f_t)^\intercal dB_t - \frac{1}{2} \int_0^1 \| \sigma_t^{-1}(g_t - f_t) \|^2 dt  \right)
\end{equation}

\begin{equation}
    \log \left( \frac{d \mathbb{P}_g}{d \mathbb{P}_f} \right) = \int_0^1 \sigma_t^{-1} (g_t - f_t)^\intercal dB_t - \frac{1}{2} \int_0^1 \| \sigma_t^{-1} (g_t - f_t) \|^2 dt
\end{equation}

We now take the expectation with respect to the target process.

\begin{equation}
    KL(\mathbb{P}_f \| \mathbb{P}_g) = -
\mathbb{E}_{\mathbb{P}_f} \left[ \int_0^1 \sigma_t^{-1} (g_t - f_t)^\intercal dB_t \right] + \mathbb{E}_{\mathbb{P}_f} \left[ \frac{1}{2} \int_0^1 \| \sigma_t^{-1} (g_t - f_t) \|^2 dt \right]
\end{equation}

Note that since the integral in the first term is with respect to a Brownian motion, this term is a martingale under the measure \(\mathbb{P}_f\),  and has an expected value of 0. Therefore, our final expression for Proposition \ref{pr:KL_Girsanov} is 

\begin{equation}
    KL(\mathbb{P}_f \| \mathbb{P}_g) = \mathbb{E}_{ \mathbb{P}_f} \left[ \frac{1}{2} \int_0^1 \| \sigma_t^{-1} (f_t - g_t) \|^2 dt \right].
\end{equation}

\section{Relationship Between Path Length and Fisher Information}\label{ap:Fisher}

Consider the practical pretrained and fine-tuned processes

\begin{equation}
    \mathbb{P}_{\theta_0}: \quad dX_t = v_{\theta_0}(X_t, t) dt + \sigma_tdB_t, \qquad X_0 \sim p_0
\end{equation}
    
\begin{equation}
    \mathbb{P}_\theta: \quad dX_t = v_\theta(X_t, t) dt + \sigma_tdB_t, \qquad X_0 \sim p_0
\end{equation}

Applying Proposition \ref{pr:KL_Girsanov} to these processes gives us a measure of the divergence between the probability distribution of their generated paths. In this section we set \(\sigma_t = 1\) for simplicity.

\begin{equation}
    KL(\mathbb{P}_\theta \| \mathbb{P}_{\theta_0}) = \mathbb{E}_{ \mathbb{P}_\theta} \left[ \frac{1}{2} \int_0^1 \| (v_\theta - v_{\theta_0}) \|^2 dt \right]
\end{equation}

With fine-tuning methods such as LoRA and head adaptation, we can assume that \(\mathbb{P}_\theta \approx \mathbb{P}_{\theta_0}\) throughout the fine-tuning process, and we can further simplify this expression using first-order Taylor approximation. 

\begin{equation}
    v_\theta - v_{\theta_0} \approx \mathbf{J}_t(X_t, t) \cdot (\theta - \theta_0)
\end{equation}

Here, \(\mathbf{J}_t(X_t, t) = \nabla_\theta v_\theta(X_t, t)  \big\rvert_{\theta_0} \in \mathbb{R}^{D \times P}\) is the Jacobian matrix of the vector field \(v\) with respect to parameters \(\theta\), evaluated at \(\theta_0\). Note that when fine-tuning with LoRA or head adaptation, \(\Delta \theta = \theta - \theta_0\), which is the guidance of the fine-tuning process. Substituting this back into the \(KL\) expression yields

\begin{equation}
    KL(\mathbb{P}_\theta \| \mathbb{P}_{\theta_0}) \approx \mathbb{E}_{ \mathbb{P}_{\theta}} \left[ \frac{1}{2} \int_0^1 \| \mathbf{J}_t \Delta \theta\|^2 dt \right]
\end{equation}

\begin{equation}
    = \mathbb{E}_{ \mathbb{P}_{\theta}} \left[ \frac{1}{2} \int_0^1 (\Delta \theta^\intercal \mathbf{J}_t^\intercal \mathbf{J}_t \Delta \theta)   dt \right].
\end{equation}

Note that the guidance \(\Delta \theta\) is not dependent on \(x_t\) or \(t\).

\begin{equation}
    =  \frac{1}{2} \Delta \theta^\intercal \left( \int_0^1 \mathbb{E}_{ \mathbb{P}_{\theta}} \left[\mathbf{J}_t^\intercal \mathbf{J}_t \right] dt \right) \Delta \theta 
\end{equation}

Notice that the integral term is related to the Fisher Information matrix centered at \(\theta_0\), \(\mathbf{F}_{\theta_0} \in \mathbb{R}^{P \times P}\). 

\begin{equation}
    \mathbf{F}_{\theta_0} = \int_0^1 \mathbb{E}_{ \mathbb{P}_{\theta}} [\mathbf{J}_t^\intercal \mathbf{J}_t] dt
\end{equation}

This is the time-integrated expected Fisher Information matrix over the time range \([0,1]\) for the path measure \(\mathbb{P}_\theta\). For practical implementation, we can rewrite the integral as an expectation over a uniformly sampled time \(\tau \sim \mathcal{U}[0,1]\).

\begin{equation}
    \mathbf{F}_{\theta_0} = \mathbb{E}_{\tau \sim \mathcal{U}[0,1], X_\tau \sim \mathbb{P}_\theta}[\mathbf{J}_\tau^\intercal \mathbf{J}_\tau ]
\end{equation}

Therefore, the final simplified approximated divergence between the probability distribution of the paths generated by the base and fine-tuned models is 

\begin{align}
    KL(\mathbb{P}_\theta \| \mathbb{P}_{\theta_0}) \approx \frac{1}{2} \Delta \theta^\intercal \mathbf{F}_{\theta_0}\Delta \theta. \label{eq:KL_Fisher}
\end{align}

\section{Minimizing the Gradual Fine-Tuning Objective}\label{ap:min_GFT}

The GFT objective function is defined as

\begin{equation}
    \mathcal{L}(\theta) = KL(\mathbb{P}_\theta \| \mathbb{P}_q) + \beta KL(\mathbb{P}_\theta \| \mathbb{P}_{\theta_0})
\end{equation}

Applying Proposition \ref{pr:KL_Girsanov} to both terms with \(\sigma_t = 1\) yields

\begin{equation}
    \mathcal{L}(\theta) = \mathbb{E}_{ \mathbb{P}_\theta} \left[ \frac{1}{2} \int_0^1 \| v_\theta - v_q \|^2 dt \right] + \beta \cdot \mathbb{E}_{ \mathbb{P}_\theta} \left[ \frac{1}{2} \int_0^1 \| v_\theta - v_{\theta_0} \|^2 dt \right]
\end{equation}

\begin{equation}
    = \mathbb{E}_{ \mathbb{P}_\theta} \left[ \frac{1}{2} \int_0^1 \| v_\theta - v_q \|^2 dt + \frac{\beta}{2} \int_0^1 \| v_\theta - v_{\theta_0} \|^2 dt \right]
\end{equation}

\begin{equation}
    = \mathbb{E}_{ \mathbb{P}_\theta} \left[  \int_0^1 \frac{1}{2}\| v_\theta - v_q \|^2  + \frac{\beta}{2} \| v_\theta - v_{\theta_0} \|^2 dt \right]
\end{equation}

Minimizing this function over \(\mathbb{P}_\theta\) is equivalent to finding the optimal drift \(v_\theta^*\). We do this by minimizing the expression inside the integral. 

\begin{equation}
    C = \frac{1}{2}\| v_\theta - v_q \|^2  + \frac{\beta}{2} \| v_\theta - v_{\theta_0} \|^2
\end{equation}

\begin{equation}
    \frac{dC}{dv_\theta} = (v_\theta - v_q) + \beta (v_\theta - v_{\theta_0}) = 0
\end{equation}

\begin{equation}
    (1+\beta) v_\theta - v_q - \beta v_{\theta_0} = 0
\end{equation}

\begin{equation}
    v_\theta^*(X_t, t) = \left( \frac{1}{1 + \beta} \right) v_q(X_t, t) + \left( \frac{\beta}{1 + \beta} \right) v_{\theta_0}(X_t, t) \label{eq:PM_optimal}
\end{equation}

While the expectation in the objective function depends on \(v_\theta\) through the path measure \(\mathbb{P}_\theta\), the pointwise minimizer of the integrand remains consistent. Since the optimizer \(v_\theta^*\) is a convex combination of continuous vector fields, the resulting optimal path measure \(\mathbb{P}_\theta^*\) is well-defined and achieves the lower bound of the objective.

\section{Gradient Equivalence between Total and OT-Conditional KL Divergence}\label{ap:Grad_equivalence}

Let \(\pi \in \Pi(p_0, q)\) be any coupling between the source and target distributions that admits the marginals \(\int \pi(X_0, X_1)dX_1 = p_0(X_0)\) and \(\int \pi(X_0, X_1)dX_0 = q(X_1)\). For each drawn pair \((X_0, X_1) \sim \pi\), we define the conditional probability path \(p_t(X_t|X_0, X_1)\) with the corresponding conditional vector field \(v_t(X_t|X_0, X_1)\), which must satisfy the continuity equation.

\begin{equation}
    \partial_t p_t(X_t|X_0, X_1) + \nabla_{X_t} \cdot (p_t(X_t|X_0, X_1) v_t(X_t|X_0, X_1)) = 0
\end{equation}

The unconditional path measure \(\mathbb{P}_\theta\) is then obtained by marginalizing over \(\pi\).

\begin{equation}
    p_t(X_t) = \int p_t(X_t|X_0, X_1) d \pi(X_0, X_1)
\end{equation}

\begin{equation}
    v_t(X_t) = \frac{\int v_t(X_t|X_0, X_1) p_t(X_t|X_0, X_1)d\pi(X_0, X_1)}{p_t(X_t)}
\end{equation}

We use this to define the conditional GFT objective for an arbitrary coupling \(\pi\).

\begin{equation}
    \mathcal{L}_\pi(\theta) = \mathbb{E}_{(X_0, X_1) \sim \pi} \left[ KL(\mathbb{P}_\theta(\cdot | X_0, X_1)\|\mathbb{P}_q(\cdot | X_0, X_1)) + \beta KL(\mathbb{P}_\theta(\cdot | X_0, X_1)\|\mathbb{P}_{\theta_0}(\cdot | X_0, X_1))   \right]
\end{equation}

Here, \(\mathbb{P}_\theta\) denotes the path measure induced by the conditional SDE:

\begin{equation}
    dX_t = v_\theta(X_t, t | X_0, X_1) dt + \sigma_t dB_t.
\end{equation}

The conditional path measure \(\mathbb{P}_q\) is defined similarly using the unknown conditional target vector field \(v_q(X_t, t|X_0, X_1)\). Note that despite this new conditioning, the base process \(\mathbb{P}_{\theta_0}\) remains fixed. We apply Proposition \ref{pr:KL_Girsanov} to the conditional GFT objective to write it in terms of these conditional vector fields.

\begin{multline}
    \mathcal{L_\pi(\theta)} = \mathbb{E}_{(X_0, X_1) \sim \pi} \mathbb{E}_{X_{[0,1]} \sim \mathbb{P}_\theta(\cdot | X_0, X_1)} \bigg[   \frac{1}{2} \int_0^1 \| v_\theta(X_t, t|X_0, X_1) - v_q(X_t, t|X_0, X_1) \|^2 dt \\ +  \frac{\beta}{2} \int_0^1 \| v_\theta(X_t, t|X_0, X_1) - v_{\theta_0}(X_t, t|X_0, X_1) \|^2 dt  \bigg]
\end{multline}

\begin{multline}
    = \mathbb{E}_{(X_0, X_1) \sim \pi} \mathbb{E}_{X_{[0,1]} \sim \mathbb{P}_\theta(\cdot | X_0, X_1)} \bigg[ \int_0^1 \frac{1}{2}  \bigg[  \| v_\theta(X_t, t|X_0, X_1) - v_q(X_t, t|X_0, X_1) \|^2 dt \\ + \beta \| v_\theta(X_t, t|X_0, X_1) - v_{\theta_0}(X_t, t|X_0, X_1) \|^2  \bigg] dt \bigg]
\end{multline}

Following the same logic used to minimize the unconditional objective (Appendix \ref{ap:min_GFT}) the integrand is minimized by the following conditional drift pointwise for each \((X_0, X_1)\), yielding the conditional minimizer.

\begin{equation}
    v_\theta^*(X_t, t|X_0, X_1) = \left( \frac{1}{1+\beta} \right)  v_q(X_t, t|X_0, X_1) + \left( \frac{\beta}{1 + \beta} \right)  v_{\theta_0}(X_t, t|X_0, X_1)
\end{equation}

Note that when the unconditional vector fields \(v_q\) and \(v_{\theta_0}\) are defined as the marginalization of their conditional counterparts over the same coupling \(\pi\), the unconditional optimum is precisely the marginalization of the conditional optimum. However, this optimum is still not usable for training, as it is written in terms of the unknown target vector field \(v_q\). We can solve this intractable dependency by following CFM \cite{lipman2023flow}, and replacing \(v_q\) with a chosen conditional vector field which marginalizes to the true target:

\begin{equation}
    \mathbb{E}_{(X_0, X_1) \sim \pi}[v_q(X_t, t|Z)] = v_q(X_t, t).
\end{equation}

Crucially, the fine-tuned model \(v_\theta\) is unconditional, and is trained to approximate the expected marginal optimal vector field

\begin{equation}
    v_\theta(X_t, t) \approx \mathbb{E}_{(X_0, X_1) \sim \pi}[v_\theta^*(X_t, t|X_0, X_1)].
\end{equation}

Applying Proposition \ref{pr:KL_Girsanov} and substituting the chosen tractable conditional target vector field, we get the final GFT loss:

\begin{equation}
    \mathcal{L}_\pi(\theta) = \mathbb{E}_{(X_0, X_1) \sim \pi} \mathbb{E}_{X_t \sim \mathbb{P}(\cdot | X_0, X_1)} \bigg[  \frac{1}{2}  \int_0^1 ( \|v_\theta(X_t, t) - v_q(X_t, t|X_0, X_1)\|^2 + \beta \|v_\theta(X_t, t) - v_{\theta_0}(X_t, t|X_0, X_1) \|^2 )dt \bigg] ~.
\end{equation}

As a result, the gradient of the conditional objective \(\mathcal{L}_\pi\) with respect to the model parameters \(\theta\) will coincide with the gradient of the original unconditional objective \(\mathcal{L}\) when the model is parameterized to satisfy the marginalization property. This follows from the same arguments used to justify the original CFM objective \cite{lipman2023flow}. Differentiating with respect to the model parameters \(\theta\) gives

\begin{multline}
    \nabla_\theta \mathcal{L}_\pi(\theta) = \mathbb{E}_{(X_0, X_1) \sim \pi} \mathbb{E}_{X_{[0,1]} \sim \mathbb{P}(\cdot | X_0, X_1)} \bigg[ \int_0^1 \langle (v_\theta(X_t, t) - v_q(X_t, t|X_0, X_1)) +  \\\beta (v_\theta(X_t, t) - v_{\theta_0}(X_t, t|X_0, X_1)), \nabla_\theta v_\theta (X_t, t) \rangle dt \bigg].
\end{multline}

By the construction of the conditional vector fields, for \(\phi = \theta, \theta_0, q\), we have

\begin{equation}
    v_\phi(X_t, t) = \mathbb{E}_{(X_0, X_1) \sim \pi} \left[ v_\phi(X_t, t|X_0, X_1)  | X_t\right].
\end{equation}

Using these identities and applying the tower property of conditional expectations, 

\begin{equation}
    \nabla_\theta \mathcal{L}_\pi(\theta) = \mathbb{E}_{X_{[0,1]} \sim \mathbb{P}} \bigg[ \int_0^1 \langle (v_\theta(X_t, t) - \mathbb{E}_\pi [v_q(X_t, t|X_t)]) + \beta (v_\theta(X_t, t) - \mathbb{E}_\pi [v_{\theta_0}(X_t, t|X_t)]), \nabla_\theta v_\theta (X_t, t) \rangle dt \bigg].
\end{equation}

This is exactly the gradient of \(\mathcal{L}(\theta)\) when the unconditional objective uses the marginal vector fields, so we can conclude that 

\begin{equation}
    \nabla_\theta \mathcal{L}_\pi(\theta) = \nabla_\theta \mathcal{L}(\theta).
\end{equation}

\section{Practical Implementation of Gradual Fine-Tuning}\label{ap:Prac_Impl}

Our proposed objective function consists of two KL divergence terms. 

\begin{equation}
    \mathcal{L}(\theta) = KL(\mathbb{P}_\theta \| \mathbb{P}_q) + \beta KL(\mathbb{P}_\theta \| \mathbb{P}_{\theta_0})
\end{equation}

We can therefore apply Proposition \ref{pr:KL_Girsanov} to both terms to rewrite the expression.

\begin{equation}
    KL(\mathbb{P}_\theta \| \mathbb{P}_q) = \mathbb{E}_{ \mathbb{P}_{\theta}} \left[ \frac{1}{2} \int_0^1 \| \sigma_t^{-1} (v_\theta - v_{q}) \|^2 dt \right]
\end{equation}

\begin{equation}
    KL(\mathbb{P}_\theta \| \mathbb{P}_{\theta_0}) = \mathbb{E}_{ \mathbb{P}_{\theta}} \left[ \frac{1}{2} \int_0^1 \| \sigma_t^{-1} (v_\theta - v_{\theta_0}) \|^2 dt \right]
\end{equation}

Here, \(v_{\theta_0}\) is the pretrained base vector field, and \(v_q\) is the optimal vector field that transports samples from the source distribution \(p_0\) to the target fine-tuning distribution \(q\). Since this vector field is unknown, it can be defined in the same way that the CFM \cite{lipman2023flow} defines a target vector field to be matched. We show empirically that the diffusion coefficient \(\sigma_t\) can simply be set to 1 while maintaining high performance. Note that both terms of the objective are now written using the drift terms of the base and target processes, and that the parameters \(\theta\) directly determine the path measure \(\mathbb{P}_\theta\). The otherwise intractable objective therefore now optimizes directly over the parameters \(\theta\). We can now write a tractable version of the proposed objective. 

\begin{equation}
    \mathcal{L}(\theta) = \mathbb{E}_{ \mathbb{P}_{\theta}} \left[ \frac{1}{2} \int_0^1 \| v_\theta - v_{q} \|^2 dt \right] + \beta \cdot \mathbb{E}_{ \mathbb{P}_{\theta}} \left[ \frac{1}{2} \int_0^1 \| v_\theta - v_{\theta_0} \|^2 dt \right]
\end{equation}

%% file: app_expt.tex
\section{Implementation Details}\label{ap:imp_details}

We use the exact U-Net architecture given by \cite{tong2024improving} and \cite{tong2023simulationfree} under MIT License, which has been pretrained on the Cifar-10 dataset \cite{Krizhevsky09learningmultiple} for 400,000 epochs. Minibatch OT couplings are used during pretraining and fine-tuning of the model, in accordance with the convergence results given in Section \ref{sc:Cond_GFT}. All GFT experiments use an inverse sigmoid \(\beta_s\) cooling schedule and an Adam Optimizer.

All experiments were run on a single A100 GPU. The large scale image generation experiments each ran for approximately 12 hours for Cross-Domain and 8 hours for In-Domain, for a total of about 60 GPU hours. The 2D experiments ran for approximately 1 hour in total. This does not include the time or computational resources used for preliminary experiments that are not included in this paper.

\subsection{Datasets} 
We fine-tune the pretrained image generation model for three datasets within the WILDS benchmarks datasets \cite{wilds2021} (MIT Liscense). 

Camelyon17 \cite{Camelyon17} is a dataset of H\&E-stained histopathology images of lymph node sections of breast cancer patients. The dataset includes both tumor and non-tumor images. The distribution shift between each split of this dataset is caused experimental batch effects and population variations. The Camelyon17 train set has a total of 302,436 samples, and 60,490 of these samples are held out from training for use in the FID calculations. The Camelyon17 validation set is substantially smaller than the train dataset, with a total of 34,904 samples, 6,981 of which are held out for use in the FID calculations. 

RxRx1 \cite{RxRx1} is a dataset of fluorescent microscopy images of human tissue sections. The distribution shift between each split of this dataset is caused experimental batch effects, such as variations in temperature, humidity, and reagent concentration. The RxRx1 experiments are conducted on a compressed version of the dataset to enable running on a single GPU. The RxRx1 train set has a total of 40,612 samples, 8,123 of which are held out for FID calculations. The RxRx1 validation set is smaller than the train dataset, with a total of 9,854 samples, and 1,971 of these samples are held out from fine-tuning for use in the FID calculations.

FMoW \cite{fmow2018} is a dataset of global satellite images of regions with varying land use. The experiments shown below are conducted on the RGB image version of the dataset, rather than the multispectral data. The distribution shift between each split of this dataset is caused temporal evolution. The FMoW experiments are conducted on a compressed version of the dataset to enable running on a single GPU. The FMoW train set has a total of 76,863 samples, and 15,373 are held out for use in the FID calculations. The FMoW validation set has a total of 19,915 samples, 3,983 of which are held out from fine-tuning for FID calculations.

\paragraph{Stability and Convergence Analysis}  We report three metrics to quantitatively analyze the convergence speed and stability of GFT. \textit{Instantaneous Variance} measures the local variance in accuracy convergence. At 50 epoch intervals, the instantaneous mean is found using a Radial Basis Function (RBF) kernel over a window of 10 FID measurements. This mean is then used to calculate variance with the same weighting and window size. The reported value is the average of all instantaneous variances over the scope of fine-tuning. \textit{Average convergence rate} is calculated by performing linear regression on windows of 10 consecutive FID measurements. The reported value is the average of the absolute value of the resulting slopes. Finally, \textit{Spearman correlation} is calculated between fine-tuning epochs and FID. For a stable fine-tuning process, we expect a strong negative correlation. A Spearman correlation coefficient close to -1 would indicate that as fine-tuning progresses, the model performance consistently improves with minimal regressive spikes or oscillation. 

\paragraph{Path Length Calculation} As a proxy to estimate the generation efficiency of each method we test, we report average probability path length. We can define the path length of a single sample through a vector field as its energy cost, or its accumulated kinetic energy over time. By the Benamou-Brenier formulation of optimal transport, this can be calculated by integrating the square of the velocity magnitude over the path of the given sample.

\begin{equation}
    L(x_0) = \int_0^1 \|\frac{dx}{dt}\|^2 dt = \int_0^1 \|\mathbf{v}(x_t, t)\|^2 dt
\end{equation}

Given a prior distribution over initial samples \(p(x_0)\), we can get the expected length of the vector field.

\begin{equation}
    \mathbb{E}_{p(x_0)}[L(x_0)] = \mathbb{E}_{x_0 \sim p(x_0)}\left[ \int_0^1 \|\mathbf{v}(x_t, t)\|^2 dt \right]
\end{equation}

The expectation over the prior distribution \(p_0\) can be approximated by averaging over a minibatch of \(M\) samples. We can apply Euler discretization with some chosen step size \(\Delta t\) to approximate the integral. 

\begin{equation}
    L(\theta) = \frac{1}{M} \sum_{j=1}^M \sum_{i=1}^N \|\mathbf{v}_\theta (x_i^{(j)}, t_i)\|^2 \cdot \Delta t \label{eq:Euler_pl}
\end{equation}

We estimate average path length for a model with parameters \(\theta\) with 100 discretization steps and 64 initial samples.

\section{2D Experiments}\label{ap:2d}

All small scale 2D experiments are run on a single A100 GPU using simulated datasets. All fine-tuning methods are run until convergence, with GFT and CFM taking 500 epochs, Adjoint Matching taking 1500 epochs, and ORW-CFM-W2 taking 1000 epochs. The ORW-CFM-W2 experiment uses the hyperparameters \(\tau = 4.5\) and \(\alpha=0.001\). 

\subsection{Extended Adjoint Matching Results}

\begin{figure}[t]
\centering
\begin{subfigure}[t]{.3\textwidth}
  \centering
  \includegraphics[width=.9\linewidth]{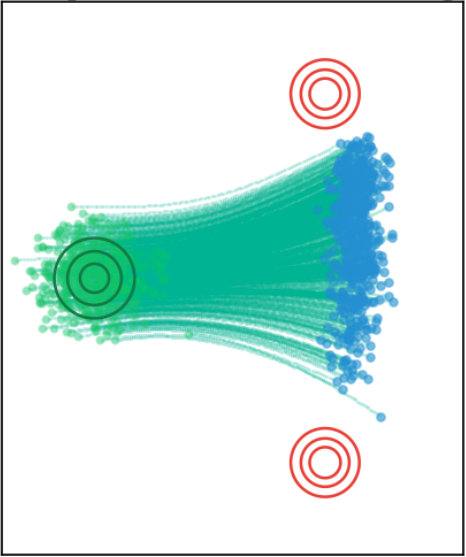}
  \caption{200 Epochs.}   \label{fig:sub1}
\end{subfigure}\hfill
\begin{subfigure}[t]{.3\textwidth}
  \centering
  \includegraphics[width=.85\linewidth]{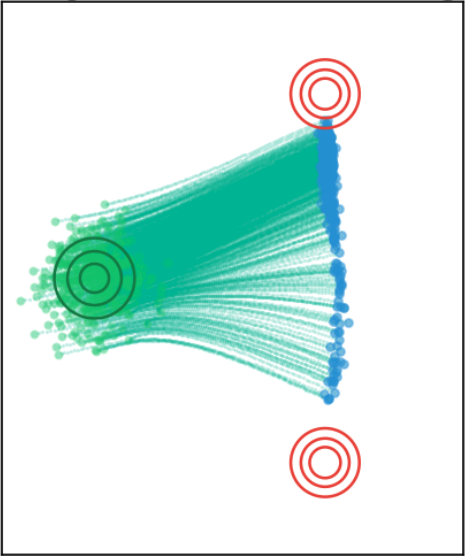}
  \caption{500 Epochs.}
  \label{fig:sub2}
\end{subfigure}\hfill
\begin{subfigure}[t]{.3\textwidth}
  \centering
  \includegraphics[width=.85\linewidth]{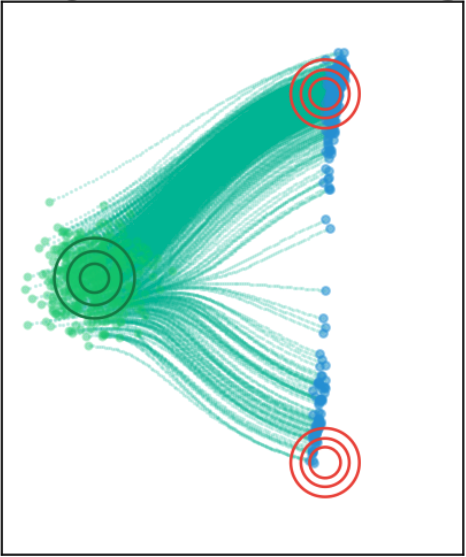}
  \caption{1000 Epochs.}
  \label{fig:sub2}
\end{subfigure}\hfill


\begin{subfigure}[t]{.3\textwidth}
  \centering
  \includegraphics[width=.85\linewidth]{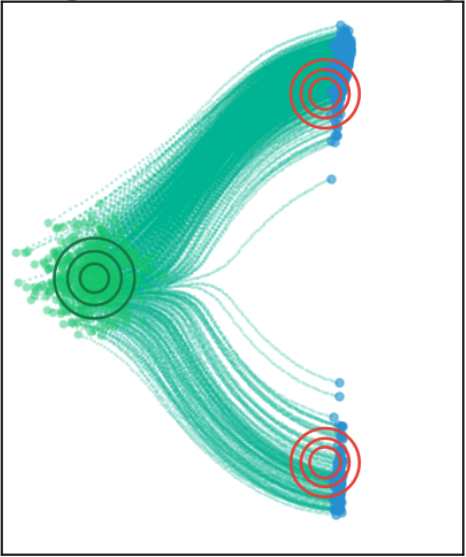}
  \caption{1500 Epochs.}
  \label{fig:sub2}
\end{subfigure}\hfill
\begin{subfigure}[t]{.3\textwidth}
  \centering
  \includegraphics[width=.85\linewidth]{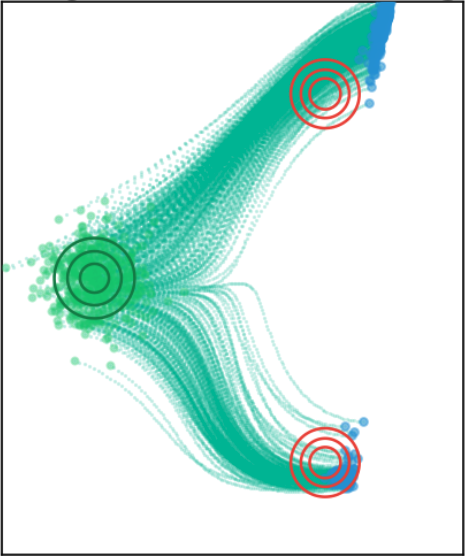}
  \caption{2000 Epochs.}
  \label{fig:sub2}
\end{subfigure}\hfill
\begin{subfigure}[t]{.3\textwidth}
  \centering
  \includegraphics[width=.85\linewidth]{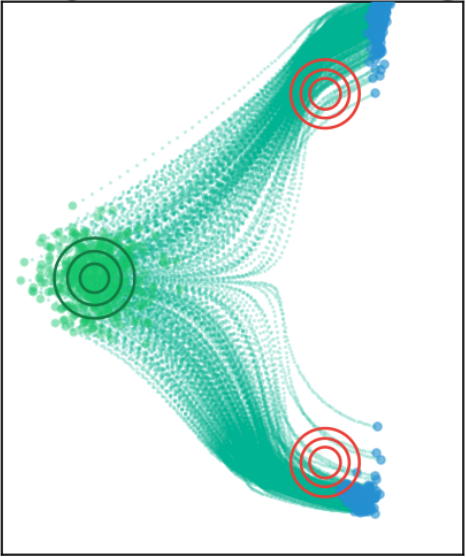}
  \caption{3000 Epochs.}
  \label{fig:sub2}
\end{subfigure}
\caption{Generated trajectories after fine-tuning with Adjoint Matching. Extended fine-tuning steadily reduces diversity and shifts the generated samples away from the mode centers.}
\label{fig:extended_AM}
\end{figure}

Fine-tuning with Adjoint Matching does not recover both modes of the target distribution until after 1000 epochs, compared to around 500 epochs for both GFT and CFM (Figure \ref{fig:extended_AM}). Additionally, the target modes are recovered with a significantly decreased diversity than exhibited by the pretrained model, and the generation diversity decreases as fine-tuning proceeds. Finally, when trained further after recovering the target modes, Adjoint Matching shifts the solution away from the mode means, decreasing accuracy as fine-tuning continues (Figure \ref{fig:extended_AM}f). This is in contrast to GFT, which maintains the diversity of the pretrained model while closely matching the covariance of both modes in the new target distribution.

\section{GFT Convergence Analaysis}
\label{ap:GFT_Convergence}

\subsection{Assumptions}

\begin{assumption}[Function space minimization]
\label{ass:exact}
At each $\beta > 0$, the vector field $v_\theta$ exactly
minimizes $\mathcal{L}_\pi(\theta)$ in function space,
yielding a pointwise minimizer $v_\theta^*$.
\end{assumption}

\begin{assumption}[Bounded mismatch under $\mathbb{P}_q$]
\label{ass:bounded}
The pretrained vector field $v_{\theta_0}$ and the marginal CFM
target $v_q$ satisfy:
\begin{equation}
    C' := \int_0^1 \mathbb{E}_{\mathbb{P}_q}
    \|v_{\theta_0}(X^q_t,t) - v_q(X^q_t,t)\|^2\,dt
    < \infty
    \label{eq:Cprime}
\end{equation}
where the expectation is over $X^q_t$ drawn from the
process induced by $v_q$ with initial state $X_0 \sim p_0$. The constant $C'$ therefore measures the $L^2$ mismatch between the pretrained and target vector fields under the target path measure, and is large when the distribution shift from $p_1$ to $q$ is large.
\end{assumption}

\begin{assumption}[Bounded conditional variance]
\label{ass:condvar}
The conditional variance of $v_q(\cdot | Z)$ around its mean $v_q(\cdot)$
is finite:
\begin{equation}
    V := \mathbb{E}_{Z \sim \pi}\int_0^1
    \|v_q(X_t,t) - v_q(X_t,t \mid Z)\|^2\,dt
    < \infty
    \label{eq:V}
\end{equation}
This quantity is the irreducible variance introduced by the coupling $\pi$, and is equal to the standard CFM loss
evaluated at $v_q(\cdot)$.
\end{assumption}

\begin{assumption}[Lipschitz vector fields]
\label{ass:lipschitz}
Both $v_q$ and $v_{\theta_0}$ are Lipschitz continuous in $x$,
uniformly in $t$, such that there exist constants $L_q, L_0 < \infty$ such that for all $x, y \in \mathbb{R}^d$ and $t \in [0,1]$:
\begin{equation}
    \|v_q(x,t) - v_q(y,t)\| \leq L_q\|x-y\|,
    \qquad
    \|v_{\theta_0}(x,t) - v_{\theta_0}(y,t)\| \leq L_0\|x-y\|
    \label{eq:lipschitz}
\end{equation}
We write $L = \max(L_q, L_0)$.
\end{assumption}

\subsection{Main Result}

\begin{theorem}[GFT Convergence]
Under Assumptions~\ref{ass:exact}--\ref{ass:lipschitz},
the pointwise minimizer of $\mathcal{L}_\pi$ at temperature
$\beta > 0$ is:
\begin{equation}
    v_\theta^*(X_t, t) = \frac{1}{1+\beta}\,v_q(X_t,t)
    + \frac{\beta}{1+\beta}\,v_{\theta_0}(X_t,t)
    \label{eq:minimizer}
\end{equation}
and the following hold:

\begin{enumerate}
    \item[(i)] $L^2$ \textbf{convergence of vector fields.}
    $v_\beta^* \to v_q$ in $L^2(\mathbb{P}_q)$ as $\beta \to 0^+$,
    with the explicit rate:
    \begin{equation}
        \int_0^1 \mathbb{E}_{\mathbb{P}_q}\|v_\beta^* - v_q\|^2\,dt
        \leq \frac{\beta^2}{(1+\beta)^2}\,C'
        \label{eq:l2_convergence}
    \end{equation}

    \item[(ii)] \textbf{Terminal distribution convergence.}
    The terminal distribution $p_\beta^*$ of the GFT process
    converges to $q$ in Wasserstein-2 distance as $\beta \to 0^+$:
    \begin{equation}
        W_2^2(p_\beta^*, q)
        \leq \frac{\beta^2\,e^{2L+1}}{(1+\beta)^2}\,C'
        \label{eq:w2_convergence}
    \end{equation}

    \item[(iii)] \textbf{Loss decomposition.}
    The GFT loss at the minimizer $v_\beta^*$ satisfies:
    \begin{equation}
        \mathcal{L}_\pi(v_\beta^*)
        = \frac{\beta}{1+\beta}\,C + V
        \label{eq:loss_decomp}
    \end{equation}
    where $C = \mathbb{E}_{Z\sim\pi}\int_0^1
    \|v_{\theta_0} - v_q\|^2\,dt$,
    and $\mathcal{L}_\pi(v_\beta^*) \to V$ as $\beta \to 0^+$.
\end{enumerate}
\end{theorem}

\textit{proof.}\\
We begin by deriving the pointwise minimizer for a fixed \(\beta > 0
\). Since $v_\theta(X_t,t)$ is unconditional and does not depend on $Z$, we minimize $\mathcal{L}_\pi(\theta)$ pointwise
in $(X_t,t)$ by differentiating inside the expectation over $Z \sim \pi$ and setting the result to zero:
\begin{equation}
    \mathbb{E}_{Z \sim \pi}\Bigl[
        2\bigl(v_\theta - v_q(X_t,t \mid Z)\bigr)
        + 2\beta\bigl(v_\theta - v_{\theta_0}(X_t,t)\bigr)
    \Bigr] = 0
    \label{eq:foc}
\end{equation}
Expanding and collecting all terms in $v_\theta$:
\begin{equation}
    (1+\beta)v_\theta
    = \mathbb{E}_{Z \sim \pi}[v_q(X_t,t \mid Z)]
    + \beta\,\mathbb{E}_{Z \sim \pi}[v_{\theta_0}(X_t,t)]
    \label{eq:collect}
\end{equation}
Since $v_{\theta_0}$ is unconditional, $\mathbb{E}_{Z\sim\pi}
[v_{\theta_0}(X_t,t)] = v_{\theta_0}(X_t,t)$. Applying the
marginalization property of the target vector field,
$\mathbb{E}_{Z\sim\pi}[v_q(X_t,t\mid Z)\mid X_t] =
v_q(X_t,t)$, and dividing by $(1+\beta)$ gives the
minimizer~\eqref{eq:minimizer}. The second derivative
of the objective with respect to $v_\theta$ is $2(1+\beta)
> 0$, confirming this is a minimum.

\textbf{Result (i): $L^2$ convergence of $v_\beta^* \to v_q$.}

Subtracting $v_q$ from both sides
of~\eqref{eq:minimizer} gives the pointwise identity:
\begin{equation}
    v_\beta^*(X_t,t) - v_q(X_t,t)
    = \frac{\beta}{1+\beta}
      \bigl(v_{\theta_0}(X_t,t) - v_q(X_t,t)\bigr)
    \label{eq:pointwise}
\end{equation}
This holds for every $(X_t, t) \in \mathbb{R}^d \times [0,1]$.
Taking the squared norm, integrating over $t$, and taking the expectation under $\mathbb{P}_q$ gives
\begin{equation}
    \int_0^1 \mathbb{E}_{\mathbb{P}-q}\|v_\beta^* - v_q\|^2\,dt
    = \frac{\beta^2}{(1+\beta)^2}
      \int_0^1 \mathbb{E}_{\mathbb{P}_q}
      \|v_{\theta_0} - v_q\|^2\,dt
    \leq \frac{\beta^2}{(1+\beta)^2}\,C'
    \label{eq:l2_bound}
\end{equation}
using Assumption~\ref{ass:bounded}. As $\beta \to 0^+$,
the prefactor $\frac{\beta^2}{(1+\beta)^2} \to 0$,
establishing $v_\beta^* \to v_q$ in $L^2(\mathbb{P}_q)$ at a rate of $O(\beta^2)$.

\textbf{Result (ii): Terminal distribution convergence
via Gronwall's Inequality.}

We couple the GFT process $X^*_t$ (driven by $v_\beta^*$)
and the target process $X^q_t$ (driven by $v_q$)
through the \emph{same} initial condition $X_0 \sim p_0$ and the \emph{same} Brownian motion $B_t$:
\begin{align}
    dX^*_t &= v_\beta
    ^*(X^*_t,t)\,dt + \sigma_t\,dB_t,
    \qquad X^*_0 = X_0
    \label{eq:gft_process}\\
    dX^q_t &= v_q(X^q_t,t)\,dt + \sigma_t\,dB_t,
    \qquad X^q_0 = X_0
    \label{eq:target_process}
\end{align}
We now define a difference process $\delta_t = X^*_t -
X^q_t$. Since both processes share the same $X_0$ and $B_t$, the diffusion terms cancel exactly:
\begin{equation}
    d\delta_t =
    \bigl(v_\beta^*(X^*_t,t) - v_q(X^q_t,t)\bigr)dt
    \label{eq:delta}
\end{equation}
with $\delta_0 = X^*_0 - X^q_0 = 0$. Thus $\delta_t$
evolves as a deterministic ODE given $X_0$, with
randomness entering only through the shared initial
condition.

We compute the time derivative of $\|\delta_t\|^2$:
\begin{align}
    \frac{d}{dt}\|\delta_t\|^2
    &= 2\delta_t^\top \frac{d\delta_t}{dt} \notag\\
    &= 2\delta_t^\top
       \bigl(v_\beta^*(X^*_t,t) - v_q(X^q_t,t)\bigr)
    \label{eq:time_deriv}
\end{align}
Adding and subtracting $v_\beta^*(X^q_t,t)$:
\begin{align}
    \frac{d}{dt}\|\delta_t\|^2
    &= 2\delta_t^\top
       \bigl(v_\beta^*(X^*_t,t) - v_\beta^*(X^q_t,t)\bigr)
     + 2\delta_t^\top
       \bigl(v_\beta^*(X^q_t,t) - v_q(X^q_t,t)\bigr)
    \label{eq:split}
\end{align}
For the first term, by Cauchy-Schwarz and the
Lipschitz continuity of $v_\beta^*$, since $v_\beta^*$ is a
convex combination of $v_q$ and $v_{\theta_0}$
\eqref{eq:minimizer}, it inherits Lipschitz
continuity with constant at most $\frac{L_q +
\beta L_0}{1+\beta} \leq \max(L_q, L_0) = L$ from
Assumption~\ref{ass:lipschitz}. Using $\|X^*_t -
X^q_t\| = \|\delta_t\|$:
\begin{equation}
    2\delta_t^\top\bigl(v_\beta^*(X^*_t,t)
    - v_\beta^*(X^q_t,t)\bigr)
    \leq 2\|\delta_t\|\cdot L\|\delta_t\|
    = 2L\|\delta_t\|^2
    \label{eq:first_term}
\end{equation}
For the second term, by Young's inequality
$2a^\top b \leq \|a\|^2 + \|b\|^2$:
\begin{equation}
    2\delta_t^\top\bigl(v_\beta^*(X^q_t,t)
    - v_q(X^q_t,t)\bigr)
    \leq \|\delta_t\|^2
    + \|v_\beta^*(X^q_t,t) - v_q(X^q_t,t)\|^2
    \label{eq:second_term}
\end{equation}
Combining~\eqref{eq:first_term}
and~\eqref{eq:second_term}:
\begin{equation}
    \frac{d}{dt}\|\delta_t\|^2
    \leq (2L+1)\|\delta_t\|^2
    + \|v_\beta^*(X^q_t,t) - v_q(X^q_t,t)\|^2
    \label{eq:gronwall_ode}
\end{equation}
Taking expectation over $X_0 \sim p_0$:
\begin{equation}
    \frac{d}{dt}\mathbb{E}\|\delta_t\|^2
    \leq (2L+1)\mathbb{E}\|\delta_t\|^2
    + \mathbb{E}_{\mathbb{P}_q}\|v_\beta^*(X^q_t,t)
      - v_q(X^q_t,t)\|^2
    \label{eq:gronwall_expect}
\end{equation}
Applying Gronwall's inequality with initial
condition $\mathbb{E}\|\delta_0\|^2 = 0$:
\begin{equation}
    \mathbb{E}\|\delta_1\|^2
    \leq \int_0^1 e^{(2L+1)(1-s)}
    \mathbb{E}_{\mathbb{P}_q}\|v_\beta^*(X^q_s,s)
    - v_q(X^q_s,s)\|^2\,ds
    \leq e^{2L+1}\int_0^1
    \mathbb{E}_{\mathbb{P}_q}\|v_\beta^*(X^q_s,s)
    - v_q(X^q_s,s)\|^2\,ds
    \label{eq:gronwall_bound}
\end{equation}
Substituting the pointwise
identity~\eqref{eq:pointwise} and
applying~\eqref{eq:l2_bound}:
\begin{equation}
    \mathbb{E}\|\delta_1\|^2
    \leq \frac{\beta^2\,e^{2L+1}}{(1+\beta)^2}\,C'
    \label{eq:delta_bound}
\end{equation}
To obtain the $W_2$ bound, we note that our coupled
construction $(X^*_1, X^q_1)$ is one particular
coupling of $p_\beta^*$ and $q$ with marginals $p^*_\beta$
and $q$ respectively. Since $W_2^2(p_\beta^*, q)$ is
defined as the infimum of $\mathbb{E}\|X-Y\|^2$ over
\emph{all} couplings of $p_\beta^*$ and $q$, it is bounded above
by the expectation under any specific coupling,
including ours:
\begin{equation}
    W_2^2(p_\beta^*, q)
    \leq \mathbb{E}\|X^*_1 - X^q_1\|^2
    = \mathbb{E}\|\delta_1\|^2
    \leq \frac{\beta^2\,e^{2L+1}}{(1+\beta)^2}\,C'
    \label{eq:w2_bound}
\end{equation}
This gives $W_2(p_\beta^*, q) \to 0$ at rate $O(\beta)$
as $\beta \to 0^+$.

\textbf{Results (iii): Loss decomposition.}

We substitute $v_\beta^*$ into $\mathcal{L}_\pi$ and handle each term. For the first term, we decompose:
\begin{equation}
    v_\beta^* - v_q(X_t,t \mid Z)
    = \underbrace{(v_\beta^* - v_q)}_
      {= \frac{\beta}{1+\beta}(v_{\theta_0} - v_q)}
    + (v_q - v_q(X_t,t \mid Z))
    \label{eq:decompose}
\end{equation}
using~\eqref{eq:pointwise}. Expanding the squared norm
and taking expectation over $Z \sim \pi$:
\begin{align}
    \mathbb{E}_Z\|v_\beta^* - v_q(\cdot \mid Z)\|^2
    &= \frac{\beta^2}{(1+\beta)^2}
       \|v_{\theta_0} - v_q\|^2 \notag\\
    &\quad + \frac{2\beta}{1+\beta}
      (v_{\theta_0} - v_q)^\top
      \underbrace{\mathbb{E}_Z[v_q - v_q(\cdot \mid Z)]}_{=\,0}
      \notag\\
    &\quad + \mathbb{E}_Z\|v_q - v_q(\cdot \mid Z)\|^2
    \label{eq:first_loss}
\end{align}
The cross term vanishes since $\mathbb{E}_{Z\sim\pi}
[v_q(X_t,t) - v_q(X_t,t \mid Z)]
= v_q - v_q = 0$ by the marginalization property of $v_q$. For the second term
of the loss:
\begin{equation}
    \|v_\beta^* - v_{\theta_0}\|^2
    = \frac{1}{(1+\beta)^2}
      \|v_q - v_{\theta_0}\|^2
    \label{eq:second_loss}
\end{equation}
using~\eqref{eq:pointwise} applied to $v_\beta^* -
v_{\theta_0} = \frac{1}{1+\beta}(v_q - v_{\theta_0})$.
Combining~\eqref{eq:first_loss}
and~\eqref{eq:second_loss}, integrating over $t$,
and collecting:
\begin{align}
    \mathcal{L}_\pi(v_\beta^*)
    &= \left(\frac{\beta^2}{(1+\beta)^2}
       + \frac{\beta}{(1+\beta)^2}\right)
       \mathbb{E}_{Z\sim\pi}\int_0^1
       \|v_{\theta_0} - v_q\|^2\,dt + V \notag\\
    &= \frac{\beta(\beta+1)}{(1+\beta)^2}\,C + V
     = \frac{\beta}{1+\beta}\,C + V
    \label{eq:loss_final}
\end{align}
As $\beta \to 0^+$, $\mathcal{L}_\pi(v_\beta^*) \to V$,
which is the standard CFM loss. The $\beta$-dependent
component $\frac{\beta}{1+\beta}C$ vanishes at
rate $O(\beta)$.

\section{GFT for Mean Flow Models}\label{ap:MeanFlow}

\subsection{Relationship Between Average and Instantaneous Optimums}

We assume \(u_{\theta_0}\), \(u_q\), and \(u_\theta\) be average velocity fields related to their corresponding instantaneous fields \(v_{\theta_0}\), \(v_q\), and \(v_\theta\) via the Mean Flow Identity:
\begin{equation}
    u(X_t, r, t) = v(X_t, t) - (t-r)\frac{d}{dt}u(X_t, r, t)~.
\end{equation}

The Mean Flow Identity can then be rewritten to express the average velocity field directly as an integral over the instantaneous velocity field.

\begin{equation}
    u(X_t, r, t) + (t-r)\frac{d}{dt}u(X_t, r, t) = v(X_t, t)
\end{equation}
\begin{equation}
    \frac{d}{dt}[(t-r)u(X_t, r, t)] = v(X_t, t)
\end{equation}

\begin{equation}
    \int_r^t \frac{d}{d \tau}[(\tau-r)u(X_{\tau}, r, \tau)]d\tau = \int_r^t v(X_\tau, \tau) d\tau
\end{equation}
\begin{equation}
    (t-r)u(X_t, r, t) = \int_r^t v(X_\tau, \tau) d\tau
\end{equation}
\begin{equation}
    u(X_t, r, t) = \frac{1}{(t-r)}\int_r^t v(X_\tau, \tau) d\tau
\label{eq:MF_int}
\end{equation}

Now suppose that the instantaneous vector field \(v_\theta\) takes the following form:
\begin{equation}
    v_\theta(X_t, t) = \left(\frac{1}{1+\beta}\right) v_q(X_t, t) + \left(\frac{\beta}{1+\beta}\right) u_{\theta_0}(X_t, t)~.
\label{eq:GFT_opt_ap}
\end{equation}

We substitute \eqref{eq:GFT_opt_ap} into \eqref{eq:MF_int}.

\begin{equation}
    u_\theta(X_t, r, t) = \frac{1}{(t-r)}\int_r^t v_\theta(X_\tau, \tau) d\tau
\end{equation}
\begin{equation}
    = \frac{1}{t - r}\int_r^t \left(\frac{1}{1+\beta}\right) v_q(X_\tau, \tau) + \left(\frac{\beta}{1+\beta}\right) v_{\theta_0}(X_\tau, \tau)\,d\tau
\end{equation}

By the linearity of the integral operator, we can separate the RHS into two terms:

\begin{equation}
    u_\theta(X_t, r, t) = \frac{1}{t - r}\int_r^t \left(\frac{1}{1+\beta}\right) v_q(X_\tau, \tau)\,d\tau + \frac{1}{t - r}\int_r^t\left(\frac{\beta}{1+\beta}\right) v_{\theta_0}(X_\tau, \tau)\,d\tau~.
\label{eq:avg_velocity_sep}
\end{equation}

By the definition of average velocity and its relation to instantaneous velocity, we have now recovered the target and pretrained average velocity fields.

\begin{equation}
    u_\theta(X_t, r, t) = \left(\frac{1}{1+\beta}\right) u_q(X_t, r, t) + \left(\frac{\beta}{1+\beta}\right) u_{\theta_0}(X_t, r, t)~.
\label{eq:avg_velocity_sep}
\end{equation}

Note that this relationship between the optimal instantaneous and average velocity fields also works in reverse. Assume now that we start from the average velocity field, written as a convex combination as given in \eqref{eq:avg_velocity_sep}. We apply the Mean Flow Identity to all average velocity terms.

\begin{equation}
    v_\theta^* - (t-r)\frac{d}{dt}u_\theta^* = \frac{1}{1+\beta}\!\left(v_q - (t-r)\frac{d}{dt}u_q\right) + \frac{\beta}{1+\beta}\!\left(v_{\theta_0} - (t-r)\frac{d}{dt}u_{\theta_0}\right)
\label{eq:sub_MF}
\end{equation}

Applying the linearity of \(\frac{d}{dt}\) to \eqref{eq:avg_velocity_sep} gives 
    
\begin{equation}
    \frac{d}{dt}u_\theta^* = \frac{1}{1+\beta}\frac{d}{dt}u_q + \frac{\beta}{1+\beta}\frac{d}{dt}u_{\theta_0}~.
\label{eq:proof_step2}
\end{equation}

Substituting this into the left-hand side of \eqref{eq:sub_MF}:
    
\begin{equation}
    v_\theta^* - (t-r)\!\left(\frac{1}{1+\beta}\frac{d}{dt}u_q + \frac{\beta}{1+\beta}\frac{d}{dt}u_{\theta_0}\right) = \frac{v_q + \beta v_{\theta_0}}{1+\beta} - (t-r)\!\left(\frac{1}{1+\beta}\frac{d}{dt}u_q + \frac{\beta}{1+\beta}\frac{d}{dt}u_{\theta_0}\right).
\label{eq:proof_step3}
\end{equation}

The second terms on both sides cancel, leaving the original instantaneous convex combination.

\begin{equation}
    v_\theta^* = \frac{v_q + \beta v_{\theta_0}}{1+\beta}~,
\end{equation}

\subsection{Mean Flow GFT Algorithm}

\begin{algorithm}[H]
\caption{MeanFlow GFT fine-tuning step}
\label{alg:GFT_MF}
\begin{algorithmic}[1]
\Require Current model $u_\theta$, frozen base model $u_{\theta_0}$, 
         data sample $x_1$, noise $x_0$, reference time $r$, time $t$, 
         temperature $\beta$
\Statex
\Statex \textbf{1. Compute interpolated sample and conditional target velocity}
\Statex $x_t \leftarrow (1-t)x_0 + t x_1$
\Statex $v_q \leftarrow x_1 - x_0$
\Statex
\Statex \textbf{2. Recover instantaneous velocities from Mean Flow models}
\Statex $v_{\theta_0} \leftarrow u_{\theta_0}(x_t,\, t,\, t)$ \quad (forward pass of frozen model at $r=t$)
\Statex $v_\theta \leftarrow u_\theta(x_t,\, t,\, t)$ \quad (forward pass of current model at $r=t$)
\Statex
\Statex \textbf{3. Compute the conditional GFT instantaneous target}
\Statex $v^\star \leftarrow \dfrac{v_q + \beta\, v_{\theta_0}}{1 + \beta}$
\Statex
\Statex \textbf{4. Compute JVP of current model using $v^\star$ as tangent}
\Statex $(u_\theta^{\text{out}},~\frac{d}{dt}u_\theta) 
    \leftarrow \texttt{jvp}\!\left(u_\theta(x_t, r, t),\;(v^\star,\;0,\;1)\right)$
\Statex
\Statex \textbf{5. Compute average velocity target via MeanFlow Identity}
\Statex $u^\star \leftarrow \mathrm{sg}\!\left(v^\star - (t-r)\dfrac{d}{dt}u_\theta\right)$
\Statex
\Statex \textbf{6. Compute loss}
\Statex $\mathcal{L} \leftarrow \left\|u_\theta^{\text{out}} - u^\star\right\|^2$
\Statex
\Return $\mathcal{L}$
\end{algorithmic}
\end{algorithm}

\subsection{Empirical Results}

\begin{figure*}
\centering
\begin{subfigure}{.45\textwidth}
  \centering
  \includegraphics[width=1.0\linewidth]{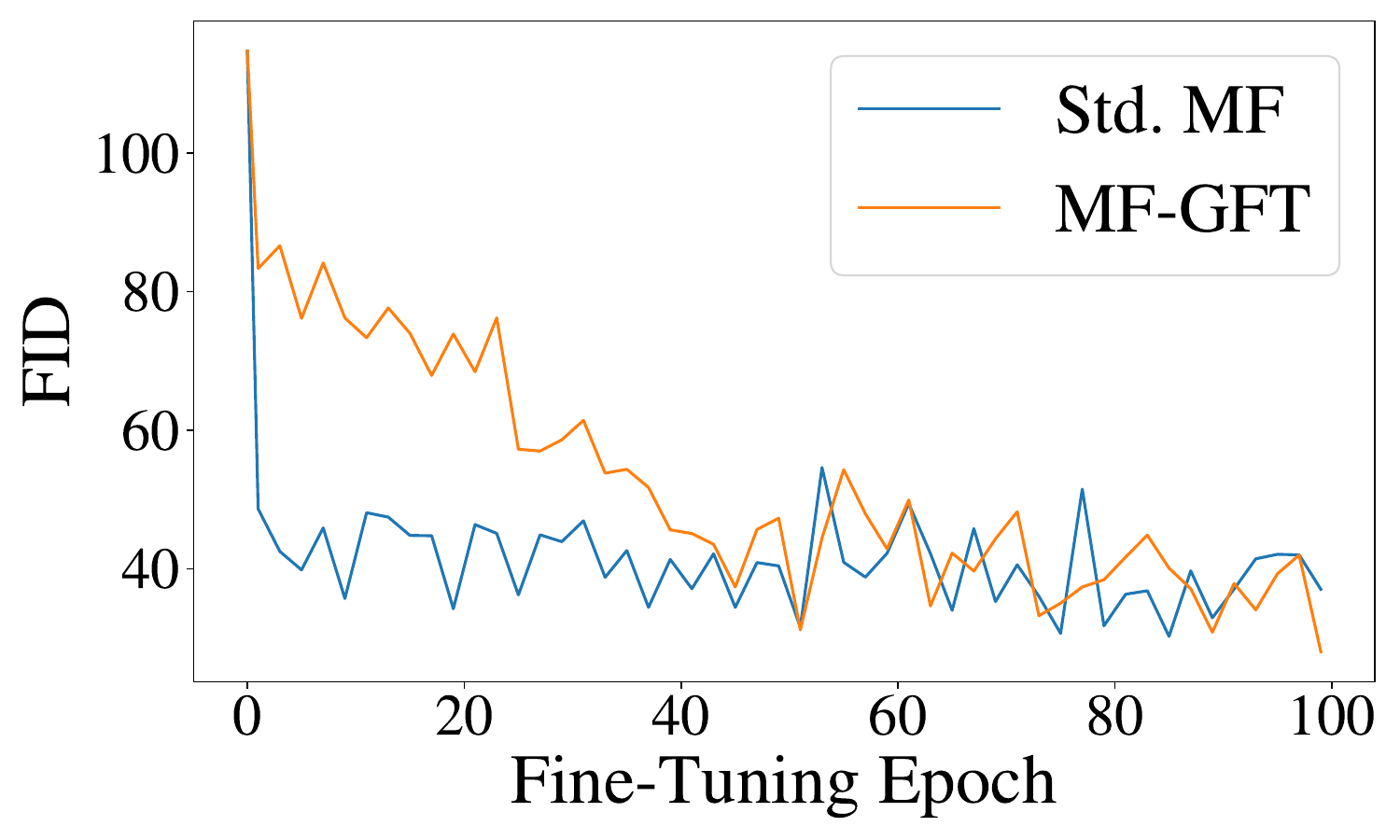}
  \caption{No EMA.}
  \label{fig:sub1}
\end{subfigure}%
\hspace{0.03\textwidth}
\begin{subfigure}{.45\textwidth}
  \centering
  \includegraphics[width=1.0\linewidth]{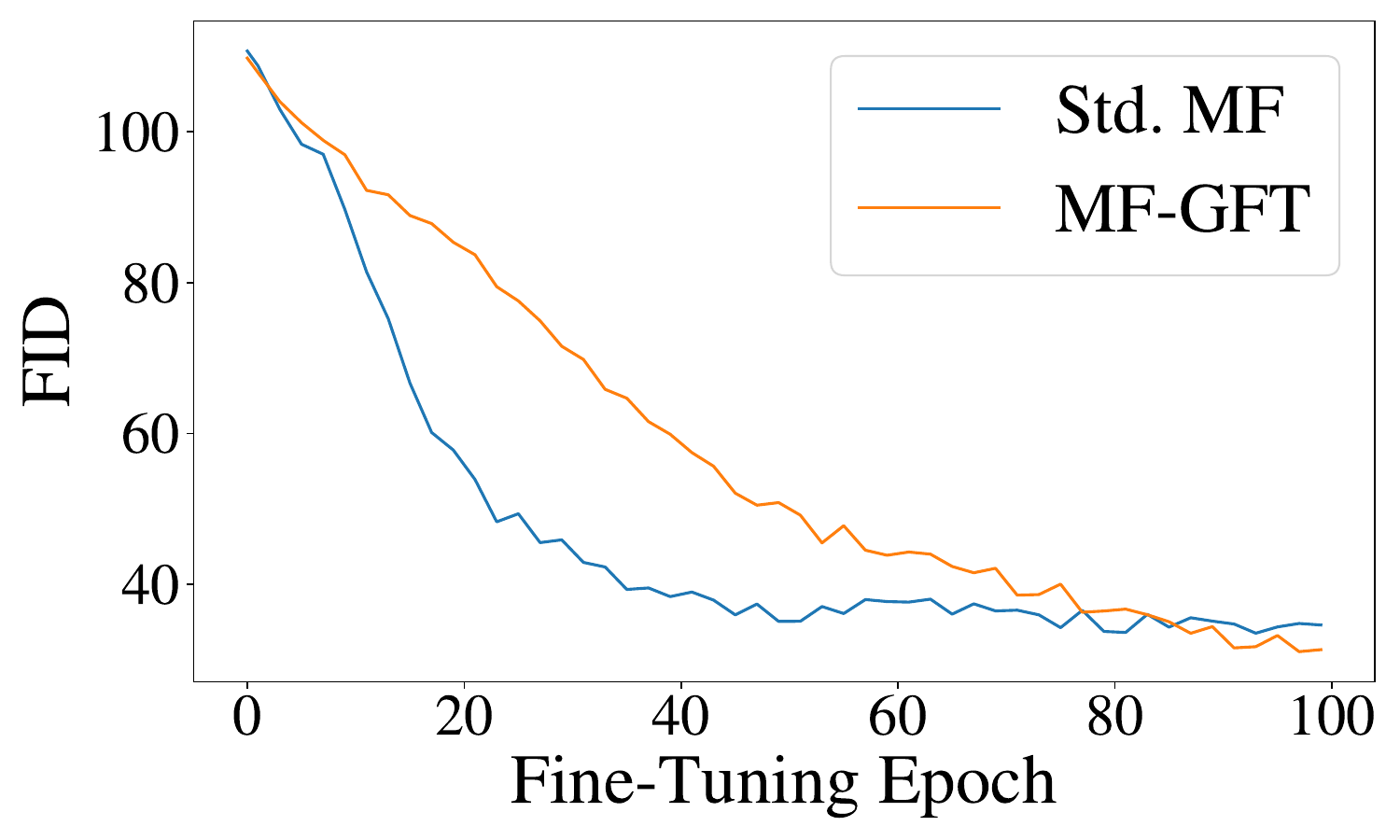}
  \caption{EMA.}
  \label{fig:sub2}
\end{subfigure}
\caption{Mean Flow Cross-domain adaptation on Camelyon17, with one-step generation. GFT achieves the same accuracy as standard MF fine-tuning, but with significantly higher stability.}
\label{fig:mean_flow}
\end{figure*}

We begin with a Mean Flow model pretrained on the Cifar10 dataset, and fine-tuning it to the Camelyon17 train dataset using the standard Mean Flow training algorithm, and the GFT Mean Flow algorithm (\ref{alg:GFT_MF}). The GFT Mean Flow algorithm reaches the same accuracy as the standard fine-tuning (Figure \ref{fig:mean_flow}), but with significantly improved stability in both the normal update and EMA update settings (Table \ref{tab:MF}). The increased stability provided by GFT is particularly important when working with one-step generative models, which are known to be highly unstable during training. Compared to an iterative generative model fine-tuned from Cifar10 to the Camelyon17 dataset, the MF-GFT model achieves comparable FID with a 50-100x reduction in model evaluations during inference, with only a minor decrease in performance. In fact, the final FID of the MF-GFT model without EMA updates (26.665) was lower than that of the LoRA CFM model with 100 NFEs (35.120).

\begin{table}
\centering
\caption{Convergence and stability analysis for MeanFlow on the Camelyon17 dataset. GFT increases stability for both normal and EMA-updated fine-tuning.}
\label{tab:stability}

\makebox[\textwidth][c]{
\begin{tabular}{clccc}
\toprule
& \textbf{Objective} 
& \textbf{Inst. Variance} $\downarrow$
& \textbf{Convergence Rate} $\uparrow$
& \textbf{Spearman $\rho$} $\downarrow$ \\
\midrule

\multirow{2}{*}{No EMA} 
&
Standard MF & 87.202 & 0.368 & -0.42 \\
& MF-GFT & \textbf{44.066} & \textbf{0.648} & \textbf{-0.88} \\

\hline

\multirow{2}{*}{EMA} 
& Standard MF & 34.433 & \textbf{0.805} & -0.93 \\
& MF-GFT & \textbf{16.341} & 0.794 & \textbf{-1.00} \\

\bottomrule
\end{tabular}
}
\label{tab:MF}
\end{table}

\section{Additional Results}

\subsection{Extended Results on Cross-Domain Adaptation} \label{ap:large_results}

\begin{figure}[H]
\centering
\setlength{\tabcolsep}{4pt}
\renewcommand{\arraystretch}{1.1}

\begin{tabular}{c c c}
    & \textbf{FID} & \textbf{Path Length}\\

\rotatebox{90}{\textbf{RxRx1}} &
\includegraphics[width=0.45\linewidth, valign=m]{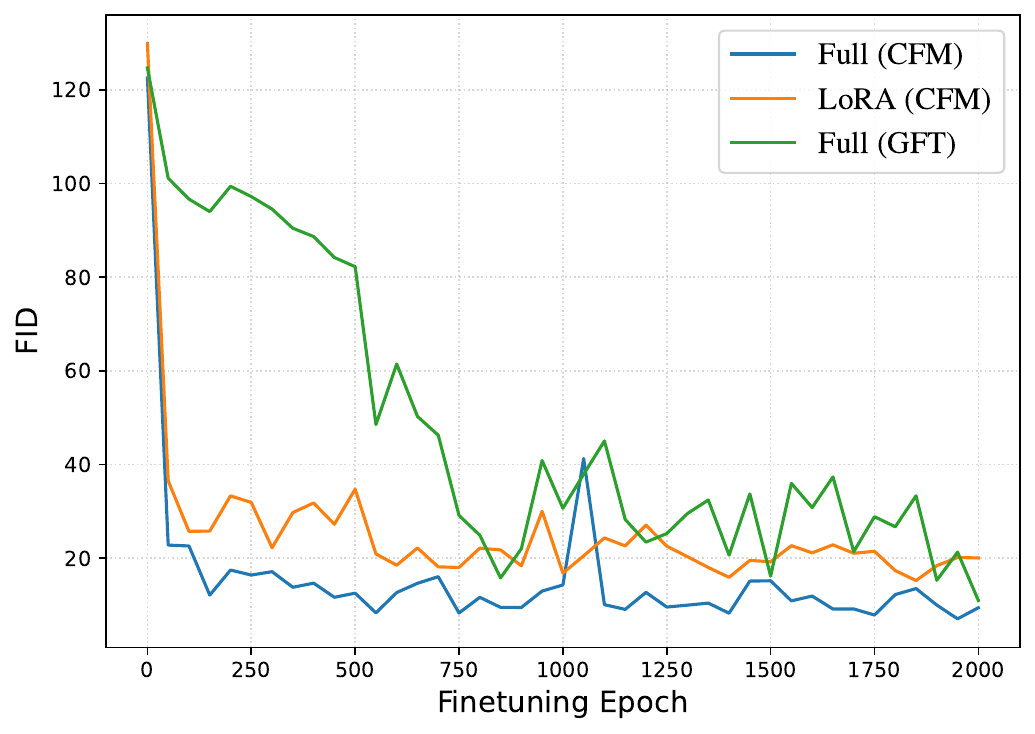} &
\includegraphics[width=0.45\linewidth, valign=m]{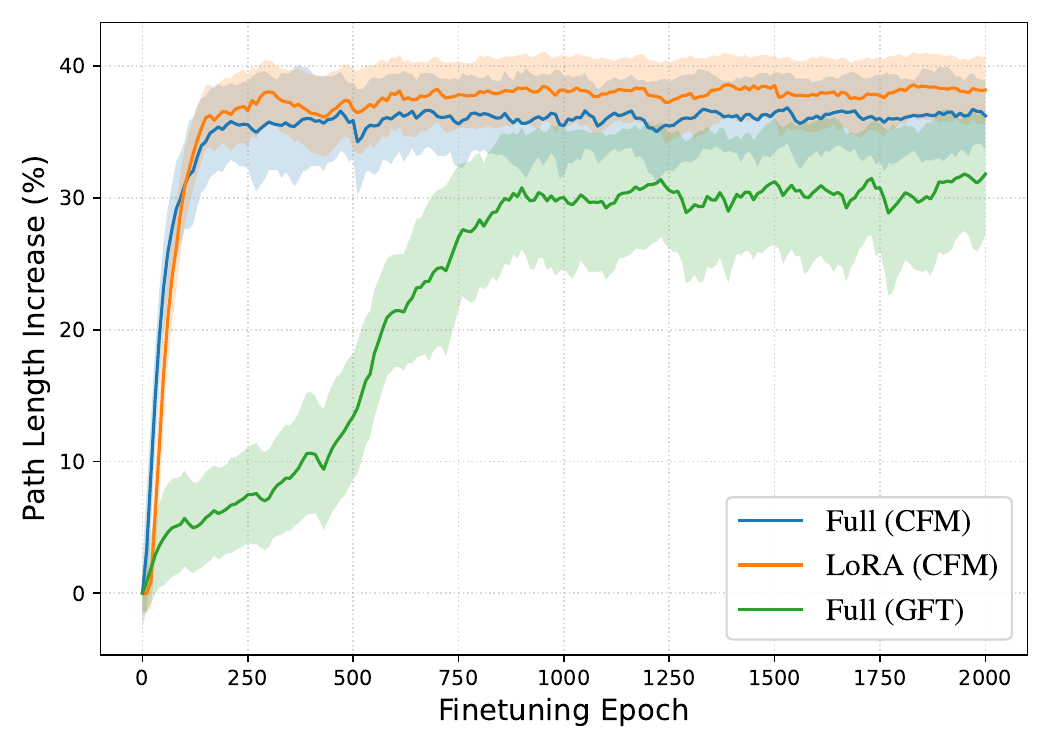} \\

\rotatebox{90}{\textbf{FMoW}} &
\includegraphics[width=0.45\linewidth, valign=m]{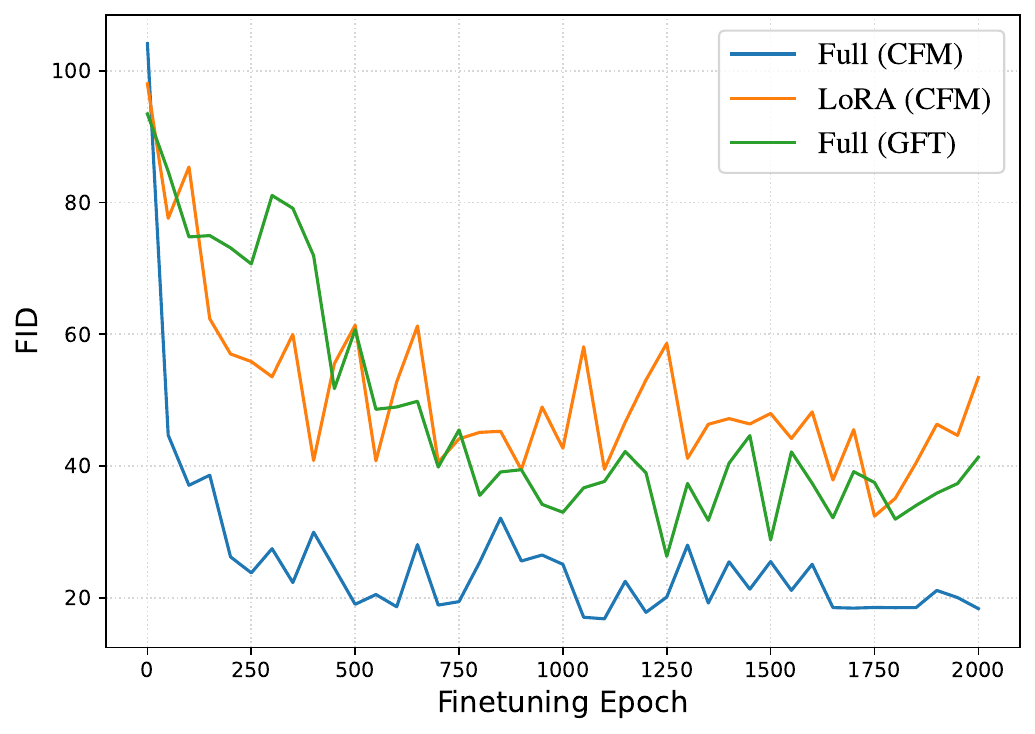} &
\includegraphics[width=0.45\linewidth, valign=m]{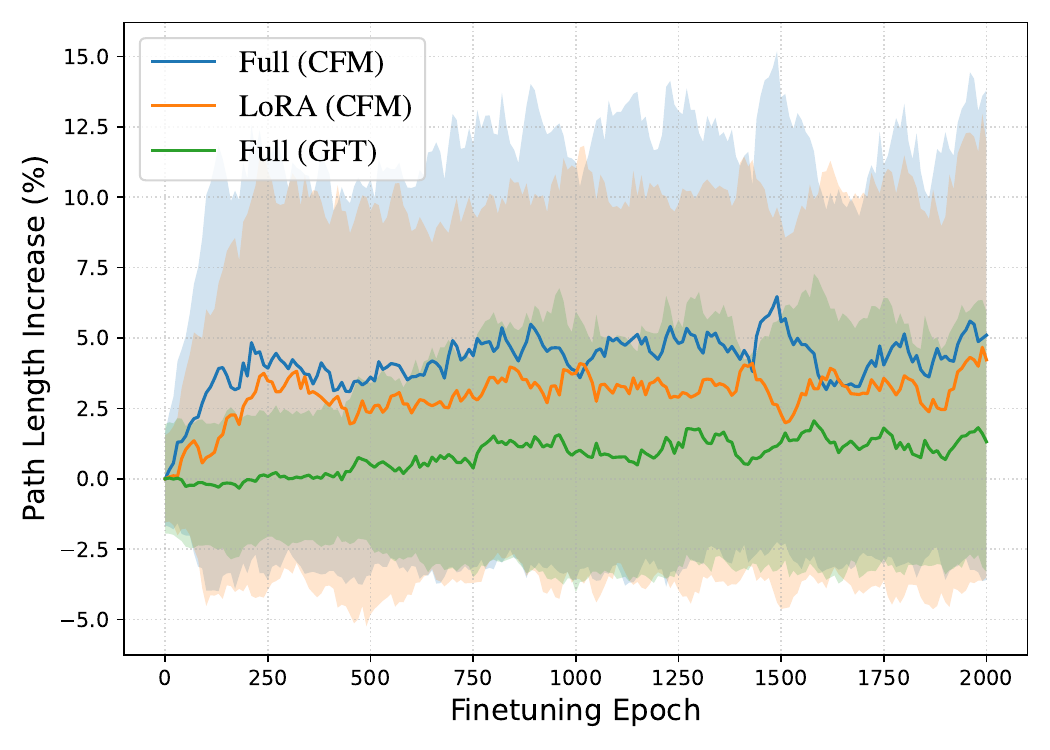} 
\end{tabular}

\caption{
Fine-tuning performance from a pretrained Cifar-10 model to various WILDS train datasets. Rows corresponding to individual datasets, and the columns show FID and probability path lengths through 2000 epochs of fine-tuning. The shaded regions of the path length graphs correspond to one standard deviation from the mean.
}
\label{fig:large_dist_shift}
\end{figure}

\subsection{Cross-Domain Adaptation Diversity}

\begin{table}
\centering
\caption{Cross-Domain Adaptation Vendi scores across datasets and fine-tuning methods. GFT matches or exceeds the diversity of CFM across all datasets.}
\label{tab:vendi_scores}
\begin{tabular}{clcc}
\toprule
\textbf{Dataset} & \textbf{Method} & \textbf{Vendi Score} & \textbf{\% of Original} \\
\midrule

\multirow{3}{*}{\textbf{Camelyon17}}
 & Original Data      & 8.51 & 100.00\% \\
 & Full (GFT) & 7.52 & 88.37\% \\
 & Full (CFM) & 6.63 & 77.91\% \\
\hline
\addlinespace
\multirow{3}{*}{\textbf{RxRx1}
} & Original Data      & 5.71 & 100.00\% \\
 & Full (GFT) & 4.66 & 81.61\% \\
 & Full (CFM) & 3.59 & 62.87\% \\
\hline
\addlinespace
\multirow{3}{*}{\textbf{FMoW}}
 & Original Data      & 21.57 & 100.00\% \\
 & Full (GFT) & 11.45 & 53.08\% \\
 & Full (CFM) & 11.75 & 54.47\% \\

\bottomrule
\end{tabular}
\label{tab:diversity}
\end{table}

\newpage
\subsection{Extended Results on In-Domain Adaptation} \label{ap:small_results}

\begin{figure}[H]
\centering
\setlength{\tabcolsep}{4pt}
\renewcommand{\arraystretch}{1.1}

\begin{tabular}{c c c}
    & \textbf{FID} & \textbf{Path Length}\\

\rotatebox{90}{\textbf{RxRx1}} &
\includegraphics[width=0.45\linewidth, valign=m]{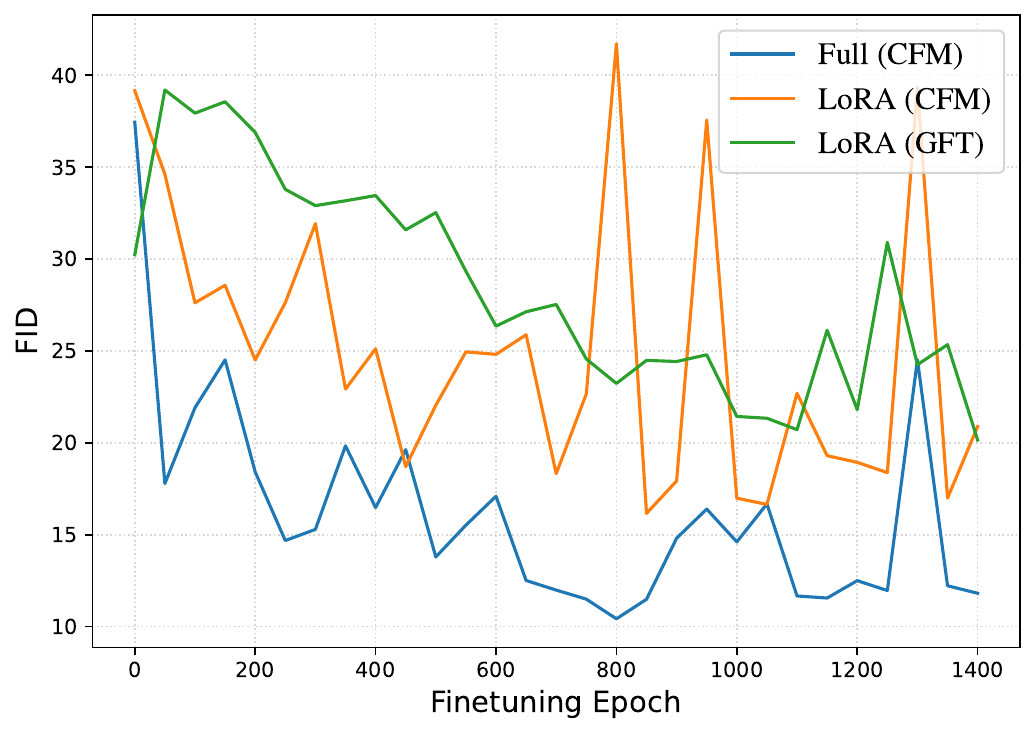} &
\includegraphics[width=0.45\linewidth, valign=m]{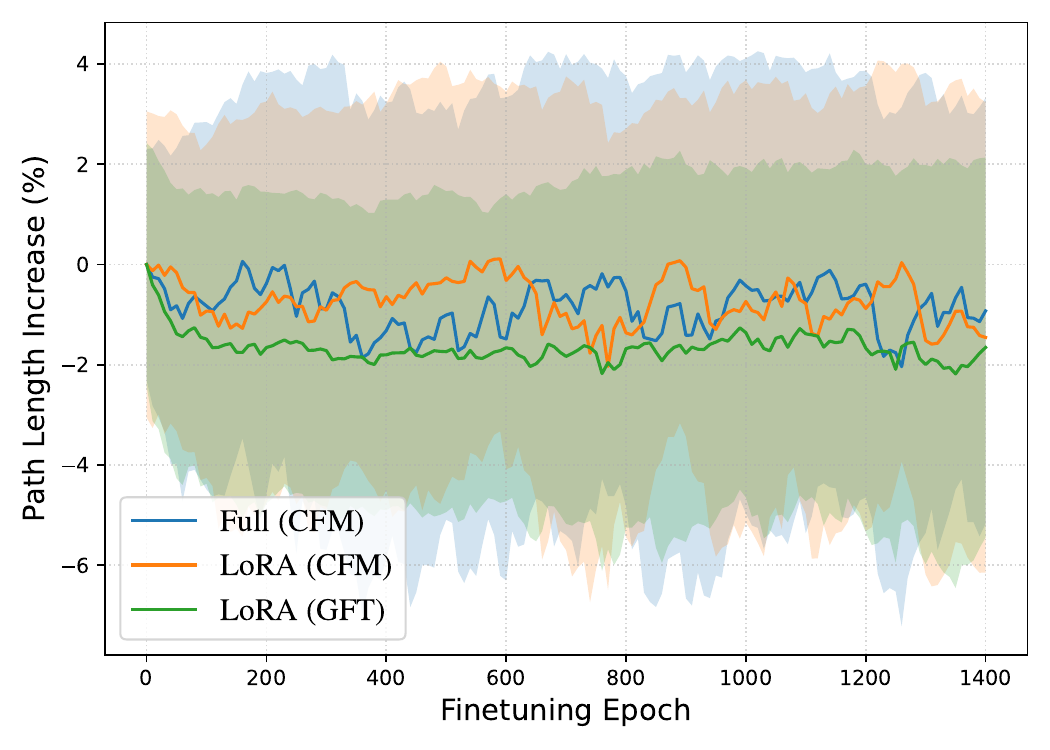} \\

\rotatebox{90}{\textbf{FMoW}} &
\includegraphics[width=0.45\linewidth, valign=m]{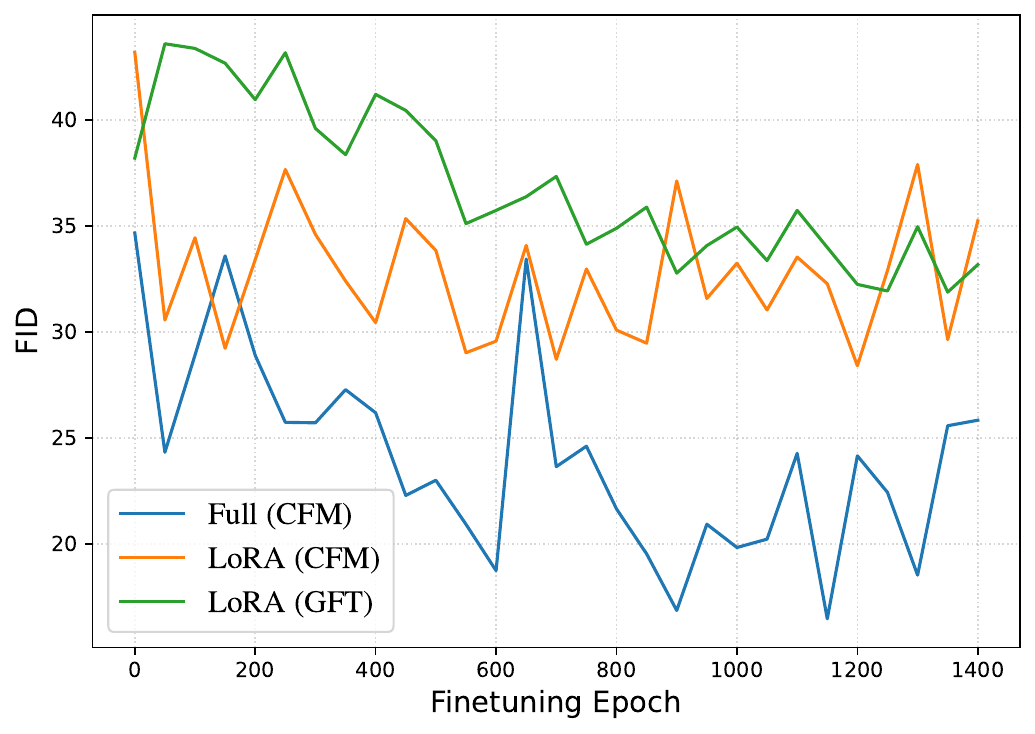} &
\includegraphics[width=0.45\linewidth, valign=m]{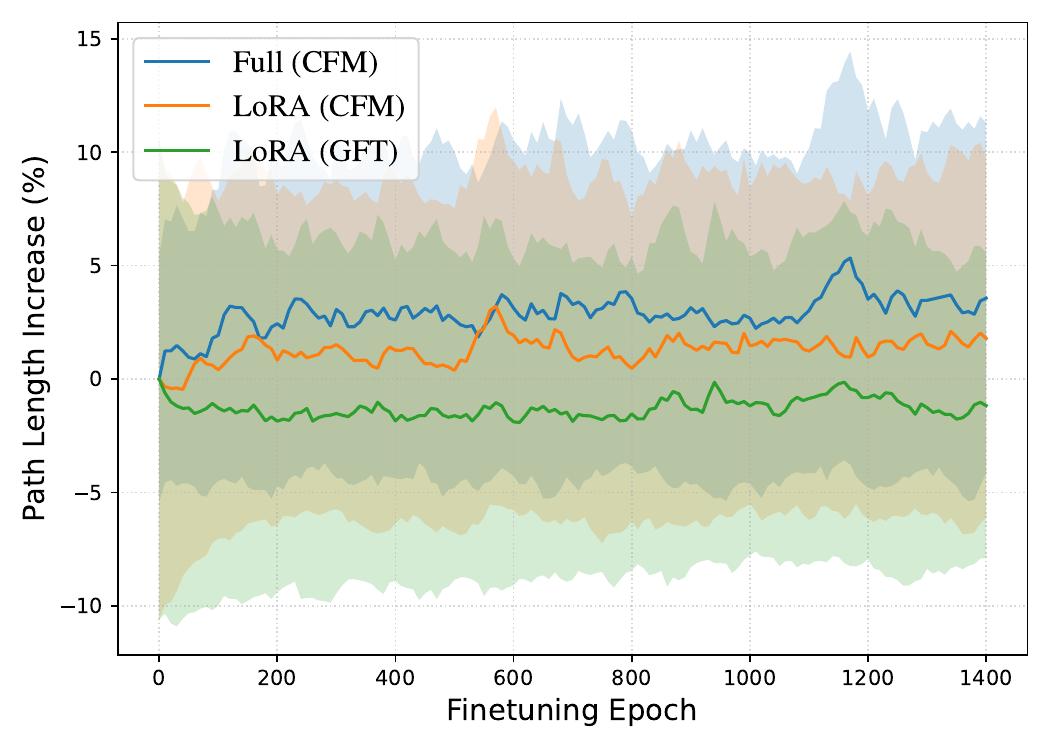} 
\end{tabular}

\caption{
Fine-tuning performance from a fully fine-tuned model to two WILDS validation datasets. Rows corresponding to individual datasets, and the columns show FID and probability path lengths through fine-tuning. The shaded region of the path length graphs show one standard deviation from the mean.
}
\label{fig:small_dist_shift}
\end{figure}

\subsection{Generated Images}\label{ap:images}

\begin{figure}[H]
\centering
\setlength{\tabcolsep}{4pt}
\renewcommand{\arraystretch}{1.1}

\begin{tabular}{c c c c}
    & \textbf{Random Init. (CFM)} & \textbf{Pretrained (CFM)} & \textbf{Pretrained (GFT)} \\

\rotatebox{90}{\textbf{10 Epochs}} &
\includegraphics[width=0.2\linewidth]{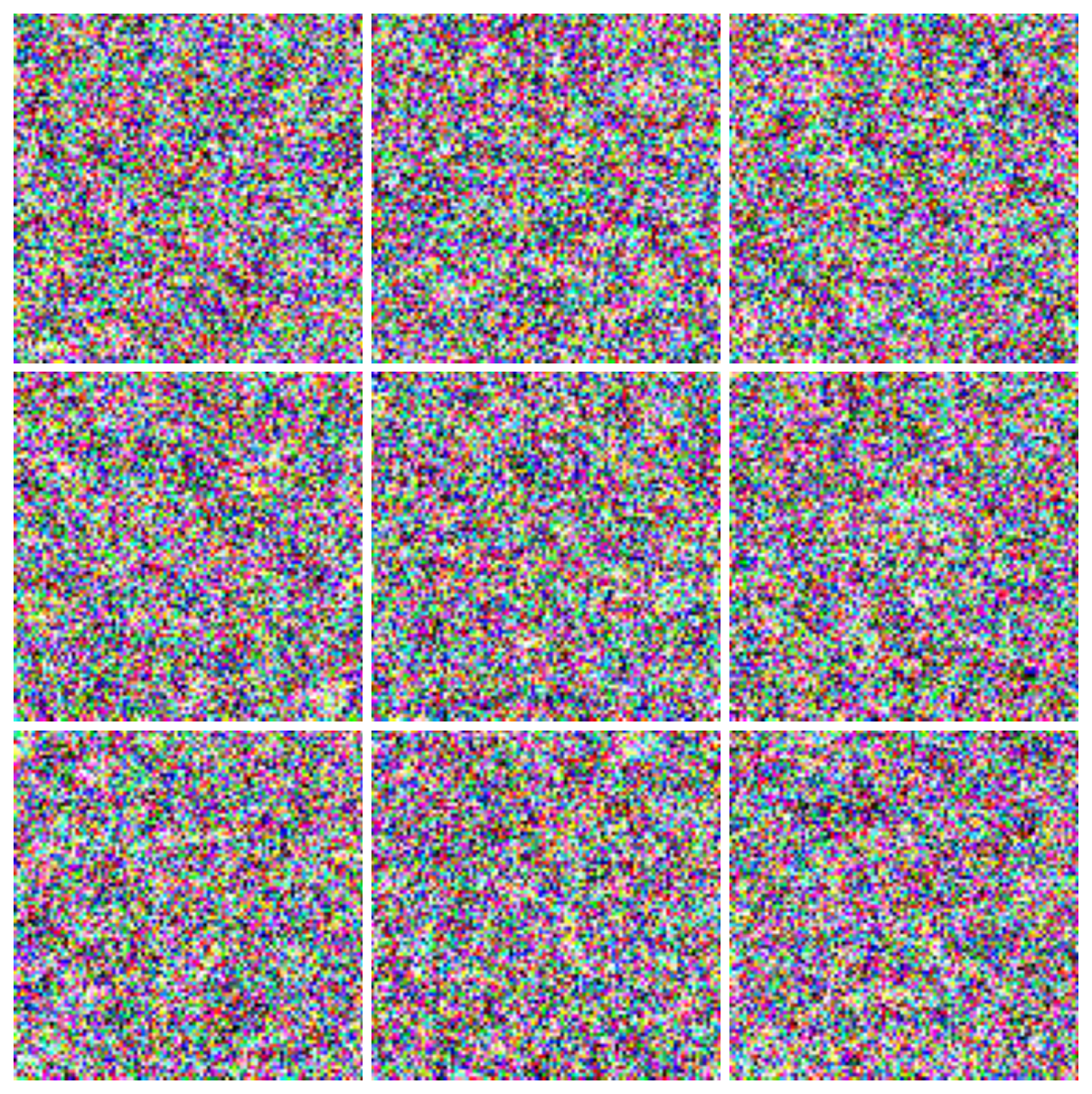} &
\includegraphics[width=0.2\linewidth]{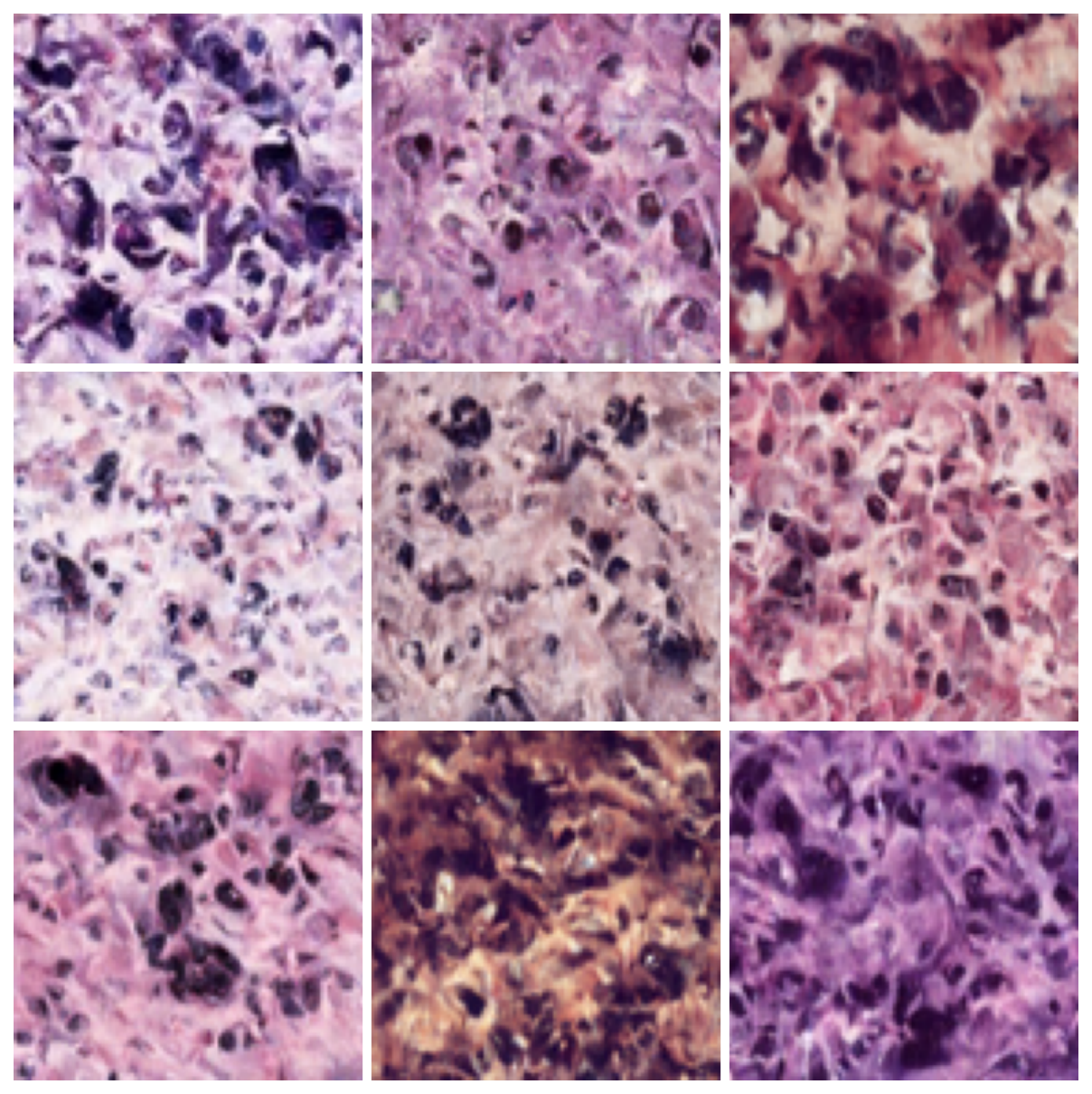} &
\includegraphics[width=0.2\linewidth]{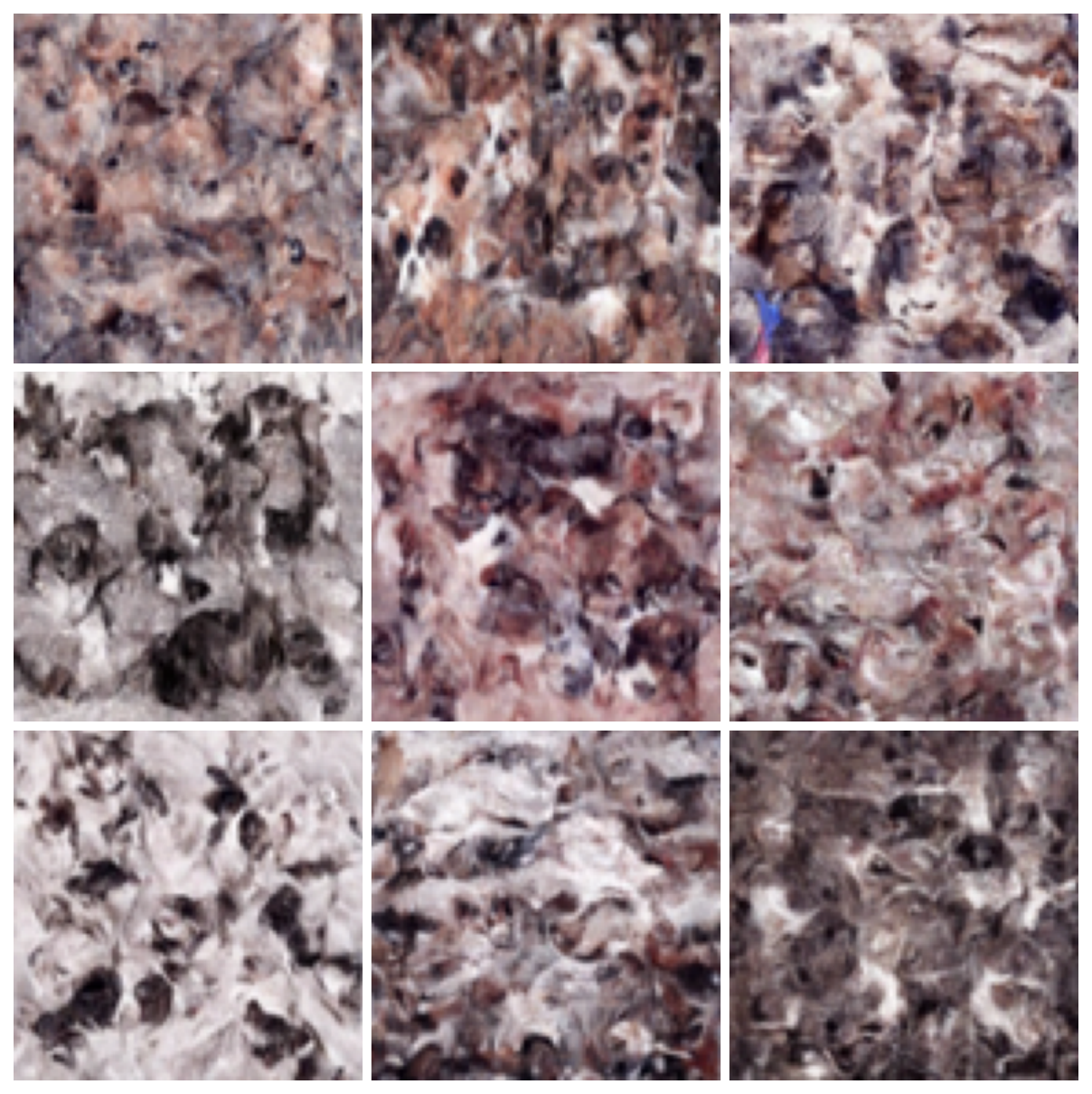} \\

\rotatebox{90}{\textbf{100 Epochs}} &
\includegraphics[width=0.2\linewidth]{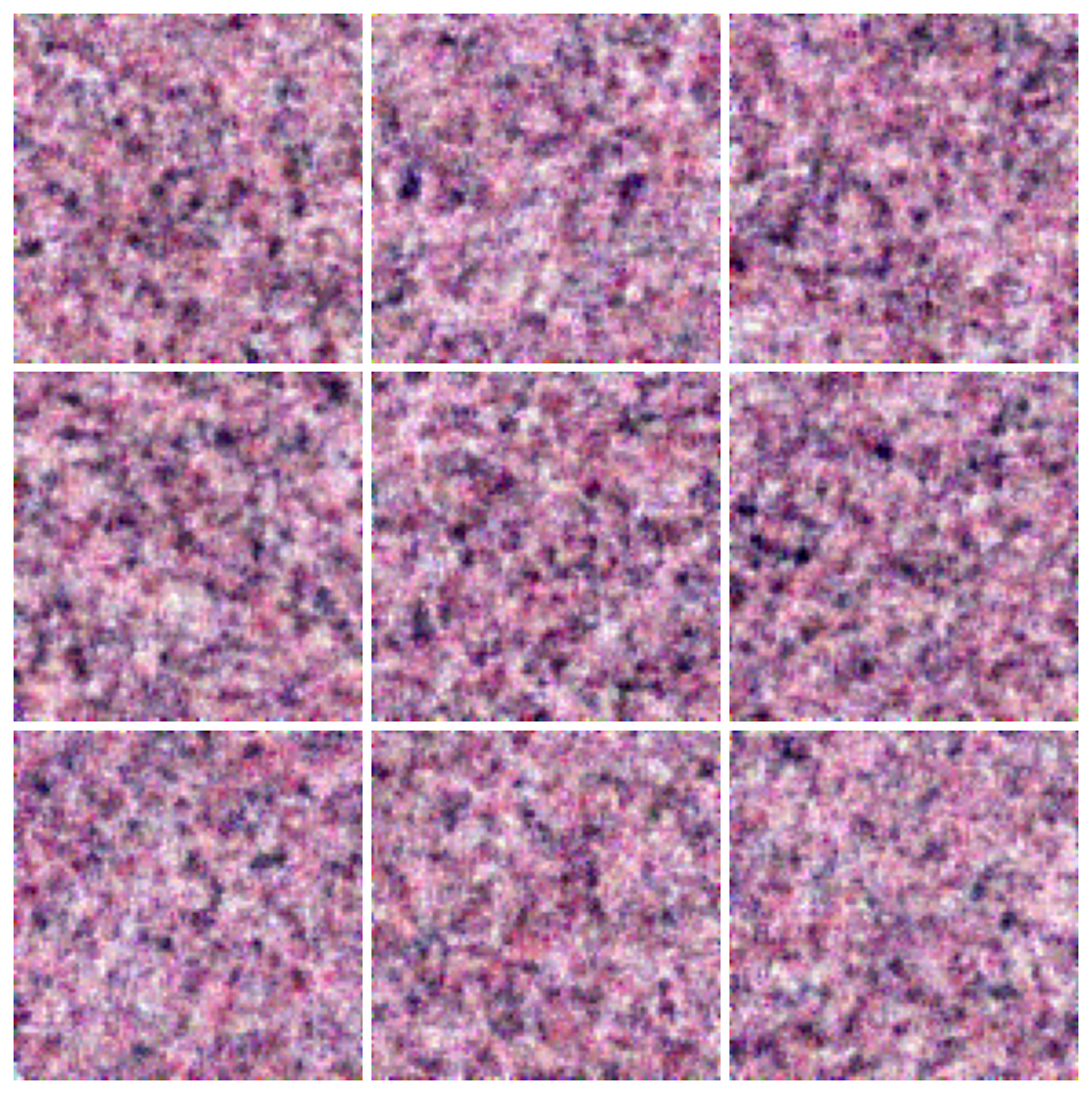} &
\includegraphics[width=0.2\linewidth]{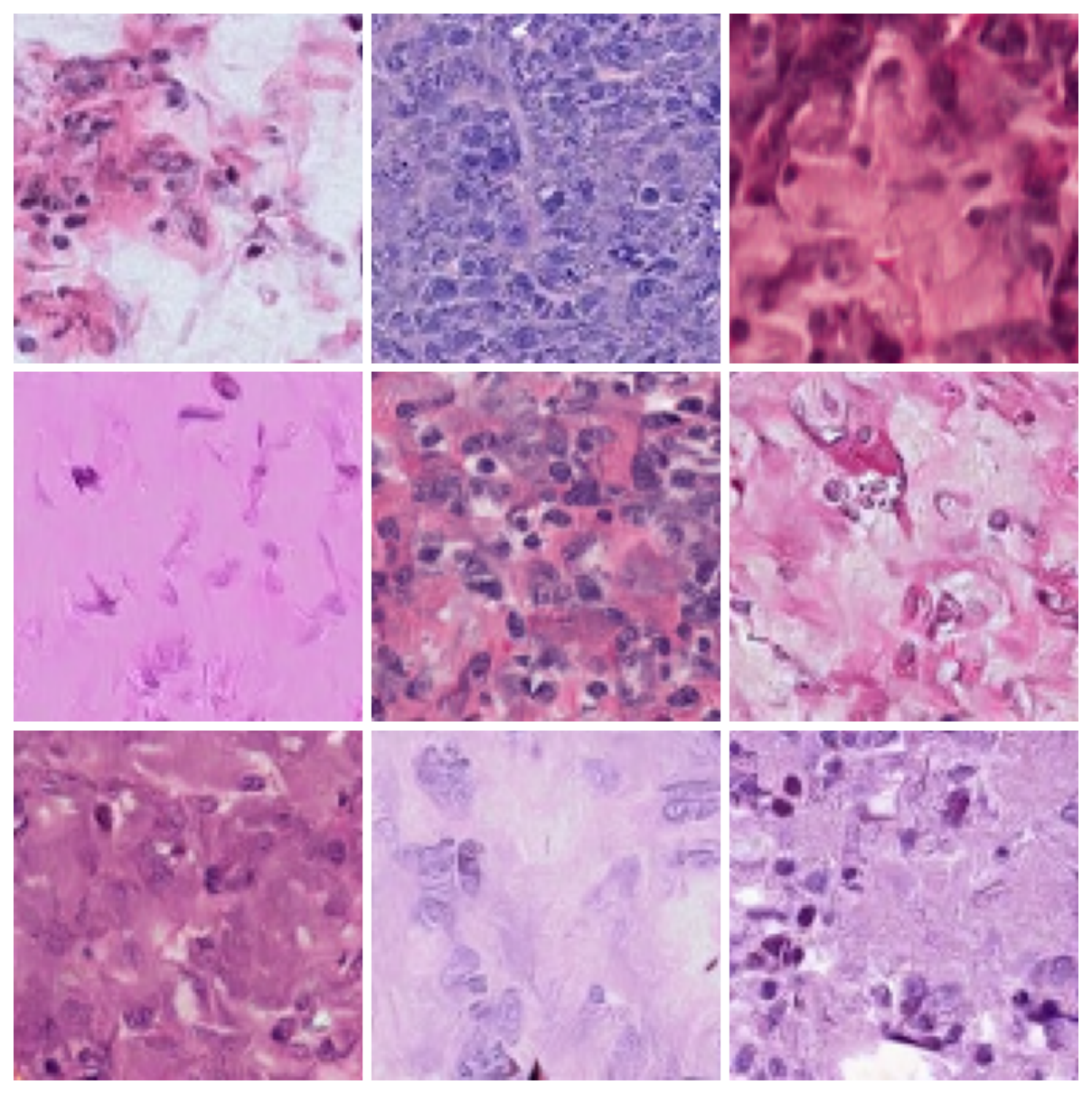} &
\includegraphics[width=0.2\linewidth]{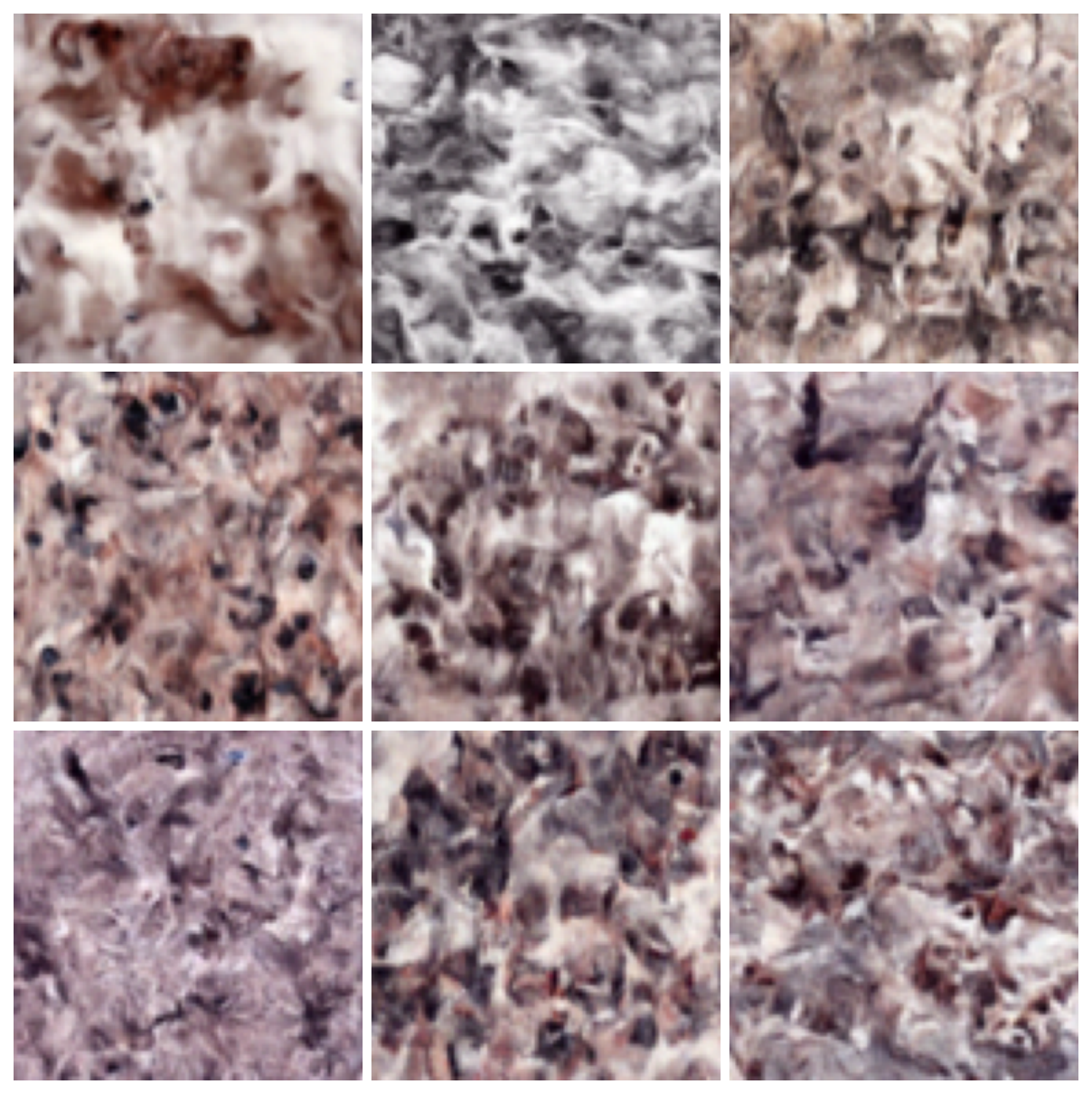} \\

\rotatebox{90}{\textbf{500 Epochs}} &
\includegraphics[width=0.2\linewidth]{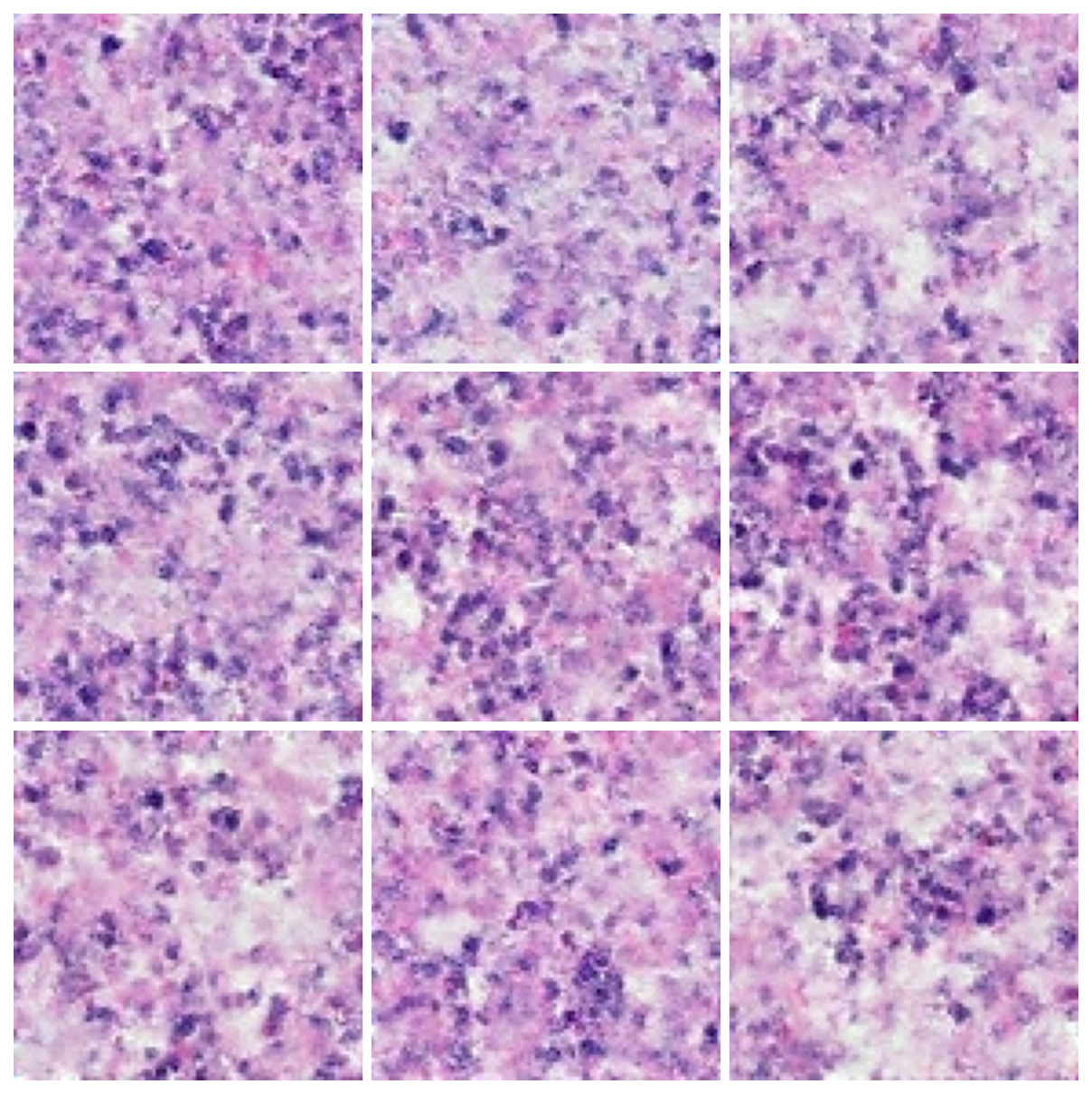} &
\includegraphics[width=0.2\linewidth]{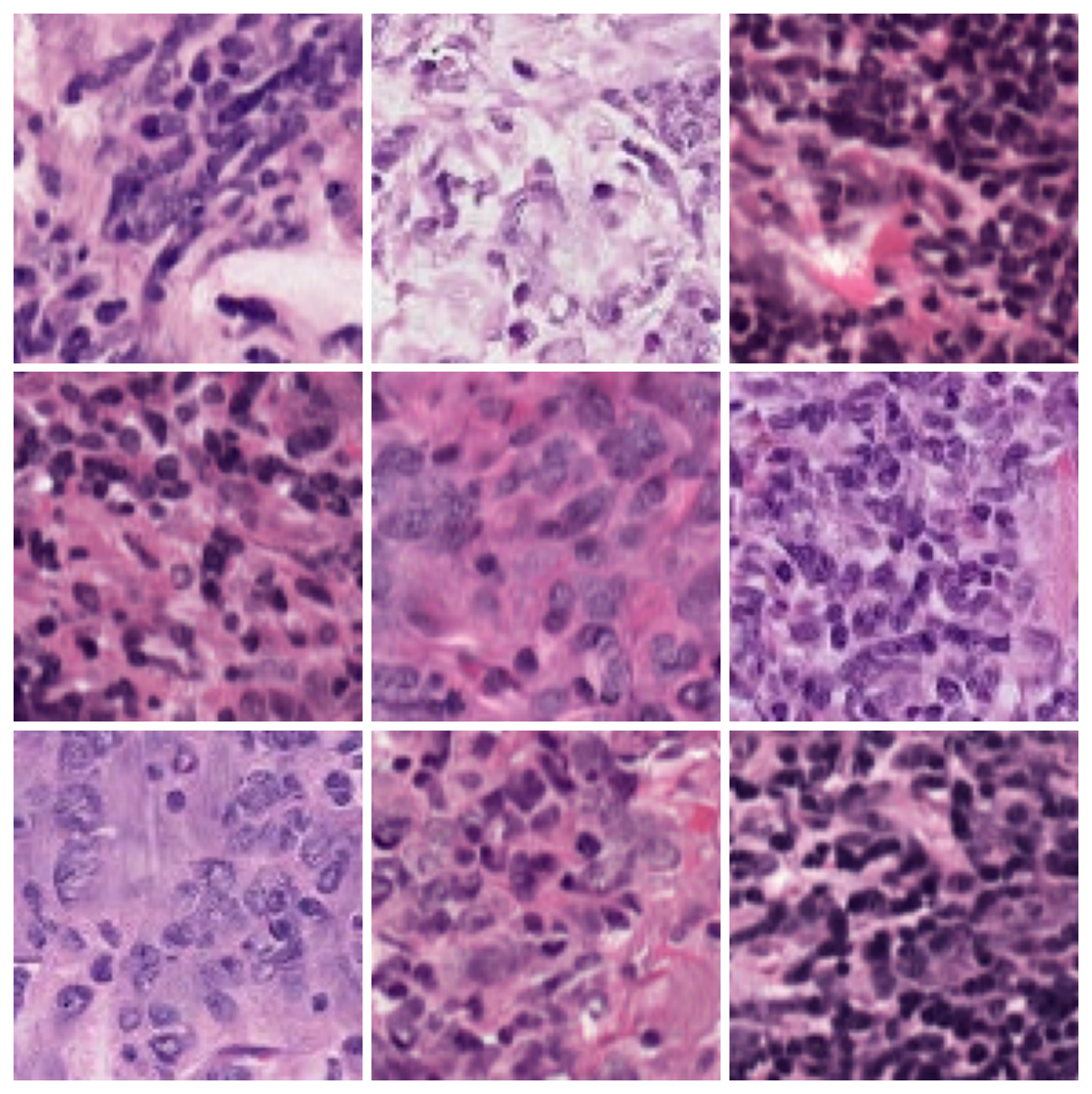} &
\includegraphics[width=0.2\linewidth]{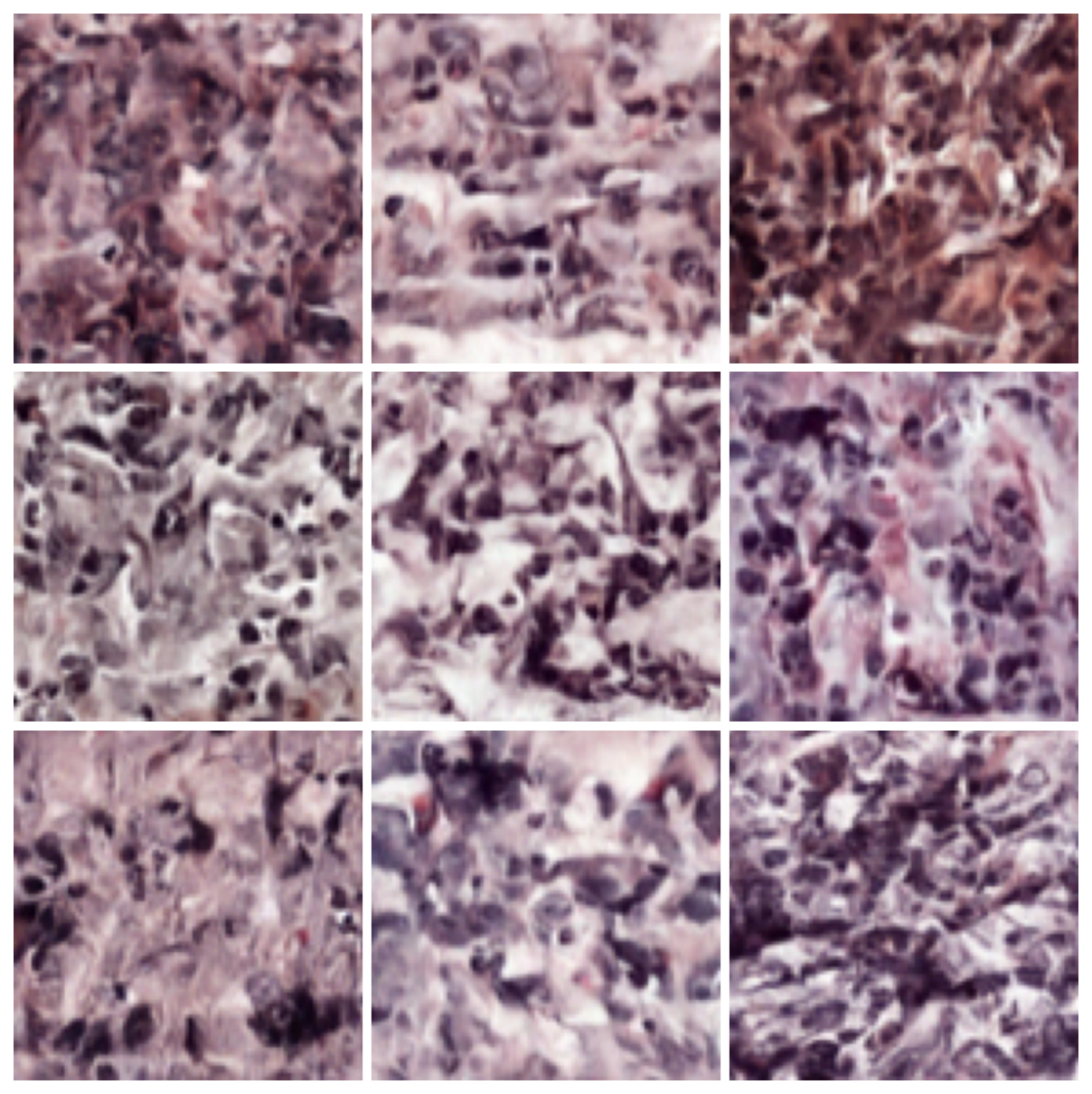} \\

\rotatebox{90}{\textbf{2000 Epochs}} &
\includegraphics[width=0.2\linewidth]{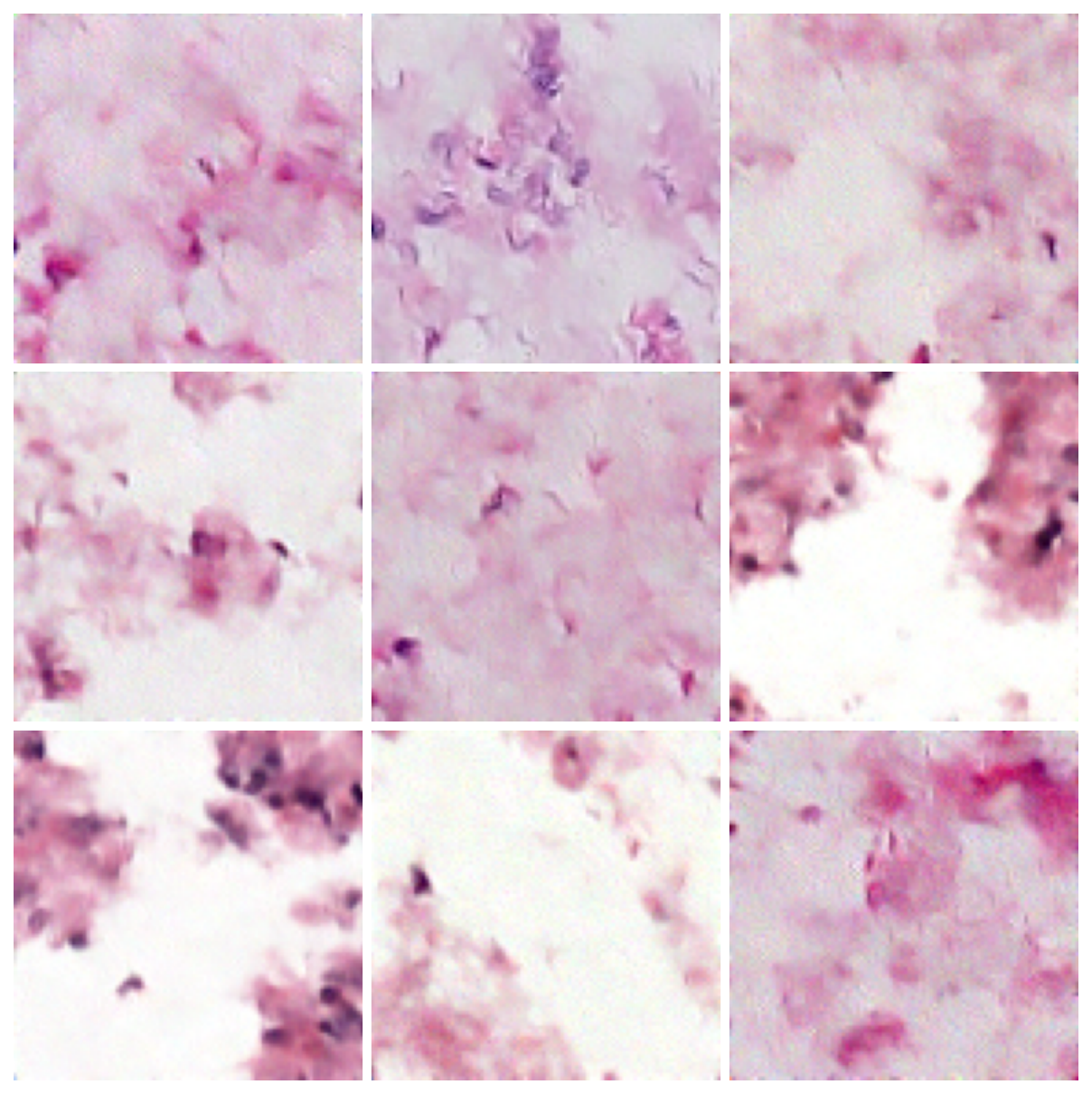} &
\includegraphics[width=0.2\linewidth]{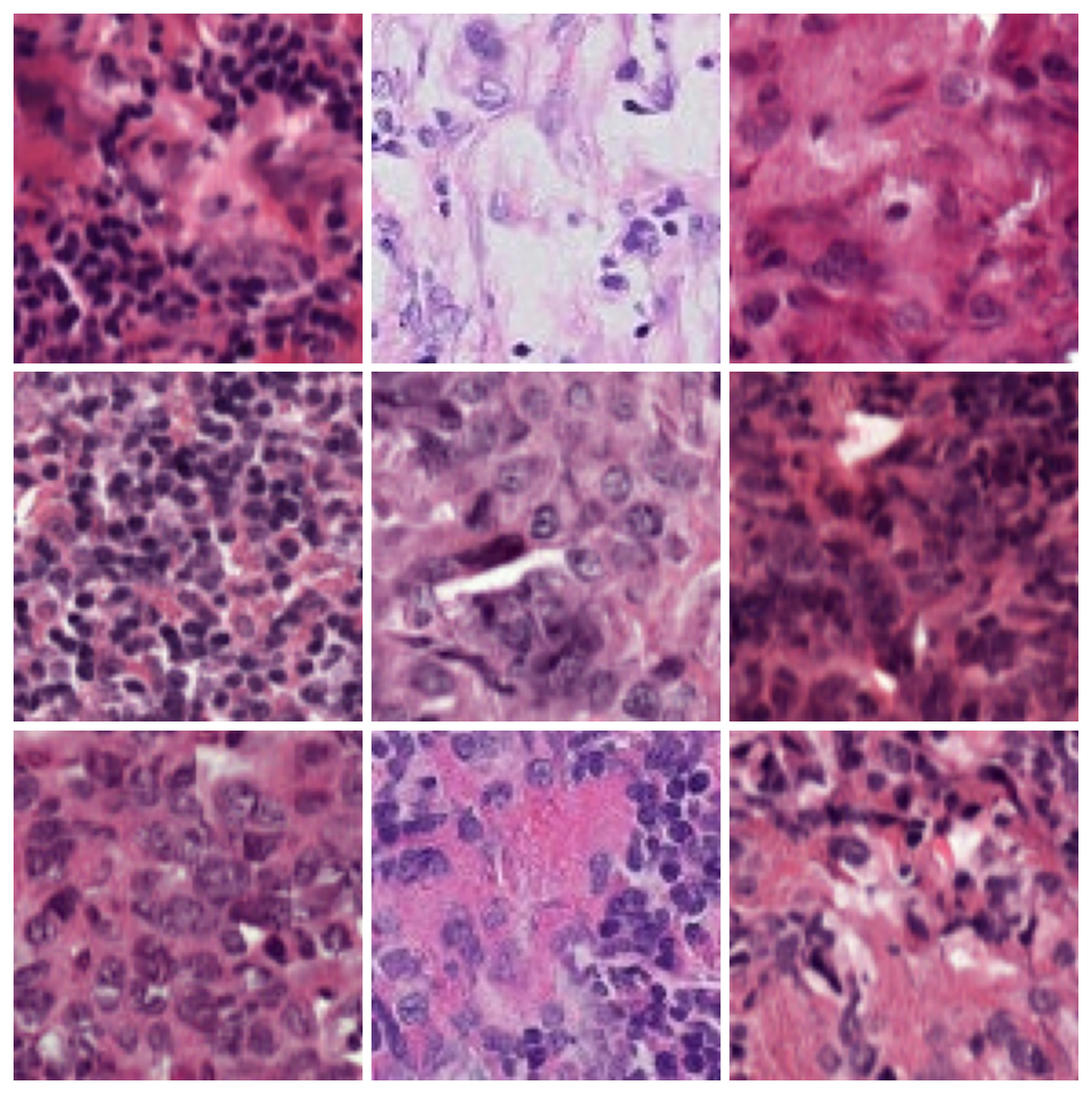} &
\includegraphics[width=0.2\linewidth]{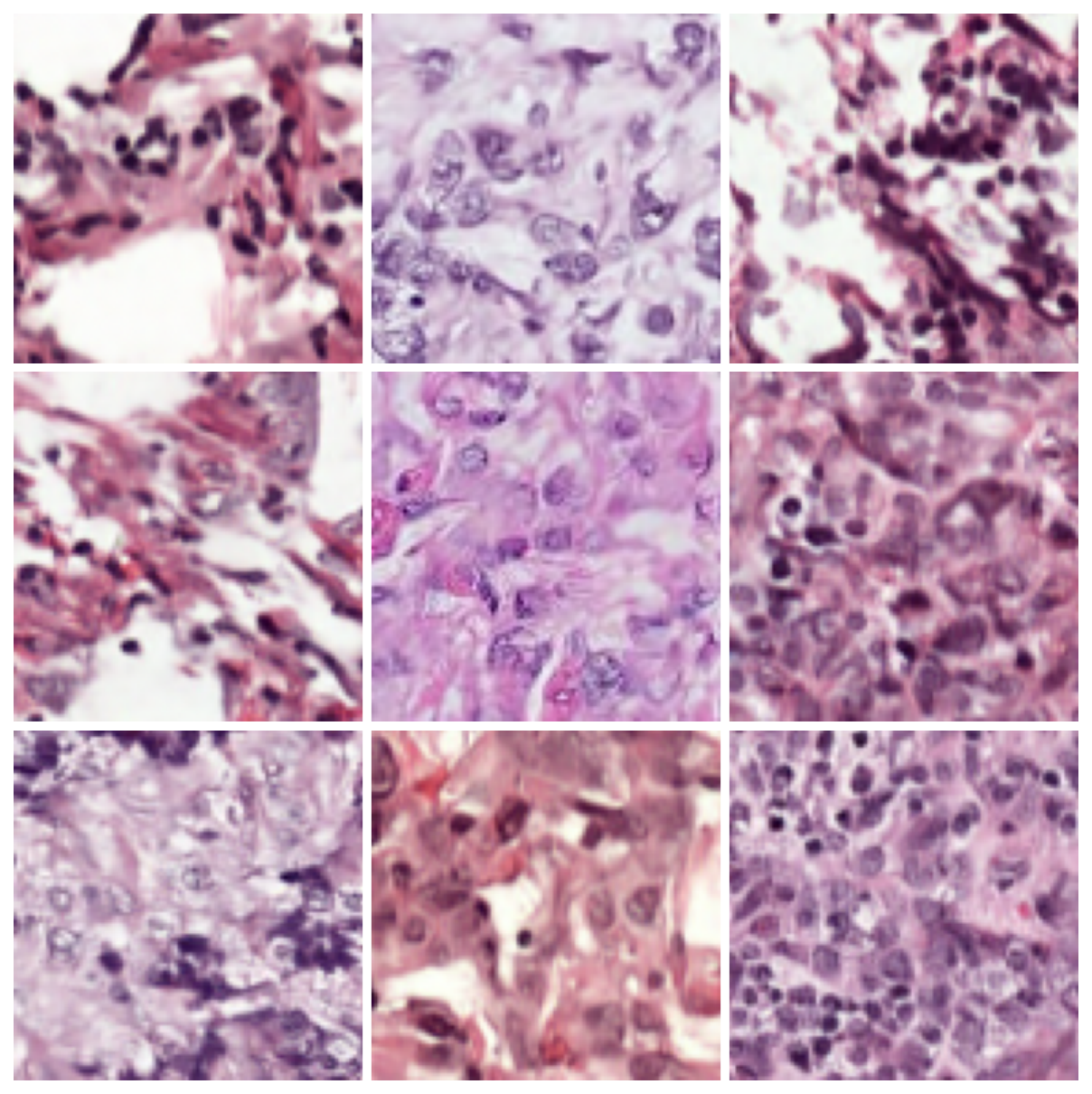}
\end{tabular}
\caption{Generated images at various training stages from models trained or fine-tuned on the 
Camelyon17 train dataset. Rows correspond to the number of training/fine-tuning steps, and columns correspond to training initialization and objective.
}
\label{fig:camelyon_scratch_vs_pre}
\end{figure}

\begin{figure}[H]
\centering
\setlength{\tabcolsep}{4pt}
\renewcommand{\arraystretch}{1.1}

\begin{tabular}{c c c c}
    & \textbf{Ground Truth} & \textbf{Full (CFM)} & \textbf{Full (GFT)}\\

\rotatebox{90}{\textbf{Camelyon17}} &
\includegraphics[width=0.3\linewidth]{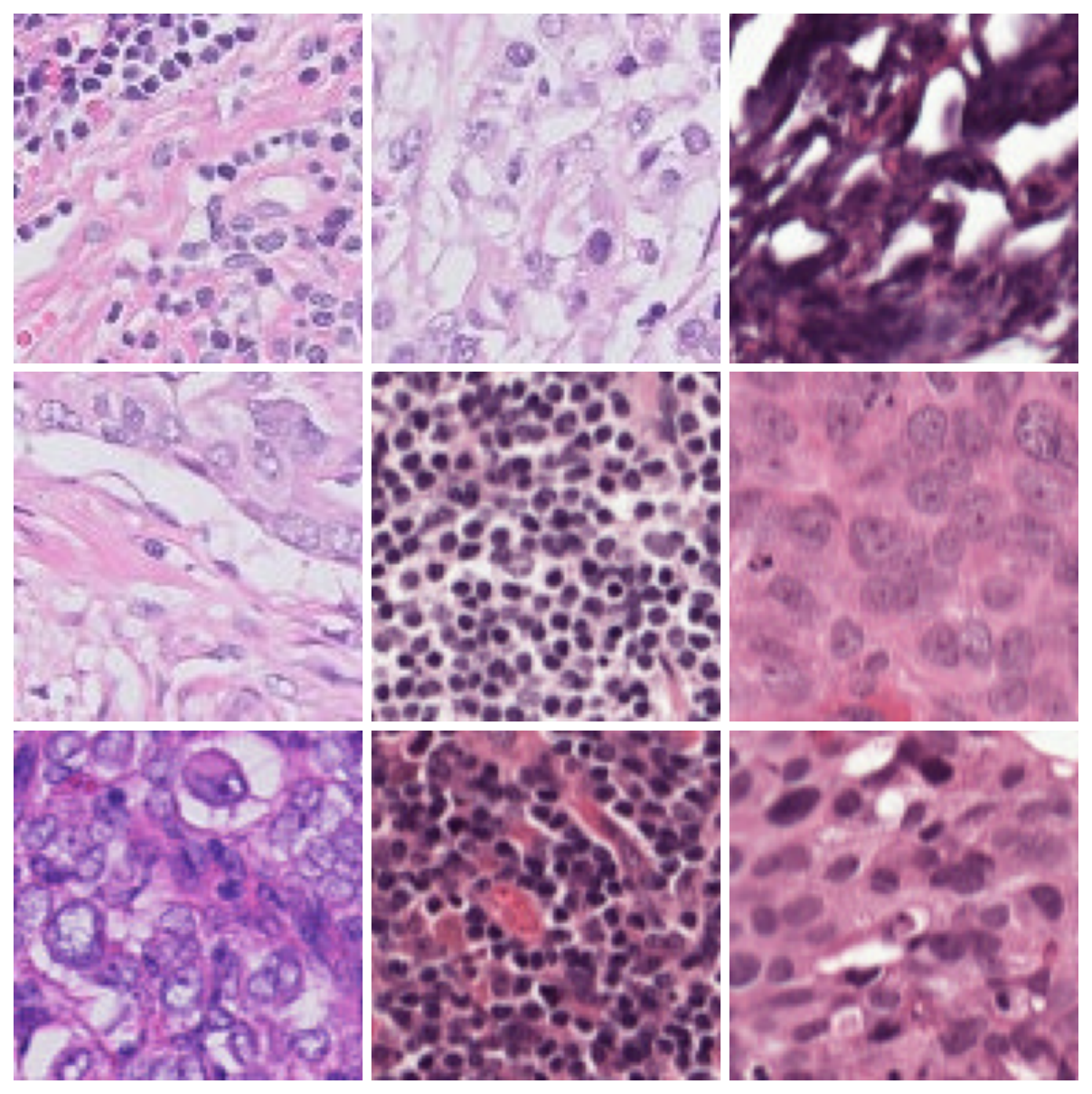} &
\includegraphics[width=0.3\linewidth]{full_Camelyon17_train.pdf} &
\includegraphics[width=0.3\linewidth]{full_L5_Camelyon17_train.pdf} \\

\rotatebox{90}{\textbf{RxRx1}} &
\includegraphics[width=0.3\linewidth]{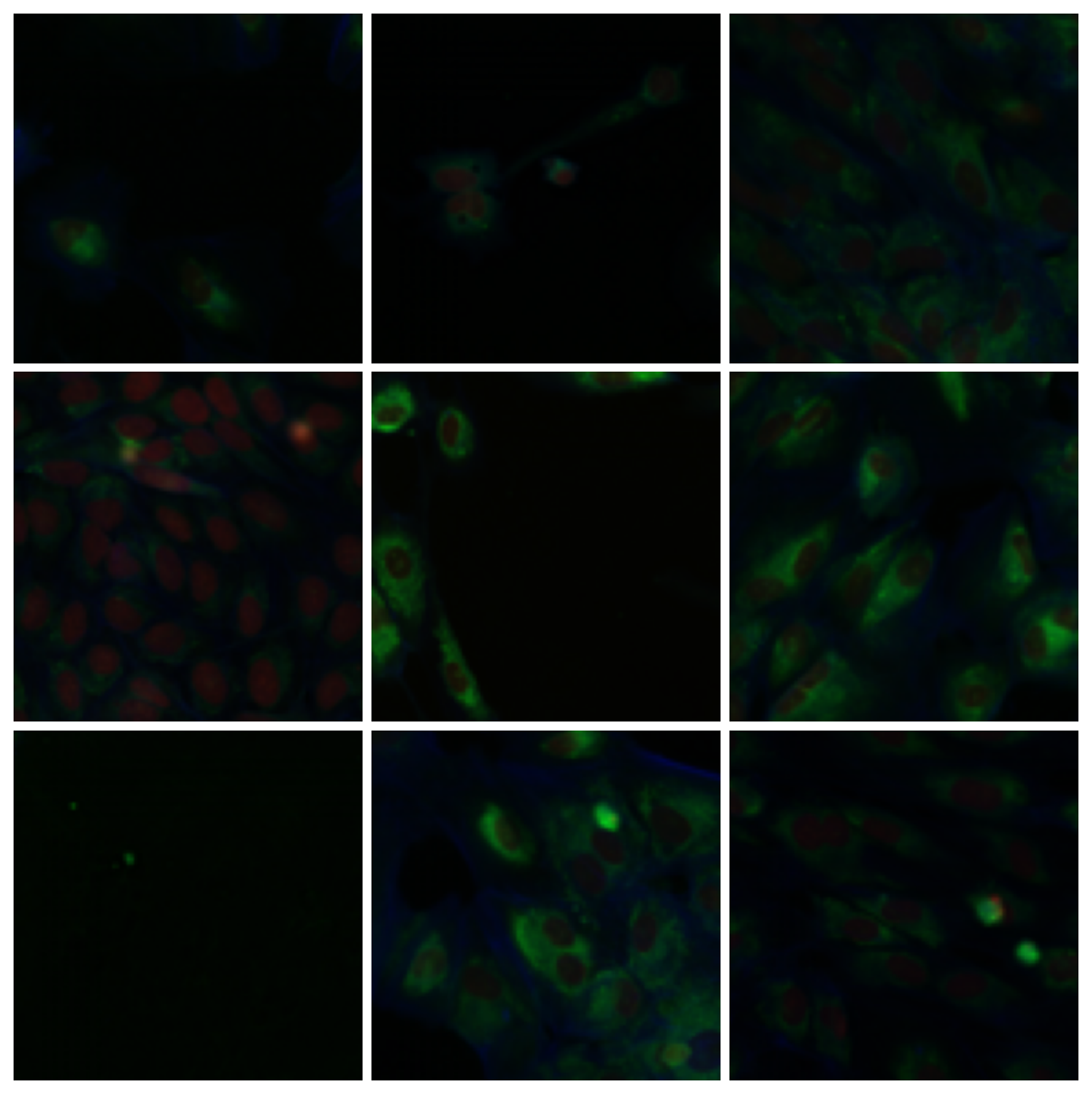} &
\includegraphics[width=0.3\linewidth]{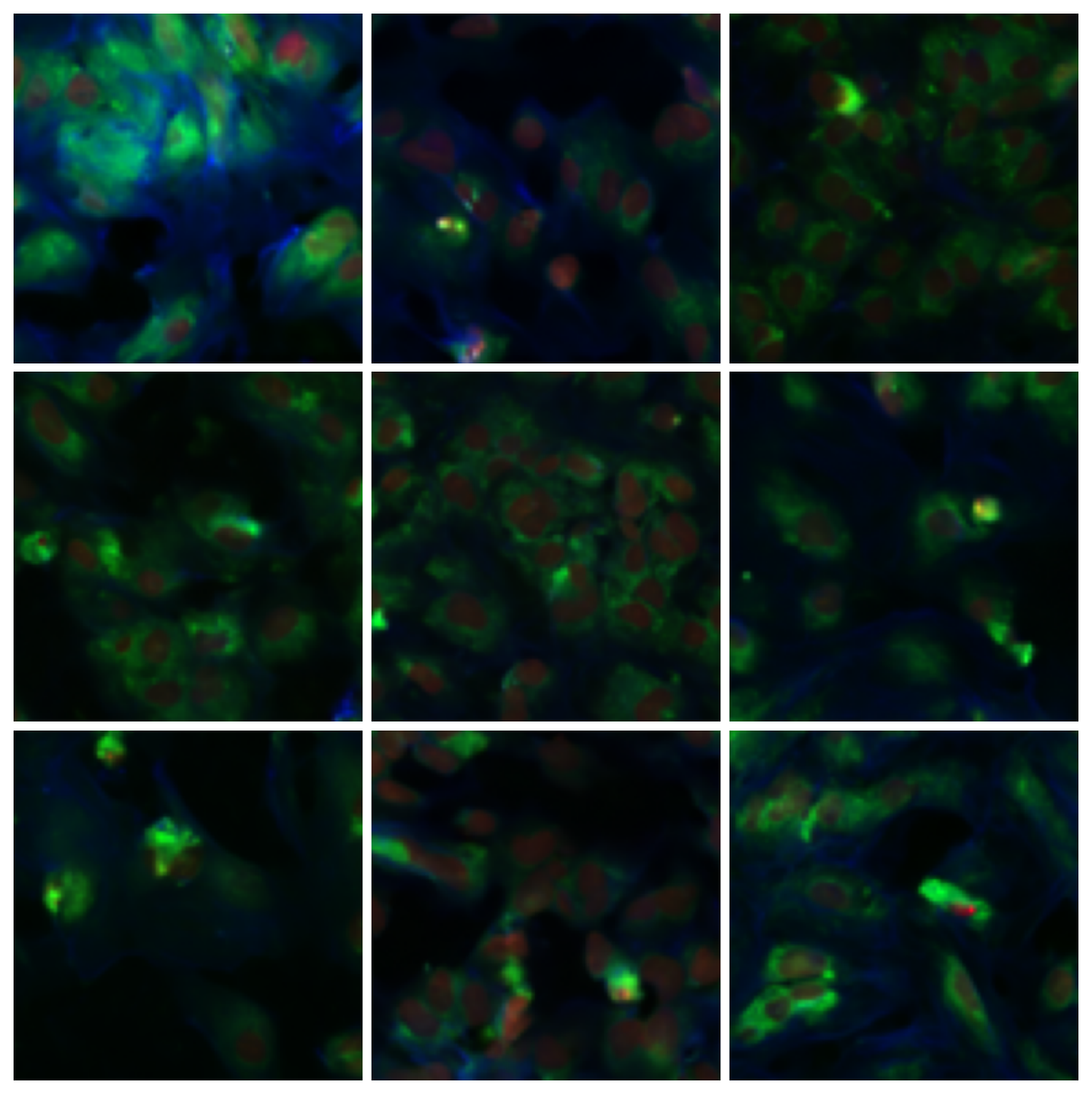} &
\includegraphics[width=0.3\linewidth]{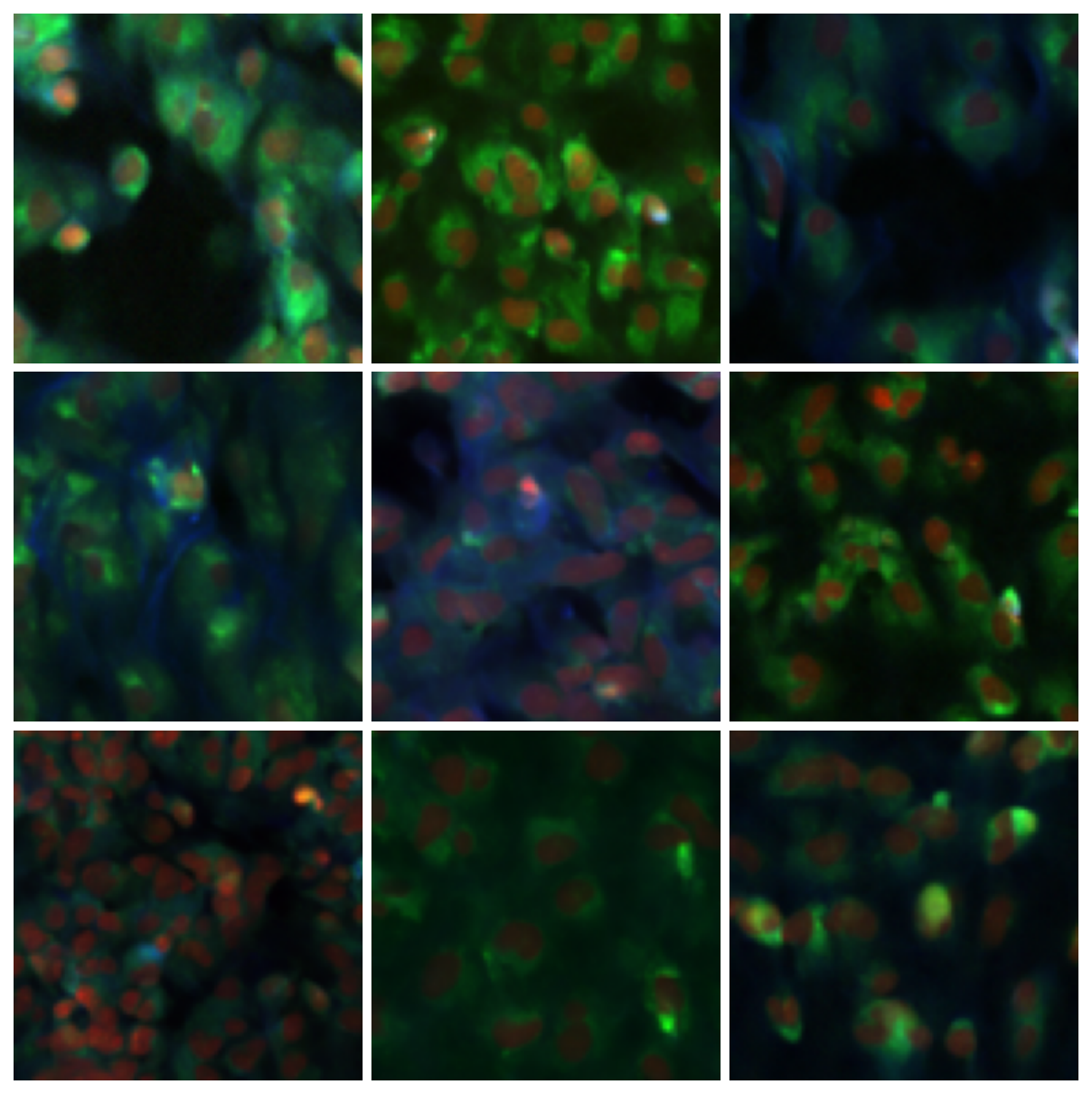} \\

\rotatebox{90}{\textbf{FMoW}} &
\includegraphics[width=0.3\linewidth]{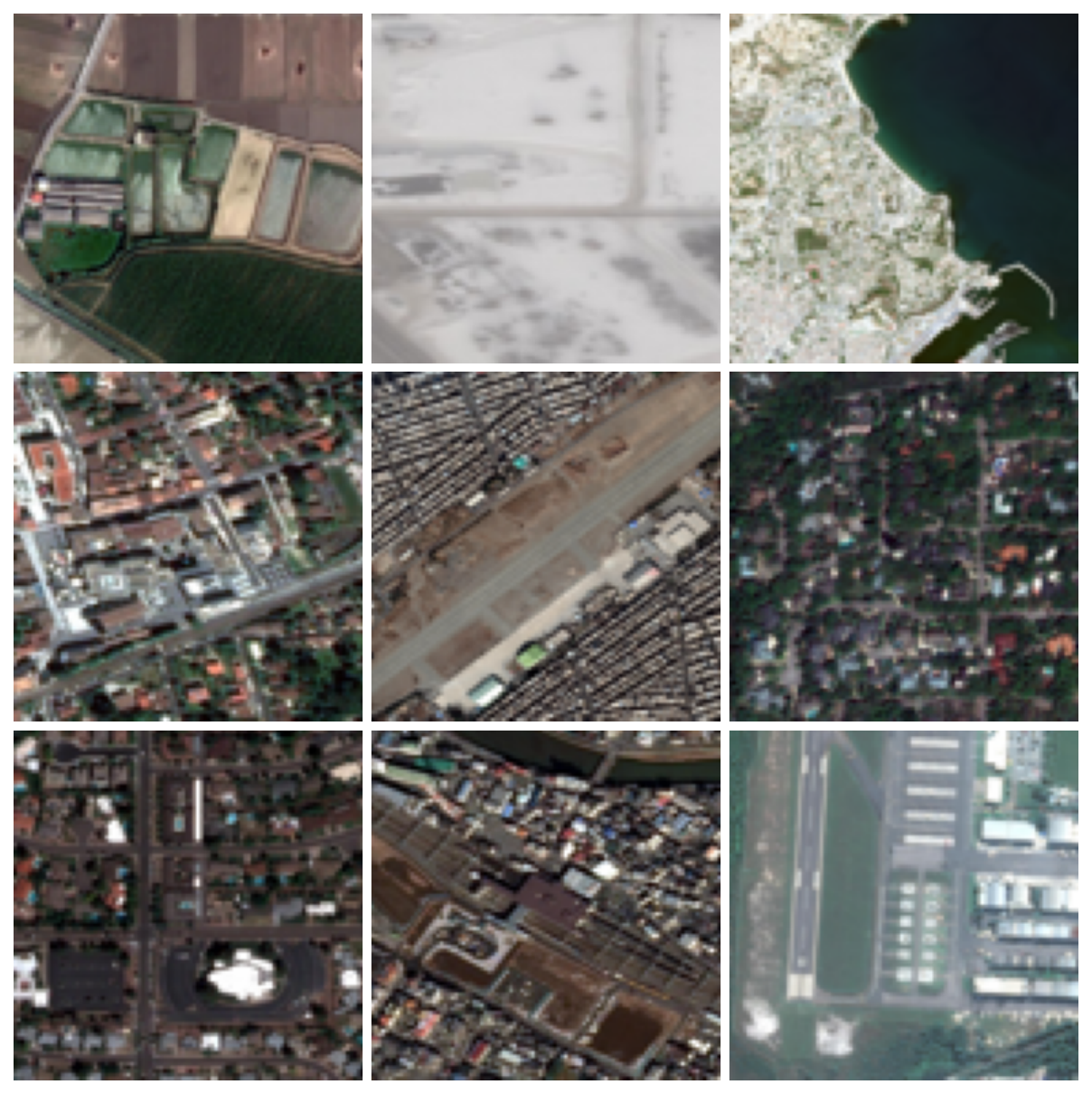} &
\includegraphics[width=0.3\linewidth]{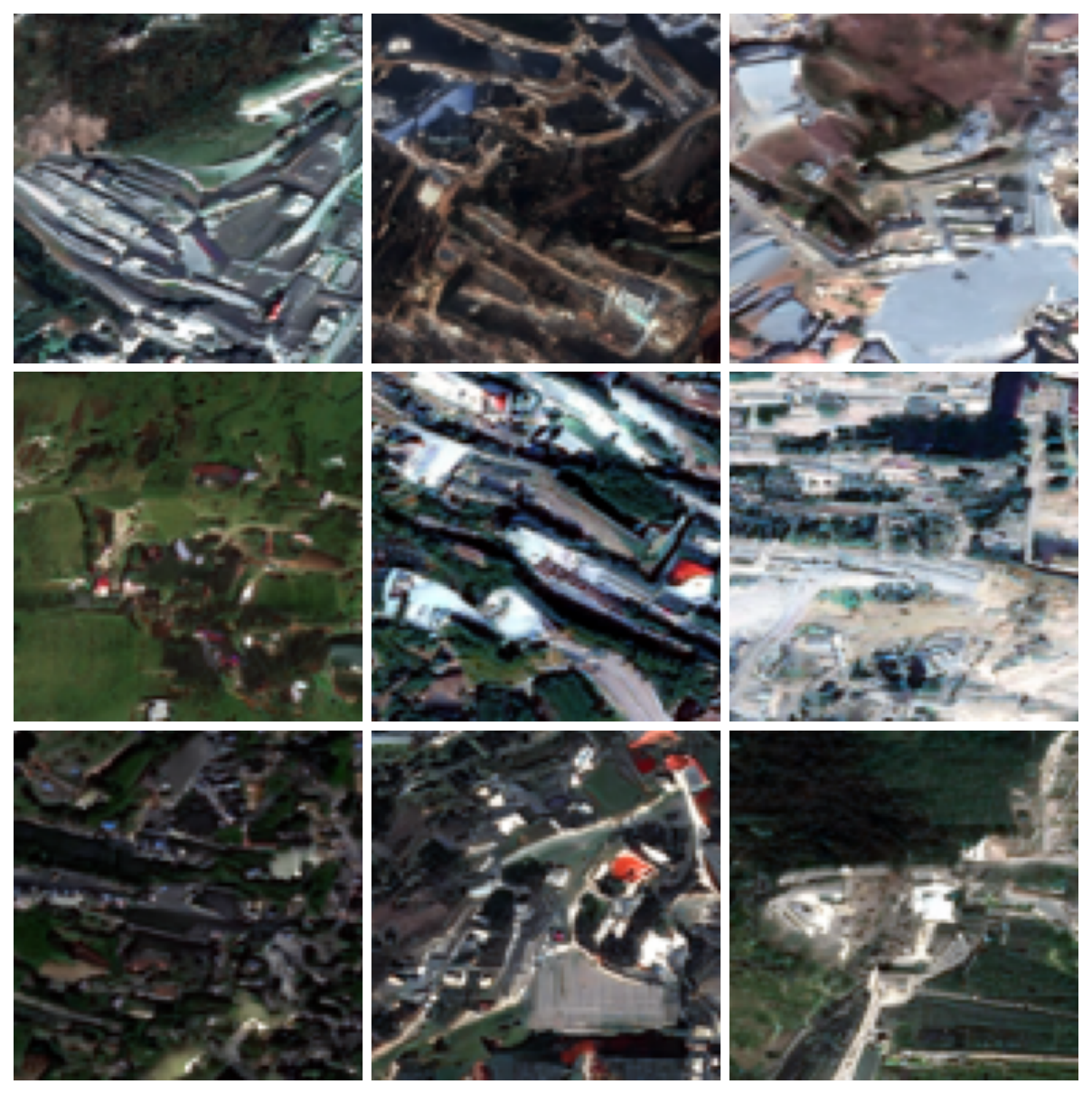} &
\includegraphics[width=0.3\linewidth]{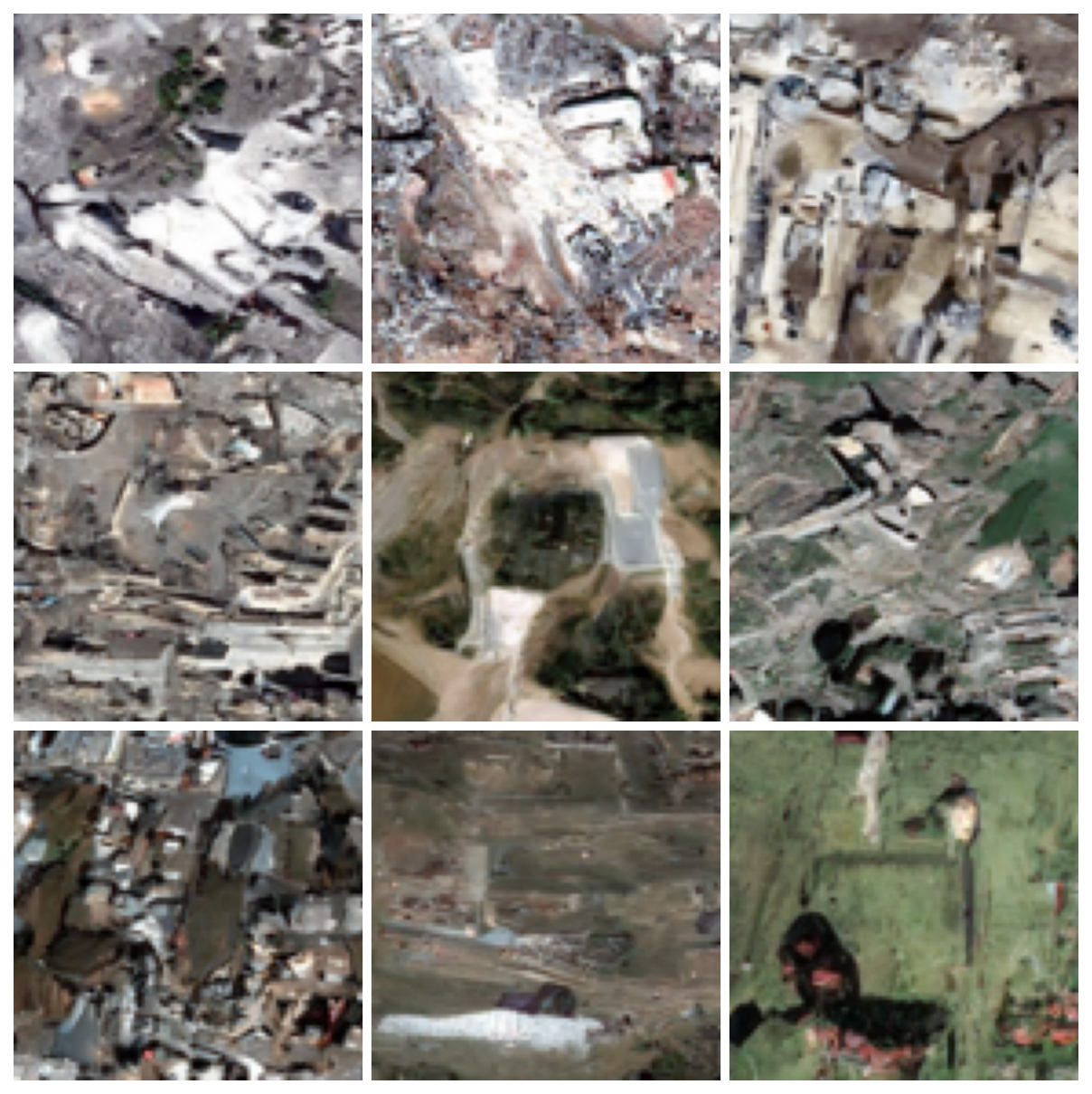}
\end{tabular}

\caption{
Generated images after 2000 epochs of fine-tuning on three WILDS datasets, starting from a pretrained Cifar-10 model.
}
\label{fig:large_dist_shift_images}
\end{figure}

\begin{figure}[H]
\centering
\setlength{\tabcolsep}{4pt}
\renewcommand{\arraystretch}{1.1}

\begin{tabular}{c c c c}
    & \textbf{Ground Truth} & \textbf{Full (CFM)} & \textbf{LoRA (GFT)}\\

\rotatebox{90}{\textbf{Camelyon17}} &
\includegraphics[width=0.3\linewidth]{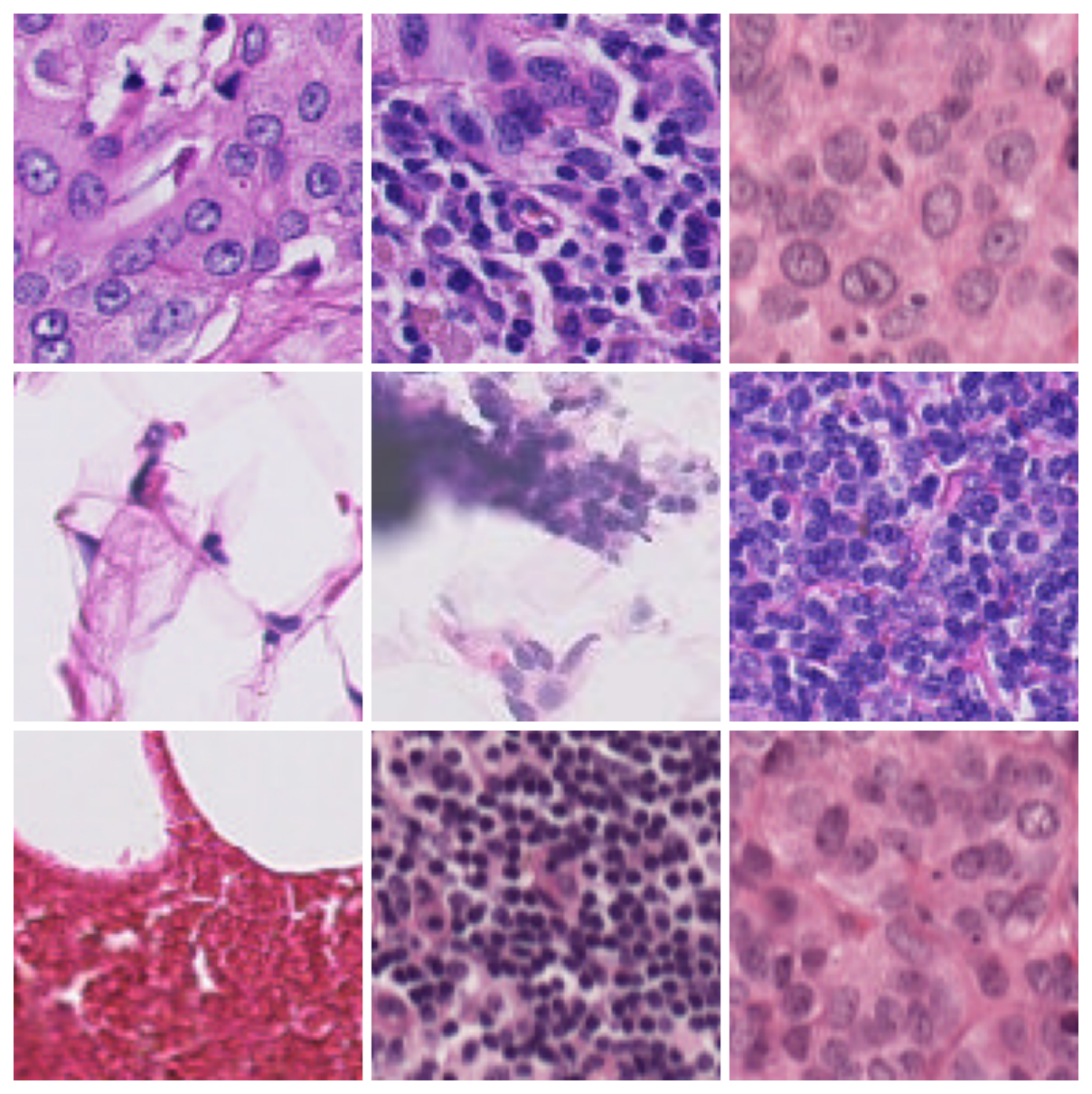} &
\includegraphics[width=0.3\linewidth]{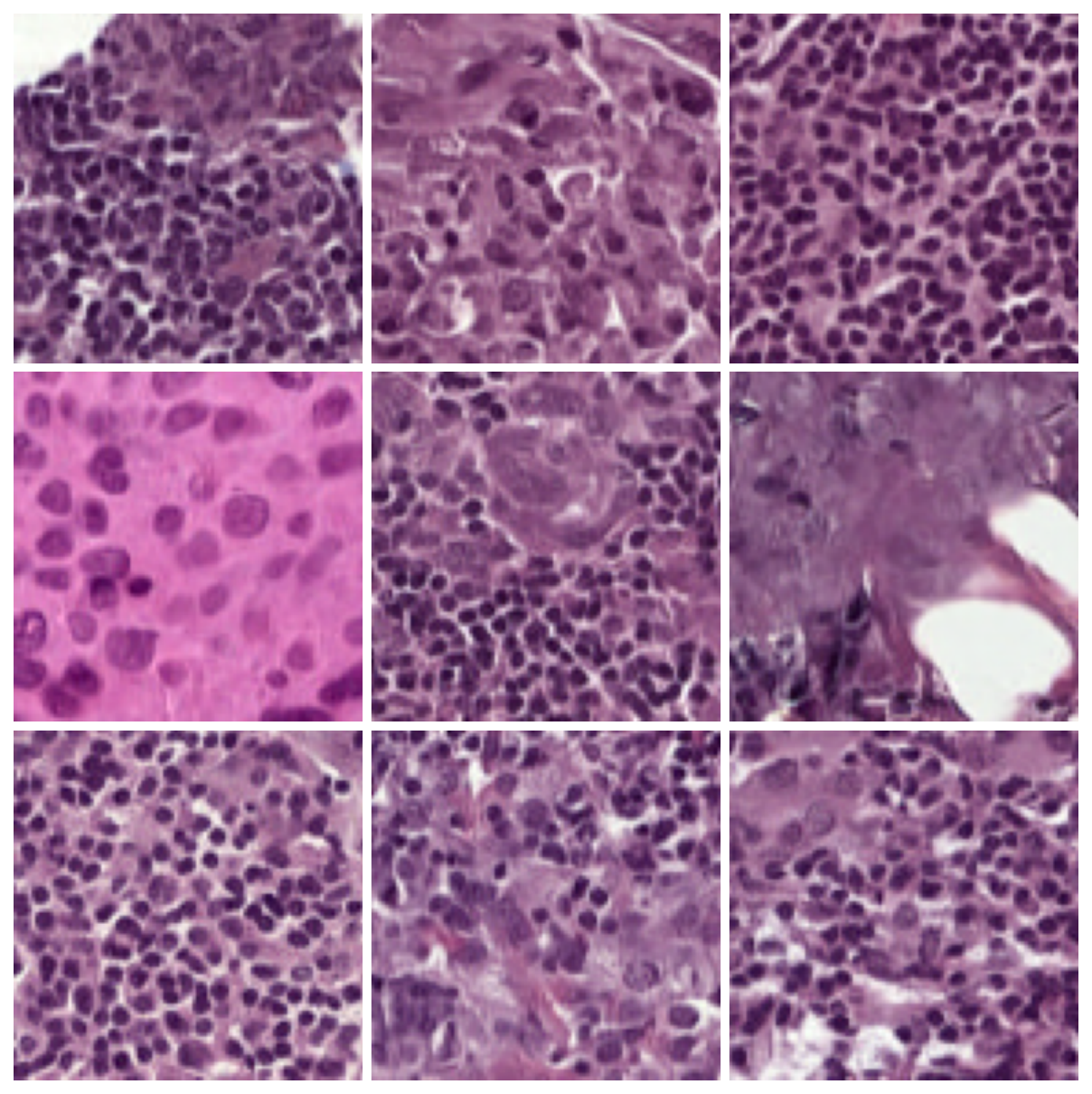} &
\includegraphics[width=0.3\linewidth]{lora_L5_Camelyon17_val.pdf} \\

\rotatebox{90}{\textbf{RxRx1}} &
\includegraphics[width=0.3\linewidth]{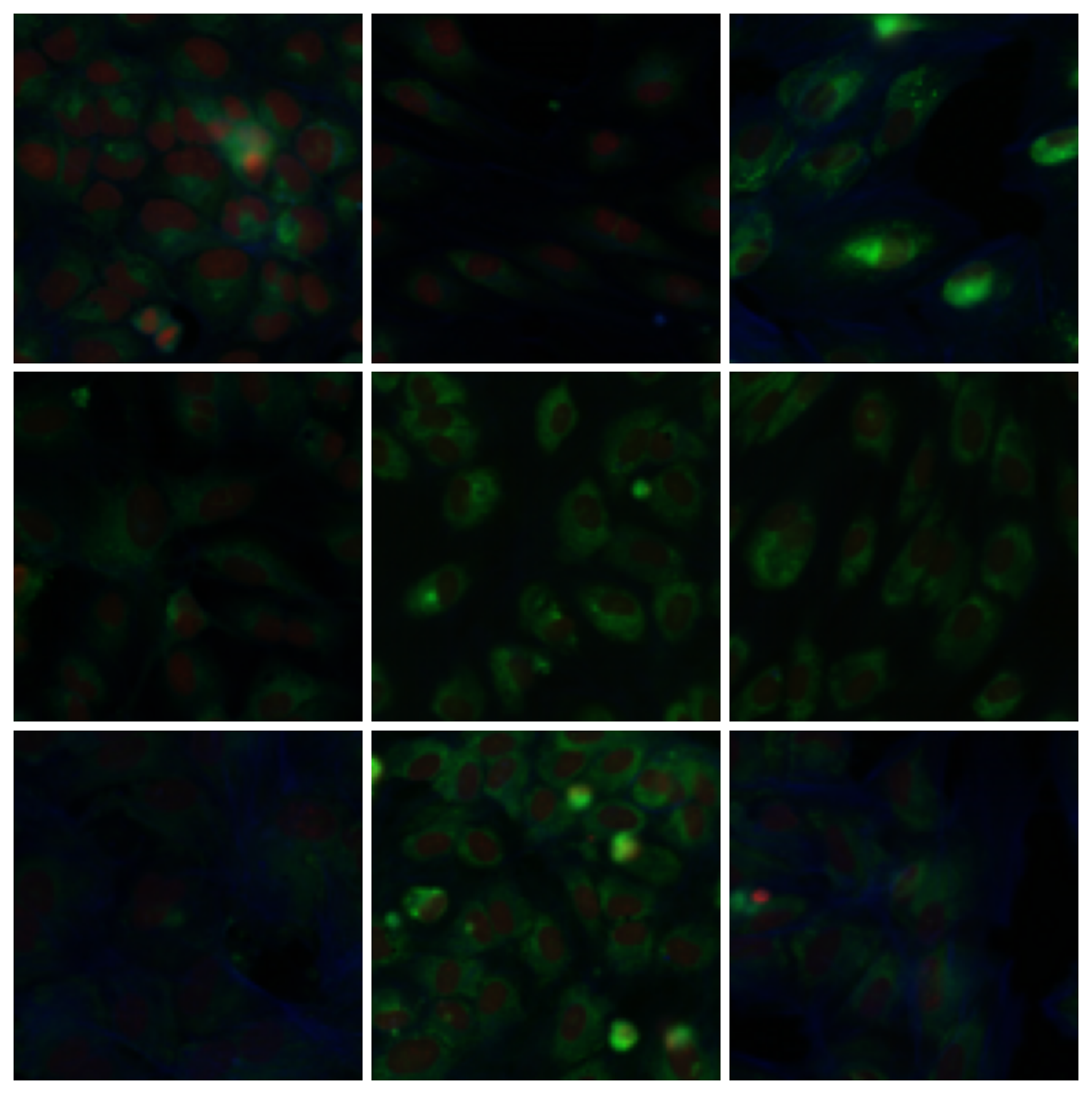} &
\includegraphics[width=0.3\linewidth]{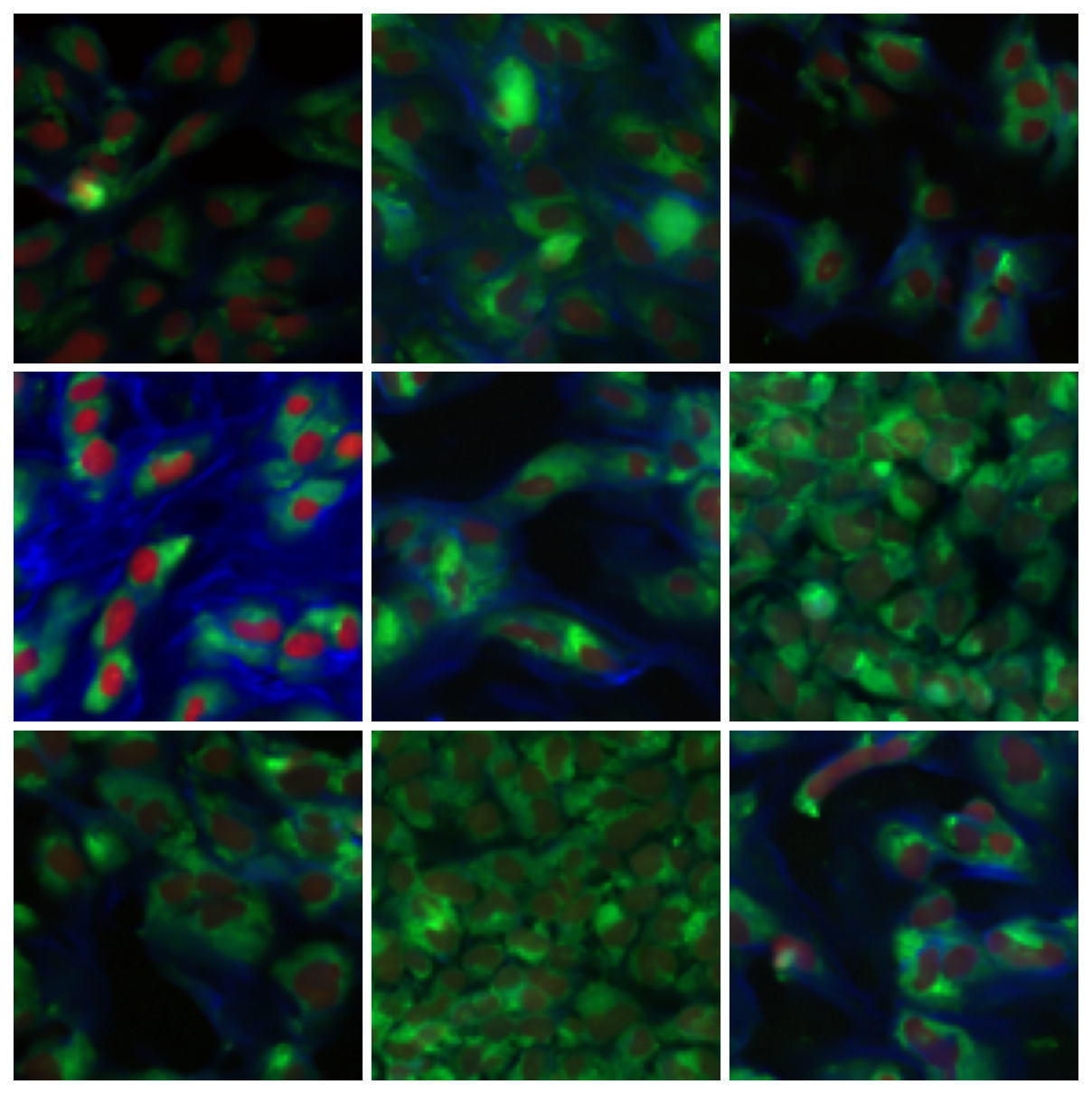} &
\includegraphics[width=0.3\linewidth]{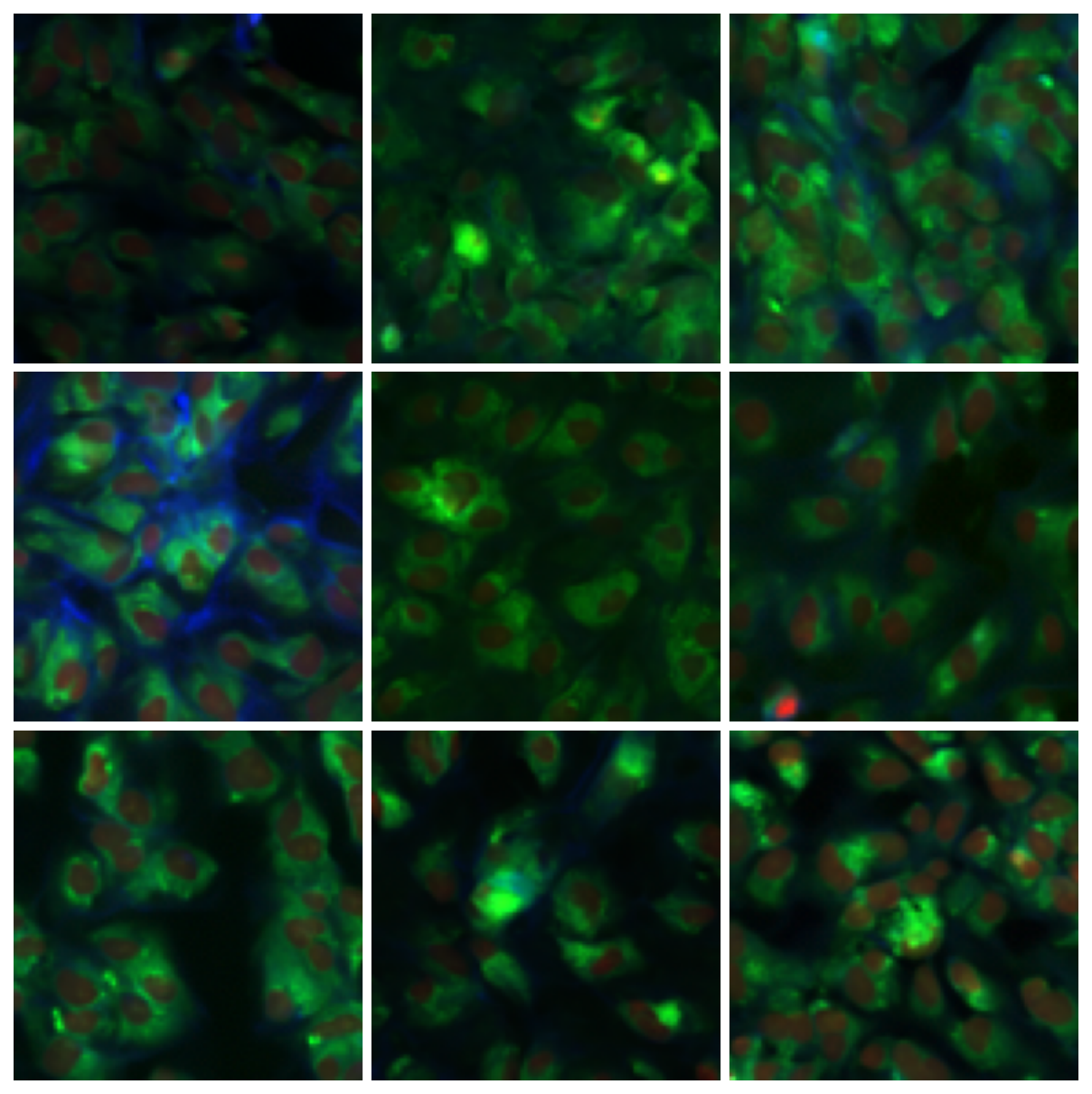} \\

\rotatebox{90}{\textbf{FMoW}} &
\includegraphics[width=0.3\linewidth]{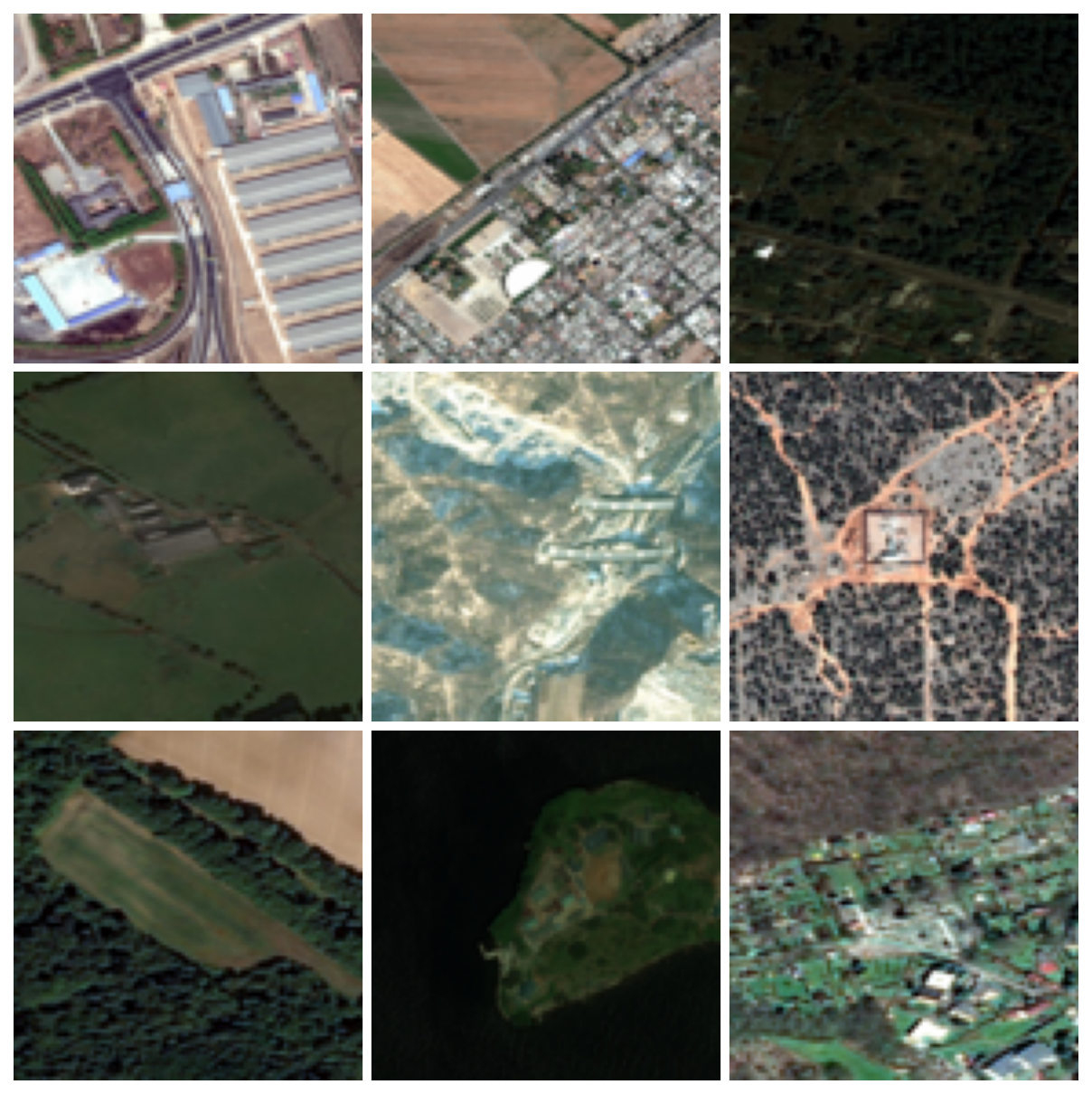} &
\includegraphics[width=0.3\linewidth]{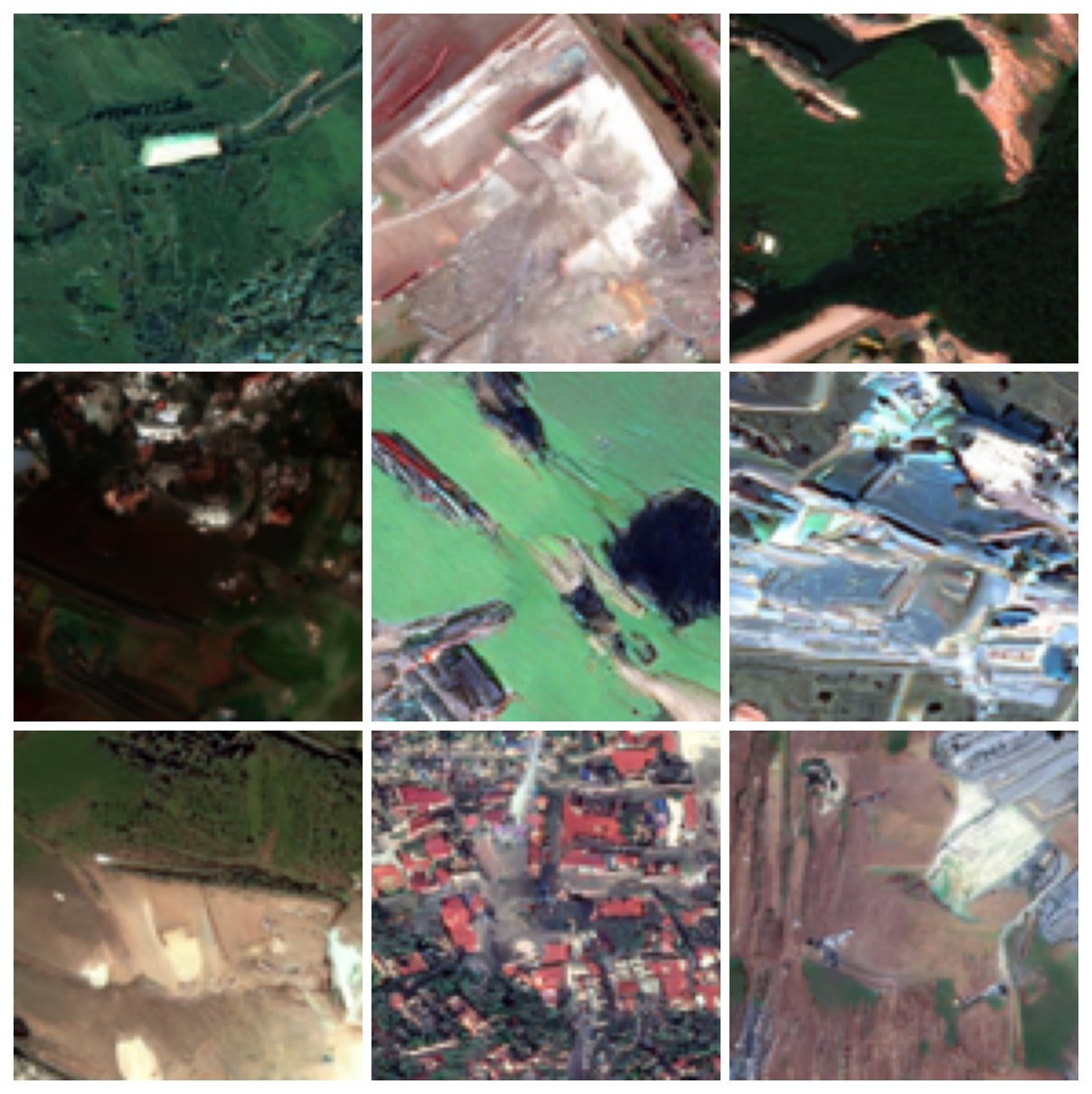} &
\includegraphics[width=0.3\linewidth]{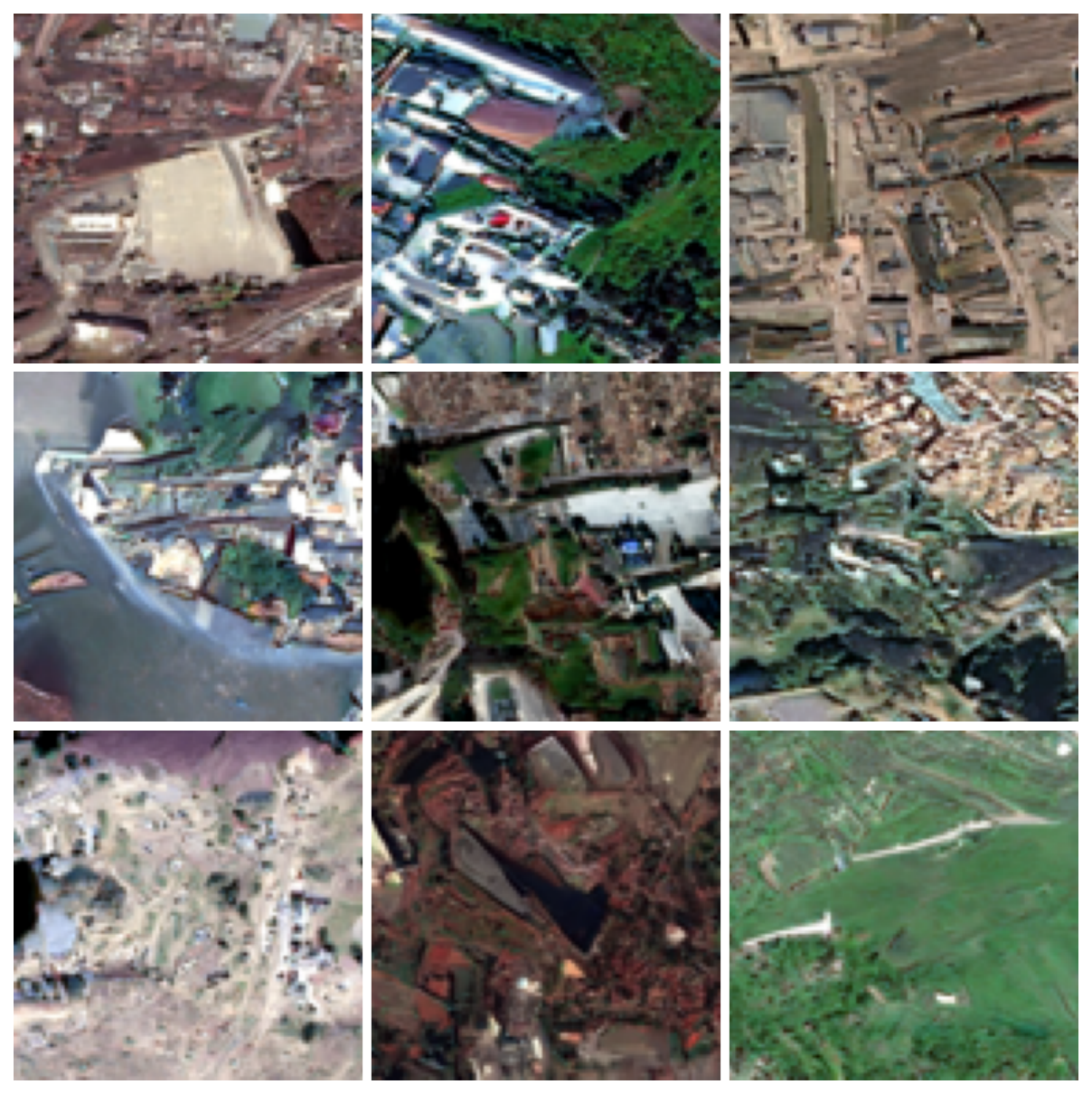}
\end{tabular}

\caption{
Generated images after fine-tuning on the validation splits of three WILDS datasets. Fine-tuning starts from a pretrained Cifar-10 model which has been fully fine-tuned on the train set of the corresponding WILDS dataset.
}
\label{fig:small_dist_shift_images}
\end{figure}

\subsection{Extended Stability and Convergence Results} \label{ap:stability}

\begin{table}[H]
\centering
\caption{Convergence stability comparison over large distribution shifts. Each method starts from a pretrained Cifar-10 model, and is fine-tuned on a WILDS train set.}
\label{tab:fid_monotonicity}
\begin{tabular}{clcccccc} 
\toprule
& \textbf{Objective} 
& \textbf{Inst. Variance} $\downarrow$
& \textbf{Avg. Conv. Rate} $\uparrow$
& \textbf{Spearman $\rho$} $\downarrow$ \\
\midrule

\multirow{3}{*}{RxRx1} 
& Full (CFM)    & 191.685 & 3.110 $\times10^{-2}$ & -0.560 \\
& LoRA (CFM)   & 166.996 & 3.058 $\times10^{-2}$ & -0.602 \\
& Full (GFT)        & \textbf{117.470} & \textbf{4.820} $\times \textbf{10}^{\textbf{-2}}$ & \textbf{-0.800} \\ \hline

\multirow{3}{*}{FMoW} 
& Full (CFM)    & 90.369 & \textbf{2.827} $\times 
\textbf{10}^{\textbf{-2}}$ & -0.584 \\
& LoRA (CFM)  & 66.178 & 2.154 $\times10^{-2}$ & -0.552 \\
& Full (GFT)       & \textbf{43.226} & 2.419 $\times10^{-2}$ & \textbf{-0.759} \\ 

\bottomrule
\end{tabular}
\end{table}

\begin{table}[H]
\centering
\caption{Convergence stability comparison over small distribution shifts. Each method starts from the base Cifar-10 model after it has been fully fine-tuned on a WILDS train set, and is fine-tuned for the corresponding validation set.}
\label{tab:fid_monotonicity}
\begin{tabular}{clccc} 
\toprule
& \textbf{Objective} 
& \textbf{Inst. Variance} $\downarrow$
& \textbf{Avg. Conv. Rate} $\uparrow$
& \textbf{Spearman $\rho$} $\downarrow$ \\
\midrule

\multirow{3}{*}{RxRx1} 
& Full (CFM)    & 10.654 & 9.779 $\times10^{-3}$ & -0.165 \\
& LoRA (CFM)   & 45.366 & \textbf{1.014} $\times \textbf{10}^{\textbf{-2}}$ & -0.329 \\
& LoRA (GFT)      & \textbf{2.347} & 3.743 $\times10^{-3}$ & \textbf{-0.518} \\ \hline

\multirow{3}{*}{FMoW} 
& Full (CFM)    & 12.334 & 4.456 $\times10^{-3}$ & -0.550 \\
& LoRA (CFM)   & 9.236 & 1.406 $\times10^{-3}$ & -0.125 \\
& LoRA (GFT)      & \textbf{2.881} & \textbf{6.157} $\times \textbf{10}^{\textbf{-3}}$ & \textbf{-0.884} \\ 

\bottomrule
\end{tabular}
\end{table}

%% file: app_tech2.tex
\section{Reward vs Sample-based Fine-Tuning Distribution}\label{ap:reward_vs_samples}

Reward-based fine-tuning methods assume access to a known reward model \(r(x)\), and the goal of fine-tuning is to train a model to generate samples from the tilted distribution

\[p^*(x) \propto p^{\text{base}}(x) \exp (r(x))\]

In the case where we are directly given a fine-tuning target distribution \(q\) (in the form of a dataset of samples from \(q\)), our goal is now to train a model which generates samples from 

\begin{align}
    q(x) = p(x)w(x), \qquad w(x) = \frac{q(x)}{p(x)}
\end{align}

In practice, we do not know the true target distribution \(q\). However, we can prove that the density ratio \(w\) is exactly the Radon-Nikodym derivative.

\textit{proof.} 

Let \(\mathbb{P}_p\) and \(\mathbb{P}_q\) be the path measures associated with the continuous processes which generate samples from the base and fine-tuned target distributions \(p\) and \(q\), respectively. In order to apply the Radon-Nikodym theorem, we assume that \(\mathbb{P}_q\) is absolutely continuous with respect to \(\mathbb{P}_p\). Then, there exists a unique function \(w(x)\) such that

\[\mathbb{P}_q(A) = \int_A w(x)  d\mathbb{P}_p\]

where \(A\) is any measurable set in the sample space. The function \(w\) is the Radon-Nikodym derivative, and is denoted as 

\[w = \frac{d \mathbb{P}_q}{d \mathbb{P}_p}\]

Given that \(\mathbb{P}_p, \mathbb{P}_q\) are defined over a space with an underlying Lebesgue base measure \(\lambda\), we can write them in terms of their probability density functions.

\[\mathbb{P}_p = \int p(x) d\lambda \qquad \qquad \mathbb{P}_q = \int q(x) d\lambda\]
\[p(x) = \frac{d \mathbb{P}_p}{d \lambda} \qquad \qquad q(x) = \frac{d \mathbb{P}_q}{d \lambda}\]

We can then apply the chain rule for Radon-Nikodym derivatives to recover the density ratio.

\[w(x) = \frac{d \mathbb{P}_q}{d \mathbb{P}_p}(x) = \frac{d \mathbb{P}_q / d \lambda}{d \mathbb{P}_p / d \lambda}(x) = \frac{q(x)}{p(x)}\]

\section{Corresponding Reward Function}\label{ap:reward}

Reward-based methods fine-tune the base model to achieve generation from a tilted distribution defined by an external reward function \(r(x)\).

\begin{equation}
    p_\theta^*(x) \propto p_{\theta_0}(x) \exp (r(x))
\end{equation}

The reward function which results in an equivalent optimum to our proposed objective function is

\begin{equation}
    r(x) = \frac{1}{1+\beta} \log \left( \frac{q(x)}{p_{\theta_0}(x)} \right)
\end{equation}

We can easily verify this by substituting this expression into the tilted distribution. 

\begin{equation}
    p_\theta^*(x) \propto p_{\theta_0}(x) \exp \left( \frac{1}{1+\beta} \log \left( \frac{q(x)}{p_{\theta_0}(x)} \right) \right)
\end{equation}

\begin{equation}
    = p_{\theta_0}(x) \left( \frac{q(x)}{p_{\theta_0}(x)} \right)^{\frac{1}{1+\beta}}
\end{equation}

\begin{equation}
    = p_{\theta_0}(x)^{\frac{\beta}{1+\beta}}q(x)^{\frac{1}{1+\beta}}
\end{equation}

Since the distribution \(q(x)\) is unknown, this reward function is not tractable in it's current form. However, in Appendix \ref{ap:reward_vs_samples} we prove the relationship between the density ratio and the Radon-Nikodym derivative, and in Appendix \ref{ap:PL_KL} we give the expression for the logarithm of the Radon-Nikodym derivative. Putting these together, we can rewrite the reward function as:

\begin{equation}
    r(x) = \frac{1}{1+\beta} \left( \int_0^1 \sigma_t^{-1}(v_q - v_{\theta_0})^\intercal dB_t   -  \frac{1}{2} \int_0^1 \| \sigma_t^{-1} (v_q - v_{\theta_0})\|^2 dt \right)
\end{equation}

The reward function is now written in terms of the drift terms of the base, target, and fine-tuned processes. As a result, this reward is only tractable when it is conditioned on given source and target samples. This is because the vector field \(v_q\) is unknown, and can only be defined conditionally given known samples. In the setting where a memoryless noise schedule is required to achieve unbiased convergence, this reward function is therefore intractable.

\section{Equivalence Between Total and Source-Conditional KL Divergence}\label{ap:KL_equivalence}

Let \(\mathbb{P}_f\) and \(\mathbb{P}_g\) be the path measures of two SDEs on the time interval \([0,1]\), with initial samples from the same source distribution and the same diffusion coefficient \(\sigma_t\). 

\begin{equation}
    \mathbb{P}_f: \quad dX_t = f_t(X_t, t) dt + \sigma_tdB_t, \qquad X_0 \sim p_0
\end{equation}
    
\begin{equation}
    \mathbb{P}_g: \quad dX_t = g_t(X_t, t) dt + \sigma_tdB_t, \qquad X_0 \sim p_0
\end{equation}

We assume that the standard conditions of Girsanov's Theorem hold, such that \(\mathbb{P}_f\) and \(\mathbb{P}_g\) are mutually absolutely continuous. The path measure \(\mathbb{P}\) for any stochastic process starting at the state \(X_0\) can be decomposed into the product of the initial distribution \(p_0(X_0)\) and the conditional law of the paths given this starting point, \(\mathbb{P}(\cdot | X_0)\).

\begin{equation}
\mathbb{P}(\cdot) = \int \mathbb{P}(\cdot|X_0) p_0(X_0) dX_0 .
\end{equation}

The path measure \(\mathbb{P}\) then admits a disintegration with respect to \(X_0\).

\begin{equation}
    d \mathbb{P} = p_0(X_0)dX_0 \cdot d \mathbb{P}(A | X_0)
\end{equation}

Since the path measures \(\mathbb{P}_f\) and \(\mathbb{P}_g\) share the same source distribution, applying this chain rule to both yields

\begin{equation}
    d \mathbb{P}_f = p_0(X_0)dX_0 \cdot d \mathbb{P}_f(\cdot | X_0)
\end{equation}

\begin{equation}
    d \mathbb{P}_g = p_0(X_0)dX_0 \cdot d \mathbb{P}_g(\cdot | X_0)
\end{equation}

We can then write the expression for the conditional Radon-Nikodym derivative of \(\mathbb{P}_f\) and \(\mathbb{P}_g\), pointwise over \(X_0\).

\begin{equation}
    \frac{d \mathbb{P}_f}{d \mathbb{P}_g} = \frac{p_0(X_0)}{p_0(X_0)} \times \frac{d \mathbb{P}_f (\cdot | X_0)}{d \mathbb{P}_g (\cdot | X_0)}
\end{equation}

\begin{equation}
    \frac{d \mathbb{P}_f}{d \mathbb{P}_g} = \frac{d \mathbb{P}_f (\cdot | X_0)}{d \mathbb{P}_g (\cdot | X_0)}
\end{equation}

Therefore, the Radon-Nikodym derivative of the path measures conditioned on a given source sample \(X_0\) is equivalent to the total derivative when the path measures share the same initial distribution. It follows from the proof given in appendix \ref{ap:PL_KL} that

\begin{equation}
    KL(\mathbb{P}_f(\cdot | X_0) \| \mathbb{P}_g (\cdot | X_0)) = - \mathbb{E}_{\mathbb{P}_f (\cdot | X_0)} \left[ \int_0^1 \sigma_t^{-1} (g_t - f_t)^\intercal dB_t \right] + \mathbb{E}_{\mathbb{P}_f (\cdot | X_0)} \left[ \frac{1}{2} \int_0^1 \| \sigma_t^{-1} (g_t - f_t) \|^2 dt \right]
\end{equation}

The stochastic integral is a Martingale under the path measure \(\mathbb{P}_f\), and its expectation vanishes.

\begin{equation}
    KL(\mathbb{P}_f(\cdot | X_0) \| \mathbb{P}_g (\cdot | X_0)) = \mathbb{E}_{\mathbb{P}_f (\cdot | X_0)} \left[ \frac{1}{2} \int_0^1 \| \sigma_t^{-1} (g_t - f_t) \|^2 dt \right]
\end{equation}

We can take the expectation over initial states \(X_0\) to get

\begin{equation}
    \mathbb{E}_{X_0 \sim p_0} \left[ KL(\mathbb{P}_f(\cdot | X_0) \| \mathbb{P}_g (\cdot | X_0)) \right] = \mathbb{E}_{X_0 \sim p_0} \left[ \mathbb{E}_{\mathbb{P}_f (\cdot | X_0)} \left[ \frac{1}{2} \int_0^1 \| \sigma_t^{-1} (g_t - f_t) \|^2 dt \right] \right]
\end{equation}

By the Law of Total Expectation, nesting these expectations is equivalent to taking the expectation over the full path measure \(\mathbb{P}_f\). We therefore find the final identity

\begin{equation}
    KL(\mathbb{P}_f \| \mathbb{P}_g) = \mathbb{E}_{X_0 \sim p_0} \left[ KL(\mathbb{P}_f(\cdot | X_0) \| \mathbb{P}_g (\cdot | X_0)) \right] = \mathbb{E}_{\mathbb{P}_f } \left[ \frac{1}{2} \int_0^1 \| \sigma_t^{-1} (g_t - f_t) \|^2 dt \right]
\end{equation}

\section{Equivalence Between Total and OT-Conditional KL Divergence}\label{ap:KL_OT_equivalence}

Let \(\mathbb{P}_f\) and \(\mathbb{P}_g\) be the path measures of two SDEs on the time interval \([0,1]\), with initial samples from the same source distribution and the same diffusion coefficient \(\sigma_t\). 

\begin{equation}
    \mathbb{P}_f: \quad dX_t = f_t(X_t, t) dt + \sigma_tdB_t, \qquad X_0 \sim p_0
\end{equation}
    
\begin{equation}
    \mathbb{P}_g: \quad dX_t = g_t(X_t, t) dt + \sigma_tdB_t, \qquad X_0 \sim p_0
\end{equation}

We assume that the standard conditions of Girsanov's Theorem hold, such that \(\mathbb{P}_f\) and \(\mathbb{P}_g\) are mutually absolutely continuous. Let \(\pi \in \Pi (p_0, q)\) be any coupling between the source and target distributions that admits the marginals \(\int \pi(X_0, X_1) dX_1 = p_0(X_0)\) and \(\int \pi(X_0, X_1) dX_0 = q(X_1)\). Consider the following expected KL divergence over jointly drawn samples:

\begin{equation}
    \mathbb{E}_{(X_0, X_1) \sim \pi} \left[  KL(\mathbb{P}_f(\cdot | X_0) \| \mathbb{P}_g(\cdot | X_0, X_1))  \right]
\end{equation}

Using the definition of \(\pi\), we can write the integral form of this expectation.

\begin{equation}
    \mathcal{L}_\pi = \int \int KL(\mathbb{P}_f(\cdot | X_0) \| \mathbb{P}_g(\cdot | X_0, X_1)) \pi(X_0, X_1) dX_0 dX_1
\end{equation}

Expanding the definition of the KL divergence for path measures:

\begin{equation} 
    \mathcal{L}_\pi = \int \pi(X_0, X_1) \left( \int \log \frac{d\mathbb{P}_f(\cdot | X_0)}{d\mathbb{P}_g(\cdot | X_0, X_1)} d\mathbb{P}_f(\cdot | X_0) \right) dX_0 dX_1
\end{equation}

For the target process \(g\), the relationship between the measure \(\mathbb{P}_g(\cdot | X_0, X_1)\) and the source-conditioned measure \(\mathbb{P}_g(\cdot | X_0)\) is given by Bayes Theorem,

\begin{equation} 
    d\mathbb{P}_g(\cdot | X_0, X_1) = \frac{d\mathbb{P}_g(\cdot | X_0) p_g(X_1 | \cdot)}{p_g(X_1 | X_0)} 
\end{equation}

where \(p_g(X_1 | \cdot)\) is the density of the terminal state of a path. Since we condition directly on \(X_1\), this terminal-state likelihood evaluates to a Dirac measure, which is independent of \(f\). We now substitute this into the Radon-Nikodym derivative.

\begin{equation} 
    \log \frac{d\mathbb{P}_f(\cdot | X_0)}{d\mathbb{P}_g(\cdot | X_0, X_1)} = \log \frac{d\mathbb{P}_f(\cdot | X_0)}{d\mathbb{P}_g(\cdot | X_0)} + \log \frac{p_g(X_1 | X_0)}{p_g(X_1 | \cdot)} 
\end{equation}

Finally, we plug this back into the coupled objective and distribute the expectation.

\begin{equation} 
    \mathcal{L}_\pi = \mathbb{E}_\pi \left[ KL(\mathbb{P}_f(\cdot | X_0) \| \mathbb{P}_g(\cdot | X_0)) \right] + \mathbb{E}_\pi \mathbb{E}_{\mathbb{P}_f(\cdot | X_0)} \left[ \log \frac{p_g(X_1 | X_0)}{p_g(X_1 | \cdot)} \right] 
\end{equation}
\begin{equation}
    = \int p_0(X_0) KL(\mathbb{P}_f(\cdot | X_0) \| \mathbb{P}_g(\cdot | X_0)) dX_0 + \mathcal{C}(\pi, g)
\end{equation}

The first term is exactly the global KL divergence \(KL(\mathbb{P}_f \| \mathbb{P}_g)\). The second term depends only on the coupling and the target process, and is therefore constant with respect to the parameters of \(f\). This is because the ratio is determined only by the process \(g\) and the terminal state \(X_1\), so the expectation over \(\mathbb{P}_f(\cdot | X_0)\) does not introduce any dependence on \(f\). For gradient-based optimization, this term does not affect convergence.

\section{Terminal Distribution Objective}\label{sc:TD_objective}

We examine a related objective function, which works on the terminal distributions of the generative process rather than the path measures. Consider the following constrained SDE,

\begin{equation}
    \min_{p_\theta} \left(  KL(p_\theta \| q) + \beta KL(p_\theta \| p_{\theta_0})  \right)\label{eq:TD_cost}
\end{equation}
\begin{equation}
    \text{s.t.  } dX_t = v_\theta(X_t, t) dt + \sigma_t dB_t
\end{equation}

where \(p_\theta\) and \(p_{\theta_0}\) are the terminal distributions from which the fine-tuned and pretrained models generates samples, respectively, and \(\beta\) is a scalar hyperparameter. The objective function corresponding to this constraint is

\begin{equation}
    \mathcal{L}(p_\theta) = KL(p_\theta \| q) + \beta KL(p_\theta \| p_{\theta_0})\label{eq:TD_objective}.
\end{equation}

We want to show that minimizing this objective function results in sampling from the desired target distribution \(q\). We start by expressing the constraint in integral form with respect to the Lebesgue base measure \(\lambda\) for the continuous Euclidean space \(\mathbb{R}^d\).

\begin{equation}
    \mathcal{L}(p_\theta) = \int p_\theta(X) \log \frac{p_\theta(X)}{q(X)} d\lambda(X) + \int \beta \cdot p_\theta(X) \log \frac{p_\theta(X)}{p_{\theta_0}(X)} d\lambda(X)
\end{equation}

We minimize \(\mathcal{L}(p_\theta)\) subject to the constraint that \(p_\theta\) is a valid probability distribution:
\(\int_{\mathcal{X}} p_\theta(X)d \lambda(X) = 1\). To do so, we introduce a Lagrange multiplier \(\mu\) for this constraint, and form the Lagrangian \(L(p_\theta)\).

\begin{equation}
    L(p_\theta) = \mathcal{L}(p_\theta) - \mu \left( \int_{\mathcal{X}} p_\theta(X)d \lambda(X) - 1 \right)
\end{equation}

\begin{equation}
    L(p_\theta) = \int_{\mathcal{X}} \left[ p_\theta(X) \log \frac{p_\theta(X)}{q(X)} + \beta \cdot p_\theta(X) \log \frac{p_\theta(X)}{p_{\theta_0}(X)} - \mu p_\theta(X)\right]d\lambda(X) + \mu 
\end{equation}

We now minimize by setting the derivative with respect to \(p_\theta\) to zero, and solving to find the optimal \(p_\theta^*\).

\begin{equation}
    \frac{dL}{dp_\theta} = (\log p_\theta + 1 - \log q) + \beta (\log p_\theta + 1 - \log p_{\theta_0}) - \mu = 0
\end{equation}

Finally, we solve for the optimal \(p^*_\theta(X)\).

\begin{equation}
    (1 + \beta) \log p_\theta + 1 + \beta - \log q - \beta \log p_{\theta_0} - \mu= 0
\end{equation}

\begin{equation}
    (1 + \beta) \log p_\theta = \log q + \beta \log p_{\theta_0} + \mu -1 -\beta
\end{equation}

\begin{equation}
    \log p_\theta = \frac{1}{1 + \beta}(\log q) +  \frac{\beta}{1 + \beta}  (\log p_{\theta_0}) + \frac{\mu - 1 - \beta}{1 + \beta}
\end{equation}

\begin{equation}
    p_\theta = \exp\left( \frac{1}{1 + \beta}(\log q) +  \frac{\beta}{1 + \beta}  (\log p_{\theta_0}) + \frac{\mu - 1 - \beta}{1 + \beta} \right)
\end{equation}

\begin{equation}
    p^*_\theta = q^{\frac{1}{1 + \beta}} \cdot p_{\theta_0}^{\frac{\beta}{1 + \beta}} \exp\left( \frac{\mu - 1 - \beta}{1 + \beta} \right)
\end{equation}

We set \(Z\) to be a constant that absorbs all terms not related to \(X\). Then, the final optimal distribution is

\begin{equation}
    p^*_\theta(X) = Z \cdot q(X)^{\frac{1}{1 + \beta}} \cdot   p_{\theta_0}(X)^{\frac{\beta}{1 + \beta}}\label{eq:TD_optimal}.
\end{equation}

\paragraph{Cooling Schedule}

Note that the exponents \(\frac{1}{1 + \beta}\) and \(\frac{\beta}{1 + \beta}\) sum to 1, meaning that we have defined a weighted average of the pretrained and fine-tuned target distributions. This means that hyperparameter \(\beta\) can be adjusted to achieve a fine-tuned model that generates from a user-defined target distribution. 

\begin{equation}
    p_\theta^*(X) \propto 
    \begin{cases}
        q(X) & \text{if } \beta = 0 \\
        \sqrt{p_{\theta_0}(X)q(X)} & \text{if } \beta = 1 \\
        p_{\theta_0}(X) & \text{if } \beta = \infty
    \end{cases}
\end{equation}

This means that the second KL term is acting as a regularization for the change from the pretrained model, and the hyperparameter \(\beta\) controls the strength of this regularization. \(\beta\) can therefore be used as a cooling schedule for simulated annealing \cite{SimAn} from the pretrained distribution \(p_{\theta_0} \approx p\) to the fine-tuned target distribution \(p_\theta \approx q\).

%% file: main.bbl
\begin{thebibliography}{45}
\providecommand{\natexlab}[1]{#1}
\providecommand{\url}[1]{\texttt{#1}}
\expandafter\ifx\csname urlstyle\endcsname\relax
  \providecommand{\doi}[1]{doi: #1}\else
  \providecommand{\doi}{doi: \begingroup \urlstyle{rm}\Url}\fi

\bibitem[Albergo et~al.(2025)Albergo, Boffi, and Vanden-Eijnden]{albergo2025stochasticinterpolantsunifyingframework}
Michael~S. Albergo, Nicholas~M. Boffi, and Eric Vanden-Eijnden.
\newblock Stochastic interpolants: A unifying framework for flows and diffusions, 2025.
\newblock URL \url{https://arxiv.org/abs/2303.08797}.

\bibitem[Bao et~al.(2024)Bao, Zhang, and Zhang]{bao2024ensemble}
Feng Bao, Zezhong Zhang, and Guannan Zhang.
\newblock An ensemble score filter for tracking high-dimensional nonlinear dynamical systems.
\newblock \emph{Computer Methods in Applied Mechanics and Engineering}, 432:\penalty0 117447, 2024.

\bibitem[Ben-Hamu et~al.(2024)Ben-Hamu, Puny, Gat, Karrer, Singer, and Lipman]{ben-hamu2024dflow}
Heli Ben-Hamu, Omri Puny, Itai Gat, Brian Karrer, Uriel Singer, and Yaron Lipman.
\newblock D-flow: Differentiating through flows for controlled generation.
\newblock In \emph{Forty-first International Conference on Machine Learning}, 2024.
\newblock URL \url{https://openreview.net/forum?id=SE20BFqj6J}.

\bibitem[Black et~al.(2024)Black, Janner, Du, Kostrikov, and Levine]{black2024training}
Kevin Black, Michael Janner, Yilun Du, Ilya Kostrikov, and Sergey Levine.
\newblock Training diffusion models with reinforcement learning.
\newblock In \emph{The Twelfth International Conference on Learning Representations}, 2024.
\newblock URL \url{https://openreview.net/forum?id=YCWjhGrJFD}.

\bibitem[Blessing et~al.(2025)Blessing, Berner, Richter, Domingo-Enrich, Du, Vahdat, and Neumann]{blessing2025trust}
Denis Blessing, Julius Berner, Lorenz Richter, Carles Domingo-Enrich, Yuanqi Du, Arash Vahdat, and Gerhard Neumann.
\newblock Trust region constrained measure transport in path space for stochastic optimal control and inference.
\newblock In \emph{The Thirty-ninth Annual Conference on Neural Information Processing Systems}, 2025.
\newblock URL \url{https://openreview.net/forum?id=6RlbOEcOS4}.

\bibitem[Boffi et~al.(2025)Boffi, Albergo, and Vanden-Eijnden]{boffi2025buildconsistencymodellearning}
Nicholas~M. Boffi, Michael~S. Albergo, and Eric Vanden-Eijnden.
\newblock How to build a consistency model: Learning flow maps via self-distillation, 2025.
\newblock URL \url{https://arxiv.org/abs/2505.18825}.

\bibitem[Bándi et~al.(2019)Bándi, Geessink, Manson, Van~Dijk, Balkenhol, Hermsen, Ehteshami~Bejnordi, Lee, Paeng, Zhong, Li, Zanjani, Zinger, Fukuta, Komura, Ovtcharov, Cheng, Zeng, Thagaard, Dahl, Lin, Chen, Jacobsson, Hedlund, Çetin, Halıcı, Jackson, Chen, Both, Franke, Küsters-Vandevelde, Vreuls, Bult, van Ginneken, van~der Laak, and Litjens]{Camelyon17}
Péter Bándi, Oscar Geessink, Quirine Manson, Marcory Van~Dijk, Maschenka Balkenhol, Meyke Hermsen, Babak Ehteshami~Bejnordi, Byungjae Lee, Kyunghyun Paeng, Aoxiao Zhong, Quanzheng Li, Farhad~Ghazvinian Zanjani, Svitlana Zinger, Keisuke Fukuta, Daisuke Komura, Vlado Ovtcharov, Shenghua Cheng, Shaoqun Zeng, Jeppe Thagaard, Anders~B. Dahl, Huangjing Lin, Hao Chen, Ludwig Jacobsson, Martin Hedlund, Melih Çetin, Eren Halıcı, Hunter Jackson, Richard Chen, Fabian Both, Jörg Franke, Heidi Küsters-Vandevelde, Willem Vreuls, Peter Bult, Bram van Ginneken, Jeroen van~der Laak, and Geert Litjens.
\newblock From detection of individual metastases to classification of lymph node status at the patient level: The camelyon17 challenge.
\newblock \emph{IEEE Transactions on Medical Imaging}, 38\penalty0 (2):\penalty0 550--560, 2019.
\newblock \doi{10.1109/TMI.2018.2867350}.

\bibitem[Christie et~al.(2018)Christie, Fendley, Wilson, and Mukherjee]{fmow2018}
Gordon Christie, Neil Fendley, James Wilson, and Ryan Mukherjee.
\newblock Functional map of the world.
\newblock In \emph{CVPR}, 2018.

\bibitem[Clark et~al.(2024)Clark, Vicol, Swersky, and Fleet]{clark2024directly}
Kevin Clark, Paul Vicol, Kevin Swersky, and David~J. Fleet.
\newblock Directly fine-tuning diffusion models on differentiable rewards.
\newblock In \emph{The Twelfth International Conference on Learning Representations}, 2024.
\newblock URL \url{https://openreview.net/forum?id=1vmSEVL19f}.

\bibitem[Davis et~al.(2024)Davis, Kessler, Petrache, Ceylan, Bronstein, and Bose]{Fisher}
Oscar Davis, Samuel Kessler, Mircea Petrache, \.{I}smail~\.{I}lkan Ceylan, Michael Bronstein, and Avishek~Joey Bose.
\newblock Fisher flow matching for generative modeling over discrete data.
\newblock In A.~Globerson, L.~Mackey, D.~Belgrave, A.~Fan, U.~Paquet, J.~Tomczak, and C.~Zhang, editors, \emph{Advances in Neural Information Processing Systems}, volume~37, pages 139054--139084. Curran Associates, Inc., 2024.
\newblock \doi{10.52202/079017-4413}.
\newblock URL \url{https://proceedings.neurips.cc/paper_files/paper/2024/file/fadec8f2e65f181d777507d1df69b92f\\-Paper-Conference.pdf}.

\bibitem[Domingo-Enrich et~al.(2025)Domingo-Enrich, Drozdzal, Karrer, and Chen]{domingo-enrich2025adjoint}
Carles Domingo-Enrich, Michal Drozdzal, Brian Karrer, and Ricky T.~Q. Chen.
\newblock Adjoint matching: Fine-tuning flow and diffusion generative models with memoryless stochastic optimal control.
\newblock In \emph{The Thirteenth International Conference on Learning Representations}, 2025.
\newblock URL \url{https://openreview.net/forum?id=xQBRrtQM8u}.

\bibitem[Esser et~al.(2024)Esser, Kulal, Blattmann, Entezari, M\"{u}ller, Saini, Levi, Lorenz, Sauer, Boesel, Podell, Dockhorn, English, and Rombach]{pmlr-v235-esser24a}
Patrick Esser, Sumith Kulal, Andreas Blattmann, Rahim Entezari, Jonas M\"{u}ller, Harry Saini, Yam Levi, Dominik Lorenz, Axel Sauer, Frederic Boesel, Dustin Podell, Tim Dockhorn, Zion English, and Robin Rombach.
\newblock Scaling rectified flow transformers for high-resolution image synthesis.
\newblock In Ruslan Salakhutdinov, Zico Kolter, Katherine Heller, Adrian Weller, Nuria Oliver, Jonathan Scarlett, and Felix Berkenkamp, editors, \emph{Proceedings of the 41st International Conference on Machine Learning}, volume 235 of \emph{Proceedings of Machine Learning Research}, pages 12606--12633. PMLR, 21--27 Jul 2024.
\newblock URL \url{https://proceedings.mlr.press/v235/esser24a.html}.

\bibitem[Fan et~al.(2025{\natexlab{a}})Fan, Cheng, Shen, Zhou, and Liu]{fan2025finetuningflowmatchinggenerative}
Jiajun Fan, Chaoran Cheng, Shuaike Shen, Xiangxin Zhou, and Ge~Liu.
\newblock Fine-tuning flow matching generative models with intermediate feedback, 2025{\natexlab{a}}.
\newblock URL \url{https://arxiv.org/abs/2510.18072}.

\bibitem[Fan et~al.(2025{\natexlab{b}})Fan, Shen, Cheng, Chen, Liang, and Liu]{fan2025online}
Jiajun Fan, Shuaike Shen, Chaoran Cheng, Yuxin Chen, Chumeng Liang, and Ge~Liu.
\newblock Online reward-weighted fine-tuning of flow matching with wasserstein regularization.
\newblock In \emph{The Thirteenth International Conference on Learning Representations}, 2025{\natexlab{b}}.
\newblock URL \url{https://openreview.net/forum?id=2IoFFexvuw}.

\bibitem[Fan et~al.(2023)Fan, Watkins, Du, Liu, Ryu, Boutilier, Abbeel, Ghavamzadeh, Lee, and Lee]{fan2023reinforcement}
Ying Fan, Olivia Watkins, Yuqing Du, Hao Liu, Moonkyung Ryu, Craig Boutilier, Pieter Abbeel, Mohammad Ghavamzadeh, Kangwook Lee, and Kimin Lee.
\newblock Reinforcement learning for fine-tuning text-to-image diffusion models.
\newblock In \emph{Thirty-seventh Conference on Neural Information Processing Systems}, 2023.
\newblock URL \url{https://openreview.net/forum?id=8OTPepXzeh}.

\bibitem[Fotiadis et~al.(2025)Fotiadis, Brenowitz, Geffner, Cohen, Pritchard, Vahdat, and Mardani]{fotiadis2025adaptive}
Stathi Fotiadis, Noah~D Brenowitz, Tomas Geffner, Yair Cohen, Michael Pritchard, Arash Vahdat, and Morteza Mardani.
\newblock Adaptive flow matching for resolving small-scale physics.
\newblock In \emph{Forty-second International Conference on Machine Learning}, 2025.
\newblock URL \url{https://openreview.net/forum?id=YJ1My9ttEN}.

\bibitem[Geng et~al.(2026)Geng, Deng, Bai, Kolter, and He]{geng2026mean}
Zhengyang Geng, Mingyang Deng, Xingjian Bai, J~Zico Kolter, and Kaiming He.
\newblock Mean flows for one-step generative modeling.
\newblock In \emph{The Thirty-ninth Annual Conference on Neural Information Processing Systems}, 2026.
\newblock URL \url{https://openreview.net/forum?id=uWj4s7rMnR}.

\bibitem[Goodfellow et~al.(2014)Goodfellow, Pouget-Abadie, Mirza, Xu, Warde-Farley, Ozair, Courville, and Bengio]{NIPS2014_f033ed80}
Ian~J. Goodfellow, Jean Pouget-Abadie, Mehdi Mirza, Bing Xu, David Warde-Farley, Sherjil Ozair, Aaron Courville, and Yoshua Bengio.
\newblock Generative adversarial nets.
\newblock In Z.~Ghahramani, M.~Welling, C.~Cortes, N.~Lawrence, and K.Q. Weinberger, editors, \emph{Advances in Neural Information Processing Systems}, volume~27. Curran Associates, Inc., 2014.
\newblock URL \url{https://proceedings.neurips.cc/paper_files/paper/2014/file/f033ed80deb0234979a61f95710dbe25-Paper.pdf}.

\bibitem[Havens et~al.(2025)Havens, Miller, Yan, Domingo-Enrich, Sriram, Levine, Wood, Hu, Amos, Karrer, Fu, Liu, and Chen]{havens2025adjoint}
Aaron~J Havens, Benjamin~Kurt Miller, Bing Yan, Carles Domingo-Enrich, Anuroop Sriram, Daniel~S. Levine, Brandon~M Wood, Bin Hu, Brandon Amos, Brian Karrer, Xiang Fu, Guan-Horng Liu, and Ricky T.~Q. Chen.
\newblock Adjoint sampling: Highly scalable diffusion samplers via adjoint matching.
\newblock In \emph{Forty-second International Conference on Machine Learning}, 2025.
\newblock URL \url{https://openreview.net/forum?id=6Eg1OrHmg2}.

\bibitem[He et~al.(2024)He, Wang, Li, and Zhao]{gradual_DA}
Yifei He, Haoxiang Wang, Bo~Li, and Han Zhao.
\newblock Gradual domain adaptation: Theory and algorithms.
\newblock \emph{Journal of Machine Learning Research}, 25\penalty0 (361):\penalty0 1--40, 2024.
\newblock URL \url{http://jmlr.org/papers/v25/23-1180.html}.

\bibitem[Ho and Salimans(2022)]{ho2022classifierfreediffusionguidance}
Jonathan Ho and Tim Salimans.
\newblock Classifier-free diffusion guidance, 2022.
\newblock URL \url{https://arxiv.org/abs/2207.12598}.

\bibitem[Hu et~al.(2022)Hu, yelong shen, Wallis, Allen-Zhu, Li, Wang, Wang, and Chen]{hu2022lora}
Edward~J Hu, yelong shen, Phillip Wallis, Zeyuan Allen-Zhu, Yuanzhi Li, Shean Wang, Lu~Wang, and Weizhu Chen.
\newblock Lo{RA}: Low-rank adaptation of large language models.
\newblock In \emph{International Conference on Learning Representations}, 2022.
\newblock URL \url{https://openreview.net/forum?id=nZeVKeeFYf9}.

\bibitem[Jing et~al.(2024)Jing, Berger, and Jaakkola]{jing2024alphafold}
Bowen Jing, Bonnie Berger, and Tommi Jaakkola.
\newblock Alphafold meets flow matching for generating protein ensembles.
\newblock In \emph{Forty-first International Conference on Machine Learning}, 2024.

\bibitem[Karatzas and Shreve(1988)]{alma9919303981206531}
Ioannis Karatzas and Steven. Shreve.
\newblock \emph{Brownian Motion and Stochastic Calculus / by Ioannis Karatzas, Steven Shreve.}
\newblock Graduate Texts in Mathematics, 113. Springer New York, New York, NY, 1st ed. 1988. edition, 1988.
\newblock ISBN 1-4684-0302-8.

\bibitem[Kingma and Welling(2014)]{Kingma2014}
Diederik~P. Kingma and Max Welling.
\newblock {Auto-Encoding Variational Bayes}.
\newblock In \emph{2nd International Conference on Learning Representations, {ICLR} 2014, Banff, AB, Canada, April 14-16, 2014, Conference Track Proceedings}, 2014.

\bibitem[Kirkpatrick et~al.(1983)Kirkpatrick, Gelatt, and Vecchi]{SimAn}
S.~Kirkpatrick, C.~D. Gelatt, and M.~P. Vecchi.
\newblock Optimization by simulated annealing.
\newblock \emph{Science}, 220\penalty0 (4598):\penalty0 671--680, 1983.
\newblock \doi{10.1126/science.220.4598.671}.
\newblock URL \url{https://www.science.org/doi/abs/10.1126/science.220.4598.671}.

\bibitem[Klein et~al.(2023)Klein, Kr\"{a}mer, and Noe]{EqFM}
Leon Klein, Andreas Kr\"{a}mer, and Frank Noe.
\newblock Equivariant flow matching.
\newblock In A.~Oh, T.~Naumann, A.~Globerson, K.~Saenko, M.~Hardt, and S.~Levine, editors, \emph{Advances in Neural Information Processing Systems}, volume~36, pages 59886--59910. Curran Associates, Inc., 2023.
\newblock URL \url{https://proceedings.neurips.cc/paper_files/paper/2023/file/bc827452450356f9f558f4e4568d553b\\-Paper-Conference.pdf}.

\bibitem[Koh et~al.(2021)Koh, Sagawa, Marklund, Xie, Zhang, Balsubramani, Hu, Yasunaga, Phillips, Gao, Lee, David, Stavness, Guo, Earnshaw, Haque, Beery, Leskovec, Kundaje, Pierson, Levine, Finn, and Liang]{wilds2021}
Pang~Wei Koh, Shiori Sagawa, Henrik Marklund, Sang~Michael Xie, Marvin Zhang, Akshay Balsubramani, Weihua Hu, Michihiro Yasunaga, Richard~Lanas Phillips, Irena Gao, Tony Lee, Etienne David, Ian Stavness, Wei Guo, Berton~A. Earnshaw, Imran~S. Haque, Sara Beery, Jure Leskovec, Anshul Kundaje, Emma Pierson, Sergey Levine, Chelsea Finn, and Percy Liang.
\newblock {WILDS}: A benchmark of in-the-wild distribution shifts.
\newblock In \emph{International Conference on Machine Learning (ICML)}, 2021.

\bibitem[Krizhevsky(2009)]{Krizhevsky09learningmultiple}
Alex Krizhevsky.
\newblock Learning multiple layers of features from tiny images.
\newblock Technical report, 2009.

\bibitem[Liang et~al.(2025)Liang, Yuan, Gu, Chen, Hang, Cheng, Li, and Zheng]{liang2025aestheticposttrainingdiffusionmodels}
Zhanhao Liang, Yuhui Yuan, Shuyang Gu, Bohan Chen, Tiankai Hang, Mingxi Cheng, Ji~Li, and Liang Zheng.
\newblock Aesthetic post-training diffusion models from generic preferences with step-by-step preference optimization, 2025.
\newblock URL \url{https://arxiv.org/abs/2406.04314}.

\bibitem[Lipman et~al.(2023)Lipman, Chen, Ben-Hamu, Nickel, and Le]{lipman2023flow}
Yaron Lipman, Ricky T.~Q. Chen, Heli Ben-Hamu, Maximilian Nickel, and Matthew Le.
\newblock Flow matching for generative modeling.
\newblock In \emph{The Eleventh International Conference on Learning Representations}, 2023.
\newblock URL \url{https://openreview.net/forum?id=PqvMRDCJT9t}.

\bibitem[Liu et~al.(2023)Liu, Gong, and Liu]{liu2023flow}
Xingchao Liu, Chengyue Gong, and Qiang Liu.
\newblock Flow straight and fast: Learning to generate and transfer data with rectified flow.
\newblock In \emph{International Conference on Learning Representations (ICLR)}, 2023.

\bibitem[Pooladian et~al.(2023)Pooladian, Ben-Hamu, Domingo-Enrich, Amos, Lipman, and Chen]{multisampleFM}
Aram-Alexandre Pooladian, Heli Ben-Hamu, Carles Domingo-Enrich, Brandon Amos, Yaron Lipman, and Ricky T.~Q. Chen.
\newblock Multisample flow matching: straightening flows with minibatch couplings.
\newblock In \emph{Proceedings of the 40th International Conference on Machine Learning}, ICML'23. JMLR.org, 2023.

\bibitem[Potaptchik et~al.(2025)Potaptchik, Lee, and Albergo]{potaptchik2025tiltmatchingscalablesampling}
Peter Potaptchik, Cheuk-Kit Lee, and Michael~S. Albergo.
\newblock Tilt matching for scalable sampling and fine-tuning, 2025.
\newblock URL \url{https://arxiv.org/abs/2512.21829}.

\bibitem[Prabhudesai et~al.(2023)Prabhudesai, Goyal, Pathak, and Fragkiadaki]{prabhudesai2023aligning}
Mihir Prabhudesai, Anirudh Goyal, Deepak Pathak, and Katerina Fragkiadaki.
\newblock Aligning text-to-image diffusion models with reward backpropagation, 2023.

\bibitem[Song et~al.(2021)Song, Sohl-Dickstein, Kingma, Kumar, Ermon, and Poole]{song2021scorebased}
Yang Song, Jascha Sohl-Dickstein, Diederik~P Kingma, Abhishek Kumar, Stefano Ermon, and Ben Poole.
\newblock Score-based generative modeling through stochastic differential equations.
\newblock In \emph{International Conference on Learning Representations}, 2021.
\newblock URL \url{https://openreview.net/forum?id=PxTIG12RRHS}.

\bibitem[Song et~al.(2023)Song, Gong, Xu, Cao, Lan, Ermon, Zhou, and Ma]{EqFM_Molecule}
Yuxuan Song, Jingjing Gong, Minkai Xu, Ziyao Cao, Yanyan Lan, Stefano Ermon, Hao Zhou, and Wei-Ying Ma.
\newblock Equivariant flow matching with hybrid probability transport for 3d molecule generation.
\newblock In A.~Oh, T.~Naumann, A.~Globerson, K.~Saenko, M.~Hardt, and S.~Levine, editors, \emph{Advances in Neural Information Processing Systems}, volume~36, pages 549--568. Curran Associates, Inc., 2023.
\newblock URL \url{https://proceedings.neurips.cc/paper_files/paper/2023/file/01d64478381c33e29ed611f1719f5a37-Paper\\-Conference.pdf}.

\bibitem[Stark et~al.(2024)Stark, Jing, Wang, Corso, Berger, Barzilay, and Jaakkola]{stark2024dirichlet}
Hannes Stark, Bowen Jing, Chenyu Wang, Gabriele Corso, Bonnie Berger, Regina Barzilay, and Tommi Jaakkola.
\newblock Dirichlet flow matching with applications to {DNA} sequence design.
\newblock In \emph{ICLR 2024 Workshop on Machine Learning for Genomics Explorations}, 2024.
\newblock URL \url{https://openreview.net/forum?id=ehYe5bz8H3}.

\bibitem[Sypetkowski et~al.(2023)Sypetkowski, Rezanejad, Saberian, Kraus, Urbanik, Taylor, Mabey, Victors, Yosinski, Sereshkeh, Haque, and Earnshaw]{RxRx1}
Maciej Sypetkowski, Morteza Rezanejad, Saber Saberian, Oren Kraus, John Urbanik, James Taylor, Ben Mabey, Mason Victors, Jason Yosinski, Alborz~Rezazadeh Sereshkeh, Imran Haque, and Berton Earnshaw.
\newblock { RxRx1: A Dataset for Evaluating Experimental Batch Correction Methods }.
\newblock In \emph{2023 IEEE/CVF Conference on Computer Vision and Pattern Recognition Workshops (CVPRW)}, pages 4285--4294, Los Alamitos, CA, USA, June 2023. IEEE Computer Society.
\newblock \doi{10.1109/CVPRW59228.2023.00451}.
\newblock URL \url{https://doi.ieeecomputersociety.org/10.1109/CVPRW59228.2023.00451}.

\bibitem[Tong et~al.(2023)Tong, Malkin, FATRAS, Atanackovic, Zhang, Huguet, Wolf, and Bengio]{tong2023simulationfree}
Alexander Tong, Nikolay Malkin, Kilian FATRAS, Lazar Atanackovic, Yanlei Zhang, Guillaume Huguet, Guy Wolf, and Yoshua Bengio.
\newblock Simulation-free schr\"odinger bridges via score and flow matching.
\newblock In \emph{ICML Workshop on New Frontiers in Learning, Control, and Dynamical Systems}, 2023.
\newblock URL \url{https://openreview.net/forum?id=adkj23mvB0}.

\bibitem[Tong et~al.(2024)Tong, Fatras, Malkin, Huguet, Zhang, Rector-Brooks, Wolf, and Bengio]{tong2024improving}
Alexander Tong, Kilian Fatras, Nikolay Malkin, Guillaume Huguet, Yanlei Zhang, Jarrid Rector-Brooks, Guy Wolf, and Yoshua Bengio.
\newblock Improving and generalizing flow-based generative models with minibatch optimal transport.
\newblock \emph{Transactions on Machine Learning Research}, 2024.
\newblock ISSN 2835-8856.
\newblock URL \url{https://openreview.net/forum?id=CD9Snc73AW}.
\newblock Expert Certification.

\bibitem[Uehara et~al.(2024)Uehara, Zhao, Black, Hajiramezanali, Scalia, Diamant, Tseng, Biancalani, and Levine]{uehara2024finetuningcontinuoustimediffusionmodels}
Masatoshi Uehara, Yulai Zhao, Kevin Black, Ehsan Hajiramezanali, Gabriele Scalia, Nathaniel~Lee Diamant, Alex~M Tseng, Tommaso Biancalani, and Sergey Levine.
\newblock Fine-tuning of continuous-time diffusion models as entropy-regularized control, 2024.
\newblock URL \url{https://arxiv.org/abs/2402.15194}.

\bibitem[Villani(2008)]{villani2008optimal}
C.~Villani.
\newblock \emph{Optimal Transport: Old and New}.
\newblock Grundlehren der mathematischen Wissenschaften. Springer Berlin Heidelberg, 2008.
\newblock ISBN 9783540710493.
\newblock URL \url{https://books.google.com/books?id=NZXiNAEACAAJ}.

\bibitem[Wallace et~al.(2023)Wallace, Dang, Rafailov, Zhou, Lou, Purushwalkam, Ermon, Xiong, Joty, and Naik]{wallace2023diffusion}
Bram Wallace, Meihua Dang, Rafael Rafailov, Linqi Zhou, Aaron Lou, Senthil Purushwalkam, Stefano Ermon, Caiming Xiong, Shafiq Joty, and Nikhil Naik.
\newblock Diffusion model alignment using direct preference optimization, 2023.

\bibitem[Xu et~al.(2023)Xu, Liu, Wu, Tong, Li, Ding, Tang, and Dong]{xu2023imagereward}
Jiazheng Xu, Xiao Liu, Yuchen Wu, Yuxuan Tong, Qinkai Li, Ming Ding, Jie Tang, and Yuxiao Dong.
\newblock Imagereward: learning and evaluating human preferences for text-to-image generation.
\newblock In \emph{Proceedings of the 37th International Conference on Neural Information Processing Systems}, pages 15903--15935, 2023.

\end{thebibliography}
